\documentclass[mnsc,nonblindrev,dvipsnames]{informs4}

\usepackage{natbib}
 \bibpunct[, ]{(}{)}{,}{a}{}{,}%
 \def\bibfont{\small}%

\usepackage[colorlinks=true,bookmarks=false,urlcolor=blue,
     citecolor=blue,linkcolor=blue,bookmarksopen=false,draft=false]{hyperref} 

\usepackage{tikz}
\usetikzlibrary{shapes.geometric, arrows.meta, positioning, calc}

\TheoremsNumberedThrough     %
\ECRepeatTheorems

\EquationsNumberedThrough    %

\MANUSCRIPTNO{MS-0001-1922.65}

\usepackage{tcolorbox}
\tcbuselibrary{skins}
\usepackage{amsmath}
\newcommand{\norm}[1]{\left\lVert#1\right\rVert}
\usepackage{amssymb}
\usepackage{algorithm}
\usepackage{cleveref}
\usepackage[noend]{algpseudocode}
\usepackage{mathtools}

\usepackage{marvosym}
\usepackage{dsfont}
\usepackage[nopar]{lipsum}
\usepackage{hyperref}
\usepackage{enumitem}
\usepackage{graphicx}
\usepackage{subcaption}
\usepackage{multirow}
\usepackage{soul}
\usepackage{bbm}
\usepackage{booktabs}

\newcommand{\eg}{\textit{e}.\textit{g}., }

\providecommand{\cA}{\mathcal{A}}

\providecommand{\cX}{\mathcal{X}}

\providecommand{\cI}{\mathcal{I}}

\providecommand{\Regret}{\textnormal{\textsf{Regret}}}
\providecommand{\regret}{\textnormal{\textsf{regret}}}

\providecommand{\checkx}{\check{x}}

\providecommand{\barx}{\bar{x}}
\usepackage{xcolor}
\providecommand{\UCB}{\textsf{UCB}}
\providecommand{\LCB}{\textsf{LCB}}
\providecommand{\CI}{\textnormal{\textsf{CI}}}

\usepackage{etoolbox}
\AtBeginDocument{%
  \renewcommand{\eqref}[1]{(\ref{#1})}%
}
\def\QED{\hfill \quad{\bf Q.E.D.}\medskip}

\usepackage{amsmath,amsfonts,bm}

\newcommand*{\rom}[1]{%
\textup{\uppercase\expandafter{\romannumeral#1}}%
}

\def\eqref#1{equation~\ref{#1}}

\def\1{\bm{1}}

\DeclareMathAlphabet{\mathsfit}{\encodingdefault}{\sfdefault}{m}{sl}
\SetMathAlphabet{\mathsfit}{bold}{\encodingdefault}{\sfdefault}{bx}{n}

\DeclareMathOperator{\sign}{sign}

\newcommand{\bE}{\mathbb{E}}

\newcommand{\bR}{\mathbb{R}}

\newcommand{\inner}[2]{\left\langle #1, #2 \right\rangle}

\begin{document}

\RUNAUTHOR{}
\RUNTITLE{}
\TITLE{\Large LIBRA: Language Model Informed Bandit Recourse Algorithm for Personalized Treatment Planning}

\ARTICLEAUTHORS{%
\AUTHOR{Junyu Cao*$^{1}$, Ruijiang Gao*$^{2}$, Esmaeil Keyvanshokooh*$^{3}$, Jianhao Ma\footnote{Alphabetical Order; all authors have equal contributions.}$^{4}$}
\AFF{$^1$McCombs School of Business, University of Texas at Austin, \EMAIL{junyu.cao@mccombs.utexas.edu}} 
\AFF{$^2$Naveen Jindal School of Management, University of Texas at Dallas, \EMAIL{ruijiang.gao@utdallas.edu}} 
\AFF{$^3$Mays Business School, Texas A\&M University,
\EMAIL{keyvan@tamu.edu}}
\AFF{$^4$ Wharton School, University of Pennsylvania, \EMAIL{jianhaom@wharton.upenn.edu}} 
}

\ABSTRACT{
We introduce a unified framework that seamlessly integrates algorithmic recourse, contextual bandits, and large language models (LLMs) to support sequential decision-making in high-stakes settings such as personalized medicine. We first introduce the recourse bandit problem, where a decision-maker must select both a treatment action and a feasible, minimal modification to mutable patient features. To address this problem, we develop the Generalized Linear Recourse Bandit (\textsf{GLRB}) algorithm.
Building on this foundation, we propose \textsf{LIBRA}, a Language Model–Informed Bandit Recourse Algorithm that strategically combines domain knowledge from LLMs with the statistical rigor of bandit learning. \textsf{LIBRA} offers three key guarantees: (i) a {\em warm-start guarantee}, showing that \textsf{LIBRA} significantly reduces initial regret when LLM recommendations are near-optimal; (ii) an {\em LLM-effort guarantee}, proving that the algorithm consults the LLM only $O(\log^2 T)$ times, where $T$ is the time horizon, ensuring long-term autonomy; and (iii) a {\em robustness guarantee}, showing that \textsf{LIBRA} never performs worse than a pure bandit algorithm even when the LLM is unreliable. We further establish {\em matching lower bounds} that characterize the fundamental difficulty of the recourse bandit problem and demonstrate the near-optimality of our algorithms. Experiments on synthetic environments and a real hypertension-management case study confirm that \textsf{GLRB} and \textsf{LIBRA} improve regret, treatment quality, and sample efficiency compared with standard contextual bandits and LLM-only benchmarks. Our results highlight the promise of recourse-aware, LLM-assisted bandit algorithms for trustworthy LLM-bandits collaboration in personalized high-stakes decision-making. 
}


\maketitle

\vspace{-0.7cm}

\section{Introduction}

In light of recent advancements in medical interventions and the growing availability of electronic health record data, there is a pressing demand for data-driven algorithmic solutions in medical decision-making to supplement clinician judgment when selecting and adapting treatments over time \citep{t2023frontiers}. 
Clinicians are tasked with navigating highly uncertain and evolving patient responses, managing multiple comorbidities, and selecting among diverse treatment options \citep{cao2025safe}. To handle this task, clinicians mostly rely on adopting ad-hoc treatment strategies, which are often suboptimal in the face of such high-stakes, uncertain settings \citep{tunc2014opportunities}. This motivates framing treatment selection as a ``{\em sequential decision-making problem under uncertainty}," where approaches such as contextual bandits and reinforcement learning (RL) algorithms that can adaptively learn from patient outcomes while balancing the need to explore new care strategies and exploit known effective ones. These methods provide a rigorous foundation for personalized medicine, ensuring that treatment strategies evolve in tandem with patient health trajectories.

On the other hand, focusing solely on the selection of medications or therapies misses a critical dimension of care. In real-world clinical settings, clinicians rarely recommend treatments in isolation; specifically, they should also provide patients with what we shall call ``{\em recourses}." We define them as actionable lifestyle adjustments or feature modifications that could improve a patient’s eligibility for, or responsiveness to, more effective treatments. For example, in managing type 2 diabetes, beyond prescribing insulin or oral medications, physicians emphasize the importance of diet control and physical activities to improve treatment effectiveness. Existing bandit and RL algorithms, however, are typically blind to these opportunities and purely optimize over potential treatment options but not over feasible patient-side changes. Accordingly, we need to bridge this important gap by designing new algorithms that are able to simultaneously recommend treatments {\em and} suggest minimal, actionable modifications to mutable patient features within their clinically plausible ranges, thereby aligning algorithmic decisions more closely with clinical practices. 

At the same time, the recent rapid progress of {\em Generative Artificial Intelligence} (GenAI) tools, in particular large language models (LLMs), has opened a new path forward. In fact, these models can do more than assist clinicians; they can act as decision-makers themselves, offering treatment and lifestyle recommendations in real time. What sets them apart is their ability to integrate various sources of knowledge, including existing clinical guidelines, research evidence, and patient records, and turn them into practical suggestions that resemble the way clinicians practice.
Beyond acting as decision-maker agents, LLMs can also serve as {\em knowledge-infused priors} for bandit and RL algorithms, feeding them strong initial guesses about what might work best. This reduces the need for costly trial-and-error at the start of learning and accelerates progress toward finding effective, personalized care. When leveraged carefully, LLM-informed decision-making has the potential to make the treatment–recourse pipeline not only faster and more efficient, but also more human-centered and relevant to everyday clinical care.

Yet, LLMs are not without limitations. They are expensive to run at scale, demanding both computational power and financial resources that limit their accessibility in everyday clinical practice. They are also prone to {\em hallucinations}, producing recommendations that sound convincing but are clinically unfounded or even unsafe. More subtly, their outputs can be inconsistent, opaque, and difficult to audit, making it hard for clinicians to fully trust them in high-stakes medical decisions where accountability and safety are paramount. These shortcomings highlight why relying on LLMs alone is risky, particularly when patient outcomes are on the line. By contrast, contextual bandit and RL algorithms bring what LLMs lack: theoretical guarantees, statistical robustness, and accountability grounded in real patient data. They adapt systematically over time, learning from outcomes to ensure that performance improves rather than drifts unpredictably. Rather than treating LLM and bandits as competing approaches, the more compelling approach is one of integration. 

In light of the above discussions, our main research question that we address in this paper is {\em how one can develop a rigorous theoretical framework that offers the best of both worlds: the creativity and knowledge base of LLMs, coupled with the reliability and statistical guarantees of online learning algorithms.}

\subsection{Main Contributions and Results}

In this work, we propose a new, rigorous theoretical framework that bridges the gap between algorithmic recourse, online learning, and LLMs. Our main contributions are summarized as follows:

\begin{enumerate}
    \item \textbf{Recourse-Aware Bandit Formulation and Algorithm.}  
    We introduce the \emph{recourse bandit} problem, in which the learner must jointly recommend an action (e.g., a treatment) along with a minimal, feasible adjustment to mutable features (e.g., lifestyle changes) to improve outcomes. To address this new setting, we propose the \emph{Generalized Linear Recourse Bandit} (\textsf{GLRB}) algorithm, and then show that it achieves a provable recourse regret bound of $\widetilde{O}(d\sqrt{KT})$, where $d$ is the feature dimension, $K$ is the number of actions, and $T$ is the time horizon.

    \item \textbf{LIBRA: A Language Model-Informed Bandit Algorithm.}  
    We develop a collaborative LLM-Bandit framework that we called the \emph{LIBRA} (Language Model Informed Bandit Recourse Algorithm), which integrates LLMs with online bandit learning theory. \textsf{LIBRA} leverages the domain knowledge and fluency of LLMs to warm-start the learning process, while relying on the statistical rigor and adaptivity of bandits to ensure reliable long-term performance. This  unified design enables efficient and trustworthy sequential decision-making in high-stakes settings, such as personalized medicine.

    \item \textbf{Theoretical Performance Guarantees.}  
    We establish rigorous theoretical performance guarantees for \textsf{LIBRA}, which formalize the benefits and safety of integrating LLMs with online bandit learning:
    \begin{itemize}
        \item \emph{Warm-start guarantee:} Whenever the LLM provides $\eta$-suboptimal recommendations (i.e., close to optimal within a small additive gap), \textsf{LIBRA} achieves significantly lower recourse regret than pure bandit methods in the early stages of learning, effectively mitigating the cold-start problem.
        \item \emph{LLM-effort guarantee:} \textsf{LIBRA} provably invokes the LLM only $O(\log^2 T)$ times across $T$ rounds. This ensures that the system becomes increasingly autonomous over time, reducing reliance on costly or latency-prone LLM queries.
        \item \emph{Robustness guarantee:} Even when the LLM is unreliable (i.e., when $\eta$ is large), \textsf{LIBRA} automatically reverts to its bandit backbone and maintains the same sublinear recourse regret bound as \textsf{GLRB}. Thus, \textsf{LIBRA} is always at least as good as a pure bandit approach.
        \item \emph{Near-optimality guarantee:} We derive a fundamental lower bound for the recourse bandit problem, showing that the recourse regret is at least $\Omega\big(\gamma d_M \sqrt{KT} \vee \sqrt{dKT}\big)$, where $d_M$ is the number of mutable features and $\gamma$ is the maximum recourse radius. In the presence of an $\eta$-suboptimal oracle (e.g., an LLM), the recourse regret is further bounded below by $\Omega\big( \eta T \wedge \big(\gamma d_M \sqrt{KT} \vee \sqrt{dKT} \big) \big)$. Our algorithms {\em match} these lower bounds up to logarithmic factors under mild conditions, establishing the theoretical {\em tightness} and {\em robustness} of our approach across regimes.
    \end{itemize}

    \item \textbf{Empirical Evaluation on Both Synthetic and Clinical Data.}  
    We validate our algorithms on both synthetic environments and a real-world case study on hypertension management using the ACCORD dataset. This dataset reflects a realistic clinical setting where treatment effectiveness depends on both pharmacological interventions and lifestyle factors, such as diet and exercise. Our results show that both \textsf{GLRB} and \textsf{LIBRA} outperform standard linear contextual bandits (LinUCB) and LLM-only baselines in terms of regret, treatment quality, and sample efficiency. In particular, LIBRA achieves rapid improvement in early rounds due to LLM guidance and seamlessly transitions to autonomous learning. These experiments highlight the practical relevance of recourse-aware bandits and their potential for LLM-Bandit integration in high-stakes applications.
\end{enumerate}

\begin{figure*}[htp]
  \centering
    \includegraphics[width=\textwidth]{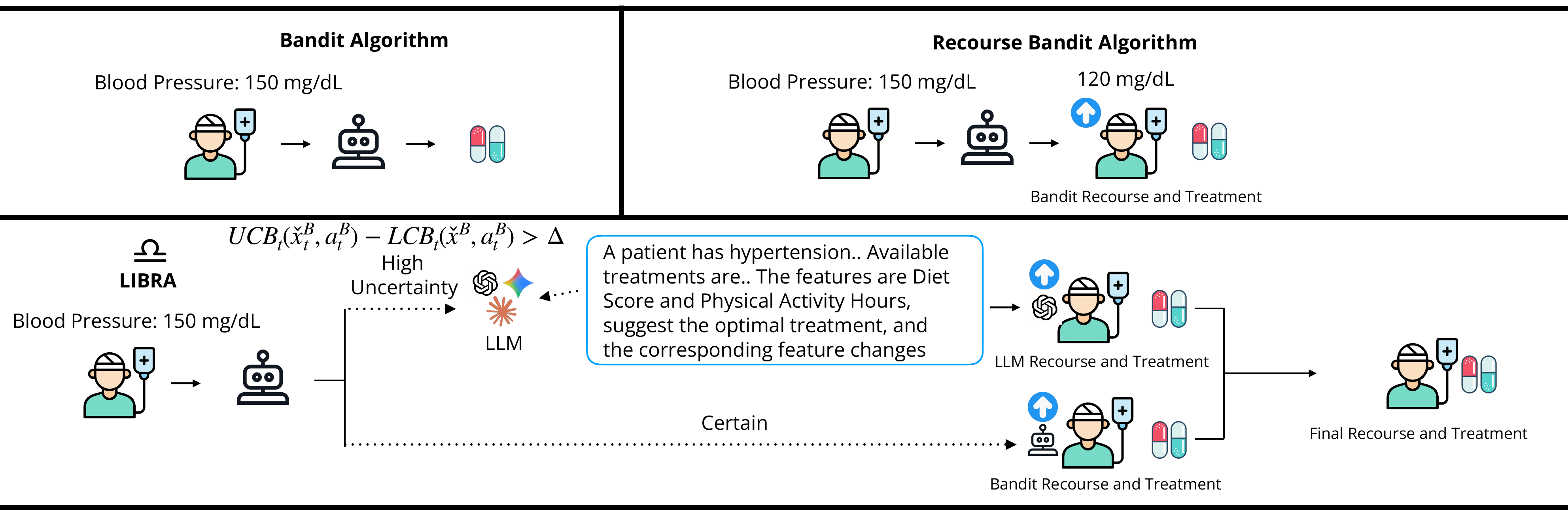}
    \caption{Illustration of \textsf{Recourse Bandit} (GLRB) and \textsf{LIBRA} frameworks. GLRB offers algorithmic recourses to patients to improve their health conditions for more effective treatment (see \S \ref{sec: GLRB} for the details of the GLRB algorithm).  \textsf{LIBRA} selectively consults LLMs based on uncertainty estimates, enabling a data-driven integration of LLM knowledge and AI recommendations (see \S \ref{sec: LIBRA} for the details of the LIBRA algorithm and their theoretical properties).~\looseness=-1}
    \label{fig:figure1}
\end{figure*}

A detailed illustration of \textsf{GLRB} and \textsf{LIBRA} is provided in \Cref{fig:figure1}.

\subsection{Literature Review}

Our work is related to the following three streams of literature. 

\noindent\textbf{Algorithmic Recourse.} 
Algorithmic Recourse seeks to help people who have been negatively impacted by algorithmic decisions by offering alternative paths to positive outcomes~\citep{wachter2017counterfactual,ustun2019actionable,van2019interpretable,pawelczyk2020learning,mahajan2019preserving,karimi2019model,karimi2020probabilistic,dandl2020multi,gao2023impact,liao2025constraint, liao2025rolemodel}.  These methods can be classified in several ways~\citep{verma2020counterfactual}, including the type of predictive model used, the level of access to the model, the preference for changing only a few features in the suggested alternatives (sparsity), whether the alternatives should resemble real data, the consideration of causal relationships, and whether the method produces one or multiple alternatives.  Research has also shown that current recourse methods may lack robustness, as small changes to the initial input~\citep{dominguezolmedo2021adversarial} or the underlying model~\citep{upadhyay2021robust,rawal2021modelshifts} can affect the suggested recourse.  Most prior research in algorithmic recourse focuses on offline classification.  This work introduces the first algorithmic recourse approach within a contextual bandit framework, combining recourse with online learning to potentially improve the performance of online algorithms by providing counterfactual explanations.

\noindent\textbf{Contextual Bandits.} Contextual Bandits have been surveyed by \citet{slivkins2019introduction, bubeck2012regret, lattimore2020bandit}. There is a growing body of literature on developing various contextual multi-armed bandit (MAB) algorithms for different settings. Many of these algorithms are based on either upper confidence bound (UCB) methods (\eg \citealt{auer2002using,dani2008stochastic, chu2011contextual, abbasi2011improved, li2017provably,gan2024contextual,wang2024smart}) or Thompson sampling (\eg \citealt{agrawal2013thompson, keyvanshokooh2025contextual, russo2014learning, zhalechian2022online,abeille2017linear,zhalechian2023data,zhou2023spoiled,keyvanshokooh2025learning}). A notable contribution by \citet{harris2022strategy} considers a scenario where human decision-makers act adversarially to a known public policy. They prove that in this setting, deriving a sublinear strategic regret is {\em generally impossible}.

In contrast to these studies, our work focuses on the new setting of \textit{recourse bandit}, where human subjects attempt to improve their conditions (features) by following the algorithm's recommendations. In this setting, we provide a theoretical analysis proving that our algorithm achieves sublinear \textit{recourse regret}. Our approach is closely related to hint-based bandits \citep{cutkosky2022leveraging, bhaskara2023bandit}, though we differ in that, unlike previous methods which rely on a single hint at the start, the recourse bandit setting involves receiving a different context at each time step. This requires the system to decide whether to consult LLM experts for an alternative course of action.
Additionally, our work shares similarities with conversational bandits \citep{zhang2020conversational, zuo2022hierarchical}, which enhance learning in recommender systems by querying user preferences on a subset of arms. However, unlike conversational bandits, where the decision to ask for additional feedback does not involve comparing UCB and lower confidence bounds (LCB) of human and AI decisions, our LLM-Bandit system dynamically decides whether to consult a LLM, depending on the bandit algorithm's confidence in a given pair of actions and recourse.

\noindent\textbf{LLMs in Online Learning Problems.} The use of external knowledge to enhance reinforcement learning (RL) algorithms has been explored primarily in offline settings. Recent works investigate human-AI collaboration in offline RL, with \citet{grand2024best} using RL-generated actions to guide human decision-making. Similarly, \citet{gao2021human} and \citet{gao2023confounding} propose a learning-to-defer framework with bandit feedback to allocate tasks between humans and AI. Meanwhile, \citet{bordt2022bandit} and \citet{wang2022blessing} explore how private human information can be integrated into RL and bandit algorithms to improve performance. Moreover, \cite{wang2024large} focuses on an offline setting by studying how to integrate LLM-generated data with human data to improve statistical estimation.

In contrast, our work focuses on improving online bandit algorithms by incorporating LLM's knowledge. We demonstrate that LLMs can enhance bandit algorithms even when given the same information. Additionally, recent studies have examined the potential of leveraging LLMs for online bandit problems. \citet{alamdari2024jump} propose using an LLM to generate synthetic data that helps warm-start bandit algorithms, but their approach does not address algorithmic recourses and provides only empirical performance improvements without theoretical analysis. \citet{ye2025lola} leverage LLMs to improve reward predictions in natural language for features in an online headline selection task. Meanwhile, \citet{park2024llm} show that state-of-the-art LLMs, such as GPT-4, fail to exhibit no-regret behavior in online learning environments. They propose a regret-loss mechanism to encourage no-regret performance from LLMs.
In contrast to these approaches, our method, \textsf{LIBRA}, elegantly combines a pre-trained LLM with online learning algorithms, improving performance while ensuring sublinear regret.

\section{Generalized Linear Recourse Bandit Algorithm}\label{sec: GLRB}
In this section, we introduce the new recourse optimization problem, develop the generalized linear recourse bandit (\textsf{GLRB}) algorithm for this model, and establish its recourse regret analysis. 

\subsection{Recourse Optimization Problem}\label{sec: RO}
We consider the following contextual decision-making problem in which a decision is made after observing patient-specific information. We denote patient information (context) as $x = (x_{I}, x_{M}) \in \cX\subset \mathbb{R}^{d}$ divided into two parts. The first $x_{I}\in \mathbb{R}^{d_I}$ represents the {\em immutable} features, such as age, race, and sex; characteristics that cannot be changed. The second  $x_{M}\in \mathbb{R}^{d_M}$ presents the {\em mutable} features, such as blood sugar level and alcohol consumption; attributes that could potentially be modified through recourse or intervention.

{\bf Generalized Linear Model (GLM)}: Let $\cA$ be the set of actions/treatments with $|\mathcal{A}|=K$. For instance, in a clinical setting, each action could represent a distinct treatment plan available to a physician. Given the observed context $x_t = (x_{t, I}, x_{t, M})$ at time $t$, we model the expected outcome (or reward) of selecting an action/arm $a\in \mathcal{A}$ using a GLM model. Specifically, for each action $a\in \mathcal{A}$, there is an unknown parameter vector $\theta_a^\star = (\theta_{a,I}^\star, \theta_{a,M}^\star) \in \mathbb{R}^d$ with $d=d_I+d_M$. The reward observed at time $t$ after selecting action $a_t$ is:
\begin{align}
r_t(x_t,a_t): 
= \mu\big({\theta_{a_t}^\star}^\top x_t\big) + \xi_t
=  \mu\big(x_{t, I}^\top  \theta_{a_t,I}^\star +  x_{t, M}^\top  \theta_{a_t,M}^\star\big) + \xi_t,
\tag{\textsf{GLM}}\label{eq: GLM}
\end{align}
where $\mu: \mathbb{R} \to \mathbb{R}$ is a known, strictly increasing link function that is $L_\mu$-Lipschitz continuous and satisfies $\inf_x \mu'(x) = c_\mu > 0$. This is a standard assumption in the literature; see e.g., Assumption 1 in \citet{filippi2010parametric}. Classical linear and logistic models fall under this framework as special cases, with $\mu(z) = z$ and $\mu(z) = 1/(1+\exp(-z))$, respectively. We assume $\xi_t$ is zero-mean, $\sigma$-sub-Gaussian noise (i.e., conditionally sub-Gaussian given the history). We also assume the true parameters are bounded: $\sup_{a \in \mathcal{A}} \|\theta_a\|_2 \leq \beta_\Theta$.

{\bf Recourse Optimization  (RO) Problem}: While the immutable context $x_{I}$ cannot be changed, the mutable context $x_M$ can be adjusted within clinically or practically plausible limits. Specifically, we allow $x_M$ to be  modified to a new value $\check{x}_M$ provided that the adjustment is controlled by $\norm{\checkx_M- x_M} \leq \gamma$, where $\norm{\cdot}$ denotes a chosen distance metric and $\gamma > 0$ controls the maximum allowable magnitude of the adjustment. For example, depending on the application, this distance could be measured using different norms, including the Euclidean norm $\norm{\cdot}_2$ or the infinity norm $\norm{\cdot}_{\infty}$. 
We refer to the adjusted feature vector $\check{x}_M$ as a ``{\em recourse}" that can be seen as an {\em actionable} modification of the patient's mutable features. Throughout the paper, we denote $\beta_{\cX}:=\max_{\norm{\checkx-x}\leq \gamma, x\in \cX}\norm{\checkx}_2<\infty$ the maximum $\ell_2$-norm of all feasible contexts. Given the observed context $x = (x_{I}, x_{M})$ and the true model parameters $\{\theta_{a}^\star\}_{a\in \mathcal{A}}$, the decision-maker seeks both the best action {\em and} the most effective recourse by solving the {\em recourse optimization} (\textsf{RO}) problem:
\begin{align}
    \max_{a\in \cA}\max_{\checkx_M\in \mathbb{R}^{d_M}} & 
    \mu\left((x_I,\checkx_M)^\top \theta_a^\star \right)
    \tag{\textsf{RO}}
    \label{eq: RO}\\
    \text{subject to}\quad &  \norm{\checkx_M- x_M} \leq \gamma. \notag
\end{align}

The goal is to maximize the expected reward by finding the best combination of action and recourse.

\begin{example}[Managing Type 2 Diabetes]
    Our formulation \eqref{eq: RO} is fairly general, but to make it more concrete, consider the case of managing type 2 diabetes. Clinicians must prescribe a medication (action), e.g., insulin or an oral drug like metformin that directly lowers blood glucose. In practice, however, pharmacological treatment alone is rarely sufficient. Clinicians may also need to ask patients for adopting a healthier diet, exercising regularly, or losing weight (recourses) to bring their BMI into a safer range. The critical point is that these lifestyle adjustments must be realistic and actionable for patients. For instance, asking a sedentary patient to immediately start running five miles every day is not actionable, but encouraging them to begin with a 20-minute walk after meals is both feasible and clinically meaningful. In the same way, reducing daily sugar intake by a small but measurable amount can move a patient's blood glucose level closer to a range where the prescribed medication is far more effective. This example highlights the central intuition of \eqref{eq: RO}: it does not only recommend what medication to give, but also what actionable changes the patient should make so that the treatment can achieve its maximum effect. By explicitly modeling these coupled decisions, \eqref{eq: RO} aligns far more closely with how clinicians design treatment plans in practice.
\end{example}

The following lemma characterizes the closed-form solution to our recourse optimization problem \eqref{eq: RO} under the full-information setting, where model parameters are known (offline setting).
\begin{lemma}[Optimal Recourse Under Full Information]\label{lem: closed-form-solution}
    For any given treatment/action $a \in \mathcal{A}$, the optimal recourse derived by solving the recourse optimization problem \eqref{eq: RO} is $\checkx_M^\star=x_M+\gamma\cdot \partial \norm{\theta^\star_{a, M}}_\star$, where $\partial f$ is a subgradient of the function $f$, and $\norm{\cdot}_\star$ is the dual norm of $\norm{\cdot}$.\footnote{Formally, the subgradient may not be unique, and each corresponds to a valid optimal recourse. Accordingly, for simplicity, we select one subgradient arbitrarily.}
\end{lemma}

This technical result admits an intuitive interpretation. The optimal recourse adjusts the mutable features in the direction that maximally increases the expected reward, subject to the allowed perturbation budget. Specifically, 
the optimal recourse direction aligns with a subgradient of the dual norm of the mutable-feature parameter vector $\theta^\star_{a, M}$. 
Below, we further illustrate the structure of this solution with two commonly used distance metrics, highlighting how the best recourse varies across different norms.
\begin{corollary}
\label{cor::examples}
    The closed-form solution in Lemma~\ref{lem: closed-form-solution} specializes to the following common norms:
\begin{enumerate}
    \item \textbf{$\ell_2$-norm}: Let $\norm{x-x'} = \norm{x-x'}_2$ be the distance function, then the optimal recourse for each given arm $a \in \mathcal{A}$ derived by \eqref{eq: RO} is $\checkx_M^\star=x_M+\frac{\theta_{a, M}^\star}{\norm{\theta_{a, M}^\star}_2}\gamma$ if $\theta_{a, M}^\star\neq 0$.
    \item \textbf{Weighted $\ell_{\infty}$-norm}: Let $\norm{x-x'} = \max_{1\leq i\leq d_M} \alpha_i|x(i)-x'(i)|$ be the distance function with weights $\alpha_i\geq 0$, then the optimal recourse for each given arm $a \in \mathcal{A}$ derived by \eqref{eq: RO} is $\checkx_M^\star(i)= x_M(i)+\frac{\gamma}{\alpha_i}\sign\left(\theta_{a, M}^\star(i)\right)$ for all $1\leq i\leq d_M$.
\end{enumerate} 
\end{corollary}

\subsection{Generalized Linear Recourse Bandit Algorithm (\textsf{GLRB})}

In Section~\ref{sec: RO}, we formulated the recourse optimization problem \eqref{eq: RO} under the offline setting in which the true model parameters are known. In practice, however, the parameters $\theta_{a}^\star$ associated with each $a \in \mathcal{A}$ are unknown and must be learned sequentially as new data are collected. To address this challenge, we propose the Generalized Linear Recourse Bandit (\textsf{GLRB}) algorithm (Algorithm~\ref{alg:GLRB}) for this online setting. This algorithm simultaneously learns the unknown model parameters while recommending both a treatment and an actionable recourse for the patient's mutable  features.

At a high level, the \textsf{GLRB} algorithm is built around the central exploration–exploitation trade-off. At each round $t$, the algorithm observes a patient context $x_t = (x_{t, I}, x_{t, M})$ divided into immutable features (e.g., age or gender) and mutable ones (e.g., lifestyle measures). For each potential action $a \in \mathcal{A}$, it then solves the {\em optimistic recourse optimization} problem \eqref{eq::ORO-Arm}. This problem optimizes for the best combination of recourse adjustments and model parameters within the current uncertainty set $\Theta_{t, a}$. Optimizing over such an uncertainty set encourages {\em exploration}, since underexplored actions are assessed not only by their point estimates but also by plausible alternatives that remain statistically consistent with the data. The algorithm selects and implements the action–recourse pair with the highest optimistic value. The observed feedback outcome $r_t$ is then used to refine $\Theta_{t, a}$. Over time, as additional feedback data shrinks the uncertainty set, the algorithm gradually shifts toward {\em exploitation}, favoring actions and recourses it has high confidence in. 

\begin{algorithm}[htp]
\caption{Generalized Linear Recourse Bandit (\textsf{GLRB}) Algorithm}\label{alg:GLRB}
\begin{algorithmic}%
\State \textbf{Input}: Time horizon $T$ and control parameter $\gamma$. 
\For{$t=1,\cdots,T$}
\State Observe the context $x_t = (x_{t, I}, x_{t, M})$.
\State Solve the optimistic recourse optimization problem \eqref{eq::ORO-Arm} for each arm $a \in \mathcal{A}$ by Algorithm~\ref{alg: two-blockCD}:
\begin{align}
   \max_{\checkx_{t, M}, \; \theta_a =  (\theta_{a, M}, \; \theta_{a, I})} \quad &\checkx_{t, M}^\top \theta_{a, M} + x_{t, I}^\top \theta_{a, I}, 
   \tag{\textsf{ORO-Arm}} 
   \label{eq::ORO-Arm} \\
    \text{subject to}\hspace{0.2cm}\quad &\hspace{-0.05cm}\norm{\checkx_{t, M} - x_{t, M}} \leq \gamma,\notag \\
    &\big\|\theta_a - \hat\theta_{t, a} \big\|_{V_{t, a}}\leq \rho_{t, a}.\notag
\end{align}
\State Implement the recourse $\checkx_{t, M}$ that maximizes \eqref{eq::ORO-Arm} and the corresponding action $a_t$.
\State Observe the feedback outcome $r_t$ and update the uncertainty set $\Theta_{t, a}$ (see Lemma~\ref{lemma: glm-radius}).
\EndFor
\end{algorithmic}
\end{algorithm}
Below, we present the details of the two main components of the \textsf{GLRB} algorithm.

(i) {\bf Data-driven Uncertainty Set}: Following the bandit literature, we first construct a high-probability uncertainty set for the unknown parameters $\theta_a^\star$ for each arm $a \in \mathcal{A}$ in our \eqref{eq: GLM} model. We denote $\mathcal{I}_{t, a} = \{ s < t \;|\; a_s = a \}$ as the set of time steps before $t$ where arm $a$ was selected, and let $n_{t, a} = |\mathcal{I}_{t, a}|$.  Given the historical observations $\{(r_s, x_s, a_s)\}_{s=1}^{t-1}$, we estimate the unknown model parameter $\theta_{a}^\star$ by minimizing the following regularized negative log-likelihood loss function for $\lambda > 0$:
\begin{align*}
    \widehat{\theta}_{t, a} := 
    \argmin_{\theta_a =(\theta_{a,M}, \theta_{a,I})\in \mathbb{R}^d} 
    \left\{
    \sum_{s \in \mathcal{I}_{t, a}}
    \ell(r_s, x_s^\top \theta_a) + \frac{\lambda}{2} \; \norm{\theta_a}_2^2
    \right\},
\end{align*}
where $\ell(r, z): = rz + m(z)$ 
with $m'(z) = \mu(z)$ 
is the negative log-likelihood loss for a single observation. Lemma~\ref{lemma: glm-radius} characterizes a high-probability uncertainty set $\Theta_{t, a}$ for the parameter $\theta_a^\star$. 

\begin{lemma}[High Probability Uncertainty Set]\label{lemma: glm-radius}
    Let $\delta\in (0,1)$. With probability at least $1-\delta$, for all time steps $t\in [T]$ and any arm $a \in \mathcal{A}$, it holds: 
    \begin{align*}
        \theta_a^\star \in \Theta_{t, a}:=
        \left\{
        \theta_a:\big\|\theta_a-\widehat{\theta}_{t, a}\big\|_{V_{t, a}}\leq \rho_{t, a}
        \right\},
    \end{align*}
where $V_{t, a}:= \lambda I + \sum_{s \in \mathcal{I}_{t, a}} x_s   x_s^\top$ is the design matrix, $\mathcal{I}_{t, a} = \{ s < t \;|\; a_s = a \}$ is the set of time steps before $t$ where arm $a \in \mathcal{A}$ was selected, $n_{t, a} = |\mathcal{I}_{t, a}|$, and the confidence radius $\rho_{t, a}$ is defined as:
$$
\rho_{t, a} = \frac{1}{c_{\mu}} 
\left(
 \sigma \sqrt{d \log \left( 1 + \frac{\beta_{\mathcal{X}}^2 n_{t, a}}{\lambda} \right) + d\log \left( \frac{K}{\delta} \right)} +
\sqrt{\lambda} \beta_{\Theta}
\right).
$$
\end{lemma}
This uncertainty set captures all statistically plausible parameter values consistent with the observed data and serves as the foundation for optimistic decision-making in the \textsf{GLRB} algorithm.

\smallskip
(ii) {\bf Optimistic Recourse Optimization (\textsf{ORO-Arm}) Problem}: The second component of Algorithm~\ref{alg:GLRB} builds on the optimism in the face of uncertainty principle that finds the best model with the highest expected reward under the data gathered thus far. Specifically, having constructed the uncertainty set $\Theta_{t, a}$ in Lemma~\ref{lemma: glm-radius}, we first formulate an {\em optimistic} version of the recourse optimization problem  \eqref{eq: RO} as:
\begin{align}
\label{eq::ORO}
   \max_{a\in \mathcal{A}} \max_{\checkx_M\in \mathbb{R}^{d_M}, \; \theta_a\in \Theta_{t, a}} &\quad \mu\left(\checkx_M^\top \theta_{a,M} +\checkx_I^\top \theta_{a,I}\right)\tag{\textsf{ORO}}\\
    \quad \text{subject to}\hspace{1cm} &  \quad \norm{\checkx_M- x_M} \leq \gamma. \notag
    \end{align}
The \eqref{eq::ORO} problem maximizes the expected reward when $\checkx_M$ is within $\gamma$-distance from the initial mutable context $x_M$ and $\theta_a$ is contained in the uncertainty set $\Theta_{t, a}$. When the true parameter $\theta_a^\star$ is contained in the uncertainty set, the optimal solution of \eqref{eq::ORO} is an {\em upper bound} for the offline recourse problem \eqref{eq: RO}. 

To solve \eqref{eq::ORO}, it suffices to decompose it across actions and solve a per-action subproblem \eqref{eq::ORO-Arm} for each arm $a \in \mathcal{A}$, which is the optimization model that we need to solve in Algorithm~\ref{alg:GLRB}. Specifically, it suffices to compute the best recourse for each arm individually and then choose the arm that yields the highest expected reward. Moreover, since the link function $\mu(\cdot)$ is strictly increasing, we can drop it from the optimization and instead directly maximize the inner product $(x_I,\checkx_M)^\top \theta_a$. The resulting \eqref{eq::ORO-Arm} is a {\em constrained nonlinear} optimization problem with a quadratic objective function and two norm-based constraints. In general, the objective function is nonconcave, rendering the problem NP-hard and limiting us to local solutions under arbitrary norms. However, the following result shows that under Mahalanobis norms---i.e., norms of the form $\norm{x} = \sqrt{x^\top A x}$ for a positive semidefinite (PSD) matrix $A$, which include the $\ell_2$-norm---the problem can be solved globally in polynomial time.
\begin{proposition}[Global Optimality of \eqref{eq::ORO-Arm}]
    \label{thm::global-optimality}
    When the Mahalanobis norm is used in \eqref{eq::ORO-Arm}, this optimistic recourse problem can be solved to global optimality in polynomial time.
\end{proposition}

While the optimistic recourse problem \eqref{eq::ORO-Arm} is solvable in polynomial time under Mahalanobis norms, existing methods often rely on either intricate semidefinite programming (SDP) reformulations or elaborate combinatorial techniques (see e.g., \citealt{yang2016two,bienstock2016note}). These methods can become computationally infeasible in high-dimensional applications, including healthcare settings with rich patient feature spaces. To overcome this challenge and accommodate general distance function metrics, we develop a {\em two-block coordinate descent} algorithm (\Cref{alg: two-blockCD}) that works under more general norms. This algorithm alternates between two key update steps. At each iteration $k$, it first fixes the current recommended recourse $\checkx_M^{(k)}$ and update the model parameter $\theta_a$ to obtain $\theta_a^{(k+1)}$; it then fixes $\theta_a^{(k+1)}$ and solve for the optimal recourse $\checkx_M^{(k+1)}$. Under general norms, the objective function of the optimistic recourse problem \eqref{eq::ORO-Arm} remains {\em biconvex}, and importantly, each subproblem admits a \textit{closed-form} solution. The details of our two-block coordinate descent approach are described in \Cref{alg: two-blockCD}. 
\begin{algorithm}
\caption{Two-Block Coordinate Descent Algorithm for Solving \eqref{eq::ORO-Arm}}\label{alg: two-blockCD}
\begin{algorithmic}
\State \textbf{Initialization:}
$\checkx_M^{(0)}=x_M$.  
\State \textbf{Repeat for $k = 0, 1, ...$ as:}
\begin{align}
&  \theta_a^{(k+1)} = \argmax_{\norm{\theta_a - \hat\theta_a}_{V_a}\leq \rho} \checkx_M^{(k)\top} \theta_{a, M} + x_I^{\top} \theta_{a, I} = \hat{\theta}_a+\rho\cdot \partial \norm{\left[\checkx_M^{(k)}; x_I\right]}_{V_a^{-1}} 
\tag{\textrm{\emph{Optimize parameter $\theta_a$}}}\\
& \checkx_M^{(k+1)} = \argmax_{\norm{\checkx_M- x_M} \leq \gamma}  \checkx_M^{\top} \theta_{a, M}^{(k+1)}=x_M+\gamma\cdot \partial \norm{\theta_{a,M}^{(k+1)}}_{\star}
\tag{\textrm{\emph{Optimize recourse $\checkx_M$}}}
\end{align}
\end{algorithmic}
\end{algorithm}

We now proceed to establish theoretical performance guarantees for the proposed \Cref{alg: two-blockCD}. Our first result guarantees that this algorithm always converges to a critical point of \eqref{eq::ORO-Arm}.
\begin{lemma}[Adapted from Corollary~2 in \citealt{grippo2000convergence}]
\label{lem::converge-to-critical-points}
    The two-block coordinate descent algorithm (\Cref{alg: two-blockCD}) converges to a critical point of \eqref{eq::ORO-Arm}.
\end{lemma}

Although \Cref{alg: two-blockCD} may converge only to a critical point due to the nonconcave nature of the optimistic recourse problem \eqref{eq::ORO-Arm}, its structure offers a key desirable guarantee. Most notably, it {\em consistently improves} upon the non-recourse formulation. This is ensured by the initialization $\checkx_M^{(0)} = x_M$, which implies that after the first iteration, the solution is strictly better than the non-recourse baseline. Since the sequence of iterates $\big(\theta_a^{(k)}, \checkx_M^{(k)}\big)$ yields a monotonically non-decreasing objective value, the final solution returned by \Cref{alg: two-blockCD} is always better than the non-recourse version.

We next analyze the convergence rate of \Cref{alg: two-blockCD}. In the special case in which the recourse procedure uses the $\ell_1$-norm as the distance metric, \Cref{cor::examples} implies that the mutable context at each iteration satisfies $\checkx_M^{(k)} \in x_M + \gamma \{\pm 1\}^{d_M}$, which forms a finite set. Since the objective value is monotonically non-decreasing across iterations, the algorithm must therefore terminate in finitely many steps. For the general case, we establish convergence under the Kurdyka–\L{}ojasiewicz (KL) property \citep{kurdyka1998gradients}, which provides a broad framework for analyzing the convergence behavior of nonconvex optimization algorithms.

\smallskip
\begin{definition}[{\bf Kurdyka--\L{}ojasiewicz (KL) Property}]\label{def: KL}
    A proper and lower-semicontinuous function $f$ is said to satisfy the KL property if for every compact set $\Omega \subset \operatorname{dom} f$ on which the function $f$ takes a constant value $f_{\Omega} \in \mathbb{R}$, there exist constants $\varepsilon>0$ and $\lambda>0$ such that for all $\bar{x} \in \Omega$ and all points $x \in\left\{z \in \mathbb{R}^d: \min_{z'\in \Omega}\norm{z-z'}<\varepsilon, f_{\Omega}<\right.$ $\left.f(z)<f_{\Omega}+\lambda\right\}$, the following inequality holds:
    \begin{equation}
        \varphi^{\prime}\left(f(x)-f_{\Omega}\right) \cdot \norm{\partial f(x)} \; \geq \; 1,
    \end{equation}
where the function $\varphi:[0, \lambda) \rightarrow \mathbb{R}_{+}$ takes the form $\varphi(t)=\frac{c}{\alpha} \; t^\alpha$ for some constants $c>0, \alpha \in(0,1]$.
\end{definition}

The KL property provides a unifying framework for analyzing the convergence behavior of first-order and block-coordinate optimization algorithms in nonconvex and nonsmooth settings. In particular, when an objective function satisfies the KL property, one can characterize not only the convergence of the generated iterates to a critical point, but also obtain explicit convergence rates that depend on the KL exponent $\alpha$. Many objective functions arising in practice, including semialgebraic and definable functions, are known to satisfy the KL property \citep{attouch2013convergence}.
We leverage this property to establish convergence guarantees for the proposed algorithm. In what follows, we show that the optimistic recourse optimization problem satisfies the KL property and derive corresponding convergence rate results for \Cref{alg: two-blockCD}.

\begin{theorem}[Theoretical Convergence Rate of Algorithm~\ref{alg: two-blockCD}]
\label{thm::convergence-rate-KL}
    Suppose that the norm $\norm{\cdot}$ is semialgebraic. The optimistic recourse optimization problem \eqref{eq::ORO-Arm} satisfies the KL property with exponent $\alpha\in (0, 1]$. Furthermore, suppose that \Cref{alg: two-blockCD} produces a sequence $(\theta_a^{(k)}, \checkx_M^{(k)})$ that converges to a critical point $(\theta_{a, \mathrm{crit}}, \checkx_{M, \mathrm{crit}})$. Then, the following convergence guarantees hold:
    \begin{enumerate}
        \item If $\alpha = 1$, the sequence $(\theta_a^{(k)}, \checkx_M^{(k)})$ converges to $(\theta_{a, \mathrm{crit}}, \checkx_{M, \mathrm{crit}})$ in finitely many iterations.
        \item If $\alpha \in\left[\frac{1}{2}, 1\right]$, there exist constants $k_0>0, C>0$, and $\tau \in[0,1)$ such that:
        $$\big\|(\theta_a^{(k)}, \checkx_M^{(k)})-(\theta_{a, \mathrm{crit}}, \checkx_{M, \mathrm{crit}})\big\| \leq C \tau^k, \;\; \text{for all $k \geq k_0$}.
        $$ 
        \item If $\alpha \in\left(0, \frac{1}{2}\right)$, there exist constants $k_0>0$ and $C>0$ such that:
        $$\big\|(\theta_a^{(k)}, \checkx_M^{(k)})-(\theta_{a, \mathrm{crit}}, \checkx_{M, \mathrm{crit}})\big\| \leq C k^{-\alpha /(1-2 \alpha)}, \;\; \text{for all $k \geq k_0$}.$$ 
    \end{enumerate}
\end{theorem}

We should highlight the assumption that the norm $\norm{\cdot}$ is semialgebraic is mild. Semialgebraic functions form a rich and well-studied class that includes many commonly used functions \citep{drusvyatskiy2013semi}. In particular, for most standard norms---including $\ell_p$-norms with $p \geq 1$ and Mahalanobis norms---it is straightforward to verify that they are semialgebraic. On the other hand, we should note that while the KL property is guaranteed, the specific KL exponent $\alpha \in (0, 1]$ depends on the detailed structure of the problem and is not characterized here. Nevertheless, our empirical results show that \Cref{alg: two-blockCD} performs effectively in practice, often converging rapidly to high-quality solutions.

\subsection{Theoretical Performance Guarantees for the \textsf{GLRB} Algorithm}

In this section, we evaluate the theoretical performance guarantee of the \textsf{GLRB} algorithm (Algorithm~\ref{alg:GLRB}) using the notion of {\em recourse regret} formally defined below. 

\smallskip
\begin{definition}[{\bf Recourse Regret}]\label{def: recregret}
    The \emph{recourse regret} of a learning algorithm $\pi$ over a  horizon \( T \) is:
    \[
       \Regret_\pi(T) := \mathbb{E}_\pi\left[\sum_{t=1}^T  \  r_t(x_t^\star, a_t^\star) 
       - 
       r_t(x_t, a_t) \right],
    \]
    where $(x_{t}^\star, a_t^\star)$ is the optimal recourse and action pair at time $t$, and $(x_{t}, a_t)$ is the recourse and action pair at time $t$ selected by the learning algorithm. The expectation is taken over all sources of randomness, including both observational noise and the algorithm’s internal randomness.
\end{definition}

\begin{theorem}[\textbf{Recourse Regret of the \textsf{GLRB} Algorithm}]\label{thm: regretbound-GLRB}
Without loss of generality, assume that $\beta_{\Theta} \cdot \beta_{\mathcal{X}} \leq 1$. With probability at least $1-\delta$, the recourse regret of Algorithm~\ref{alg:GLRB} can be bounded as follows:
\begin{align*}
    \Regret_\pi(T) 
    \leq 2L_{\mu} \cdot \rho_{T}  \sqrt{2dKT \log\left(1 + \frac{\beta_{\mathcal{X}}^2 T}{\lambda d} \right)},
\end{align*}
where $\rho_{T} = \frac{1}{c_{\mu}} 
\left(
 \sigma \sqrt{d \log \left( 1 + \frac{\beta_{\mathcal{X}}^2 T}{\lambda} \right) + d\log \left( \frac{K}{\delta} \right)} +
\sqrt{\lambda} \beta_{\Theta}
\right)$. In particular, upon setting the regularization coefficient $\lambda=\Theta(\sigma^2d/\beta^2_{\Theta})$, we have:
\begin{align*}
    \Regret_\pi(T) 
    \lesssim \frac{\sigma L_{\mu}}{c_\mu} \cdot d \sqrt{KT \left(\log\left(1 + \frac{\beta^2_\Theta\beta_{\mathcal{X}}^2 T}{\sigma^2 d^2} \right)+\log\left(\frac{K}{\delta}\right)\right)}.
\end{align*}
\end{theorem}

Theorem~\ref{thm: regretbound-GLRB} establishes that the recourse regret of our \textsf{GLRB} algorithm scales on the order of $\tilde{O}(d\sqrt{KT})$. Unlike the standard linear contextual bandit setting,  where a common parameter is shared across all arms, our setting assigns a distinct  parameter to each arm, resulting in a total of $dK$ parameters. While the regret  lower bound in the shared-parameter case has been characterized in \cite{chu2011contextual}, the disjoint-parameter setting without recourse admits a lower bound of $\Omega(\sqrt{dKT})$. In Section~\ref{sec::lower-bound}, we derive a matching lower bound for the generalized linear recourse bandit problem, showing that our regret upper bound is tight when recourse is available for all features.

\section{Language Model Informed Bandit Recourse Algorithm (\textsf{LIBRA})}\label{sec: LIBRA}

In this section, we propose a new algorithm, which we call the Language Model Informed Bandit Recourse Algorithm (\textsf{LIBRA}). We then analyze the theoretical performance and properties of \textsf{LIBRA}. 

{\bf Motivation}: \textsf{LIBRA} is designed to establish a principled {\em collaboration} between bandit learning methods and LLMs. In practice, neither approach is sufficient on its own. Pure bandit algorithms offer strong theoretical guarantees, but suffer from a severe ``{\em cold start}" problem:  they require costly and unethical exploration before achieving reliable performance. In contrast, LLMs come preloaded with rich medical knowledge and can immediately generate plausible treatment--recourse pairs. However, they also come with their own limitations: they are expensive to deploy, prone to hallucinations that may produce  unsafe or clinically implausible recommendations, and remain difficult to fully trust in high--stakes medical applications. This tension motivates the design of \textsf{LIBRA}, which unifies the statistical reliability and accountability of bandit learning with the expressiveness and prior knowledge embedded in LLMs. The goal is to leverage the speed and clinical fluency of LLMs while retaining the rigorous statistical guarantees of bandit learning methods.

{\bf Technical Challenges}: To achieve this objective, \textsf{LIBRA} treats LLMs and bandits as collaborators rather than competitors, and addresses the following three key technical challenges for LLM--Bandits systems. First, when LLM performance declines (e.g., due to hallucinations or inconsistent recommendations), the bandits component remains robust, does not fail, and continues to learn reliably, creating a reliable LLM--Bandits system ({\em robustness guarantee}). Second, when the LLM does perform well, its  domain knowledge is leveraged to efficiently warm--up the bandits component, enabling faster and more effective learning than bandits alone ({\em warm--start guarantee}). Lastly, \textsf{LIBRA} ensures that interaction with LLM is required only for a finite number of times, leading to a more automated and cost-conscious LLM--Bandits system ({\em LLM--effort guarantee}). Our \textsf{LIBRA} provides all these three technical performance guarantees formally established in \Cref{LIBRA performances}.

\subsection{The \textsf{LIBRA} Algorithm}
In this section, we present our \textsf{LIBRA} algorithm. We first introduce some notation. We define the upper and lower confidence bounds for any context-action pair $(x, a)$ at time $t$, respectively:
\begin{align*}
    \UCB_t(x,a) & := \mu(x^\top \hat{\theta}_{t, a})+ L_\mu \cdot \rho_{t, a}\|x\|_{V_{t, a}^{-1}}, \\
    \LCB_t(x,a) & := \mu(x^\top \hat{\theta}_{t, a})- L_\mu \cdot \rho_{t, a}\|x\|_{V_{t, a}^{-1}}.
\end{align*}
We also define the confidence interval length as $\CI_t(x,a) :=\UCB_t(x,a)-\LCB_t(x,a)= 2 L_\mu \cdot \rho_{t, a} \|x\|_{V_{t, a}^{-1}}$. These quantities summarize the algorithm's current knowledge about the expected reward and its associated uncertainty for any context-action pair $(x, a)$. They form the basic building blocks of \textsf{LIBRA}'s decision-making procedure. The details of our proposed \textsf{LIBRA} are provided in Algorithm~\ref{alg: LIBRA}.

The core steps of \textsf{LIBRA} unfold as follows. At each round \( t \), the algorithm first identifies the optimistic recourse--treatment pair \( (\checkx_t^B,a_t^B) \) by solving the bandit-based recourse optimization problem~\eqref{eq::ORO-Arm}. It then evaluates the reliability of this decision by checking whether the associated confidence intervals, measured by the gap \( \UCB_t(\checkx_t^B,a_t^B) - \LCB_t(\checkx_t^B,a_t^B) \), exceeds a user-specified threshold \( \Delta \). If this interval is too wide, indicating insufficient confidence (i.e., the decision is deemed uncertain), \textsf{LIBRA} queries the LLM and adopts its recommended decision \( (\checkx_t^L,a_t^L) \).
Conceptually, \textsf{LIBRA} acts as a \emph{gatekeeper}: the LLM is allowed to override the default bandit decision only when the bandit lacks sufficient confidence. After executing the selected recourse--treatment action, whether chosen by the bandit or the LLM, the algorithm observes the resulting outcome and updates the corresponding uncertainty set \( \Theta_{t, a} \), enabling progressively more confident bandit decisions over time.

\floatname{algorithm}{Algorithm}
\begin{algorithm}
\caption{Language Model Informed Bandit Recourse Algorithm (\textsf{LIBRA})}\label{alg: LIBRA}
\begin{algorithmic}
\State \textbf{Input}: Time horizon $T$, control parameter $\gamma$, and threshold parameter $\Delta$.
\For{$t=1,\cdots,T$}
\State Observe the context $x_t = (x_{t, I}, x_{t, M})$.
\State Generate the Bandit-based recourse-treatment pair 
$(\checkx_t^B,a_t^B)$ 
by solving the optimistic recourse optimization problem \eqref{eq::ORO-Arm}
, where $x_t^B=(x^B_{t, I}, \check{x}_{t, M}^B)$.
\If{$\UCB_t(\checkx_t^B,a_t^B)-\LCB_t(\checkx_t^B,a_t^B)>\Delta$}
\State LLM generates recourse--treatment pair $(\checkx_t^L,a_t^L)$. 
\State Set the recourse $\checkx_t = \checkx_t^L$ and select the treatment $a_t=a_t^L$.
\Else
\State Set the recourse $\checkx_t = \checkx_t^B$ and select the treatment $a_t = a_t^B$.
\EndIf 
\State Implement the recourse $x_t$ and corresponding treatment $a_t$. 
\State Observe the feedback outcome $r_t$ and update the uncertainty set $\Theta_{t, a}$. 
\EndFor 
\end{algorithmic}
\end{algorithm}

\subsection{Theoretical Performance Guarantees of \textsf{LIBRA}}\label{LIBRA performances}

In this section, we establish the theoretical performance properties of the \textsf{LIBRA} algorithm (Algorithm~\ref{alg: LIBRA}). We begin with the {\em warm-start guarantee}, which formalizes how \textsf{LIBRA} benefits from high-quality recommendations provided by the LLM, particularly during the initial rounds when pure bandit-based methods typically struggle with severe cold-start effects.

\begin{theorem}[Warm-Start Guarantee of \textsf{LIBRA}]\label{thm: warm start guarantee}
    Suppose that the LLM provides an $\eta$-suboptimal recommendation at round $t$, i.e.,
     $r_t(\checkx_t^L,a_t^L)\geq r_t(\checkx_t^\star, a_t^\star)-\eta$ for some constant $\eta>0$. Assume further that 
     $\theta^\star_a \in \Theta_{t, a}$ for all $a\in \mathcal{A}$. Then, the one-step regret of the \textsf{LIBRA} is bounded by: ~\looseness=-1
    \[
    \regret(t) \leq \max\left\{1, \frac{\eta}{\Delta}\right\}\cdot\min \big\{ \Delta, \CI_t(\checkx_t^B,a_t^B) \big\}.
    \]   
\end{theorem}
Intuitively, when the confidence threshold $\Delta$ satisfies the condition $\Delta \ge \eta$, the regret bound then simplifies to $\regret(t) \le \min\{\Delta, \CI_t(\checkx_t^B,a_t^B)\} < \CI_t(\checkx_t^B,a_t^B)$ whenever $\Delta < \CI_t(\checkx_t^B,a_t^B)$. This implies that \textsf{LIBRA} is guaranteed to improve over the bandit-only baseline whenever the LLM provides sufficiently accurate advice and the threshold $\Delta$ is chosen conservatively. Moreover, as shown in \Cref{lemma: glm-radius}, the coverage condition $\theta^\star_a \in \Theta_{t, a}$ for all $a\in \mathcal{A}$ holds uniformly for all $t \in [T]$ with high probability, ensuring that this guarantee applies throughout the learning horizon. In particular, during early rounds when pure bandit algorithms often incur large regret due to exploration, \textsf{LIBRA} can effectively exploit the LLM's prior knowledge to reduce regret, highlighting the practical benefit of integrating generative AI's guidance with principled statistical exploration.

We next establish the {\em LLM-effort guarantee}, which shows that interaction with the LLM is required only a finite number of times under certain regularity conditions. This property ensures that \textsf{LIBRA} does not over-rely on costly or potentially inconsistent LLM feedback, but instead becomes increasingly self-sufficient as more data are collected.

\begin{theorem}[LLM-Effort Guarantee]\label{thm: LLM-effort}
With probability at least $1-\delta$ and over a time horizon $T$, the total number of rounds in which \textsf{LIBRA} adopts the LLM's recommendation is bounded by:
\[
N_{\mathrm{LLM}}(T,\Delta):=\frac{8 L_\mu^2}{\Delta^2} \; \rho_T^2 \;
dK \; \log \Bigl(1+\frac{\beta_\cX^2 T}{\lambda dK}\Bigr),
\]
where $\rho_{T} = \frac{1}{c_{\mu}} 
\left(
 \sigma \sqrt{d \log \left( 1 + \frac{\beta_{\mathcal{X}}^2 T}{\lambda} \right) + d\log \left( \frac{K}{\delta} \right)} +
\sqrt{\lambda} \beta_{\Theta}
\right)$.
\end{theorem}
This theoretical result shows that the number of rounds in which the LLM is actively queried grows only logarithmically with the time horizon, i.e., \(N_{\mathrm{LLM}}(T,\Delta) = O(\log^2 T)\). Consequently, \textsf{LIBRA} primarily relies on the LLM during early rounds to bootstrap learning, while gradually transitioning to the autonomous bandit-driven decisions as more data are gathered.
Moreover, the above bound is inversely proportional to the confidence threshold \(\Delta\), which aligns with our key intuition: if the LLM is deemed more reliable (i.e., we use a smaller \(\Delta\)), its recommendations are trusted more often. Conversely, a larger threshold \(\Delta\) reflects lower confidence in the LLM and triggers an earlier fallback to the bandit-based decisions. This highlights \textsf{LIBRA}'s ability to {\em adaptively calibrate} its reliance on the LLM based on both trust in the model and the amount of data available.

Building on the one-step regret result in Theorem~\ref{thm: warm start guarantee}, we now present our next guarantee for \textsf{LIBRA}. ~\looseness=-1

\begin{theorem}[Improvement Guarantee of \textsf{LIBRA}]\label{thm: LIBRA improvementguarantee}
    Under the same conditions as in \Cref{thm: regretbound-GLRB},  suppose that the LLM is $\eta$-suboptimal in every round, i.e.,
     $r_t(\checkx_t^L,a_t^L)\geq r_t(x^\star,a^\star)-\eta$ for all $1\leq t\leq T$. Then, with probability at least $1-\delta$, the cumulative recourse regret of \textsf{LIBRA} is bounded by:
    \[
    \begin{aligned}
    \Regret_\pi(T) 
    &\leq 
    \max\left\{1, \frac{\eta}{\Delta}\right\}\cdot\min \Bigg\{ \Delta T,\  2L_{\mu} \cdot \rho_{T}  \sqrt{2dKT \log\left(1 + \frac{\beta_{\mathcal{X}}^2 T}{\lambda d} \right)}\Bigg\},
    \end{aligned}
    \]
where $\rho_{T} = \frac{1}{c_{\mu}} 
\left(
 \sigma \sqrt{d \log \left( 1 + \frac{\beta_{\mathcal{X}}^2 T}{\lambda} \right) + d\log \left( \frac{K}{\delta} \right)} +
\sqrt{\lambda} \beta_{\Theta}
\right)$. In particular, upon setting $\Delta=\Theta(\eta)$ and $\lambda=\Theta(\sigma^2d/\beta^2_{\Theta})$, the cumulative recourse regret reduces to $\Regret_\pi(T)=\tilde{O}\left(\min\{\eta T, d\sqrt{KT}\}\right)$.
\end{theorem}

Below, we discuss several key insights derived by the above technical performance guarantee. 

\paragraph{(i) Benefits from Accurate LLM Guidance.}
The theorem highlights the benefit of incorporating LLM guidance when its recommendation is accurate. In particular, when the LLM is consistently $\eta$-suboptimal with a small $\eta$, and the confidence threshold $\Delta$ is chosen to closely match this accuracy level (i.e., $\Delta \approx \eta$), \textsf{LIBRA} achieves a cumulative regret of order $\tilde{O}(\eta T)$, which can be substantially smaller than the $\tilde{O}(d\sqrt{KT})$ regret of a pure bandit algorithm. In this regime, \textsf{LIBRA} effectively leverages the LLM's prior knowledge to reduce exploration costs during the early rounds, providing a strong warm start. As more data are collected, \textsf{LIBRA} gradually shifts its reliance toward the bandit component, which continues to refine the policy and guide the learning process toward the true optimal solution. This dynamic interplay allows \textsf{LIBRA} not only to benefit from the LLM's warm start, but also to ultimately surpass the LLM—achieving better asymptotic performance than the LLM alone could offer.

\paragraph{(ii) Robustness Against Poor LLM Performance.}
Theorem~\ref{thm: LIBRA improvementguarantee} further establishes a robustness guarantee for \textsf{LIBRA} when the LLM provides unreliable guidance. Even when $\eta$ is large, indicating poor LLM performance, so long as the confidence threshold $\Delta$ is chosen as a conservative overestimate of $\eta$, the regret bound smoothly degrades and never exceeds that of the underlying bandit algorithm. In such cases, \textsf{LIBRA} quickly reduces its reliance on the LLM and effectively reverts to standard bandit-based exploration, consistent with the finite LLM-effort guarantee as in \Cref{thm: LLM-effort}. As a result, this ensures that \textsf{LIBRA} is {\em safe} to deploy even when the quality of the LLM is uncertain, as it can never perform worse than the bandit baseline once an accurate estimate of $\eta$ is obtained.

\paragraph{(iii) Impact and Calibration of the Confidence Threshold \texorpdfstring{$\Delta$}{Delta}.}
The choice of the confidence threshold $\Delta$ plays a central role in balancing the trust placed on the LLM versus the bandit algorithm. Fortunately, Theorem~\ref{thm: LIBRA improvementguarantee} ensures that the regret remains stable over a fairly wide range of $\Delta$ values. As long as $\Delta = \Theta(\eta)$, the regret guarantee stays within a constant factor of the optimal tradeoff. This insensitivity makes \textsf{LIBRA} {\em robust} to modest tuning errors in $\Delta$. In practice, when some prior knowledge is available (e.g., assessments from expert clinicians or other human decision-makers regarding the reliability of the LLM), this information can be deployed to guide the calibration of $\Delta$ for improved performance. Alternatively, when no prior estimate of $\eta$ is available, $\Delta$ may be adapted online using held-out validation or pessimistic reward estimates, which could further enhance \textsf{LIBRA}'s flexibility. Overall, $\Delta$ serves as a simple yet interpretable knob that governs how quickly the algorithm transitions from relying on pure generative advice to trusting its own learned experience.

\subsection{Lower Bounds on Recourse Regret}
\label{sec::lower-bound}

Having established the recourse regret upper bounds in \Cref{thm: regretbound-GLRB,thm: LIBRA improvementguarantee}, we now derive matching lower bounds to demonstrate that our recourse regret upper bounds are nearly {\em tight}.
\begin{theorem}[Recourse Regret Lower Bounds]
\label{thm::lower-bounds}
    For the generalized linear recourse bandit problem introduced in \Cref{sec: GLRB}, the recourse regret of any policy $\pi$ satisfies:
    \begin{equation}
        \Regret_\pi(T) =\Omega\left(\gamma d_{M}\sqrt{KT} \vee \sqrt{d KT}\right).
    \end{equation}
    Furthermore, suppose there exists an $\eta$-suboptimal oracle—that is, at each round $t$, the oracle outputs a recourse $\checkx_t$ and an arm $a_t$ satisfying $r(\checkx_t,a_t)\geq r(\checkx_t^\star,a_t^\star)-\eta$. Then, the regret is lower bounded by:
    \begin{equation}
        \Regret_\pi(T) =\Omega\left(\left(\gamma d_{M}\sqrt{KT} \vee \sqrt{d KT}\right) \wedge \eta T\right).
    \end{equation}
\end{theorem}

The first lower bound highlights two distinct sources of difficulty.  
The first term ({\em recourse complexity}), $\gamma d_M \sqrt{KT}$, captures the fundamental challenge introduced by the recourse procedure, which is governed by the perturbation radius $\gamma$ and the dimension $d_M$ of the mutable context. The perturbation radius $\gamma$ limits the magnitude of admissible adjustments to the mutable features, while the dimension $d_M$ dictates how hard it is to optimize over possible adjustments. The second term ({\em contextual complexity}), $\sqrt{dKT}$, reflects the intrinsic difficulty of learning the underlying contextual model over the full context dimension $d$, though it scales more benignly, as the immutable context does not require optimization through recourse and is therefore easier to learn. In the presence of an $\eta$-suboptimal oracle, the recourse regret is additionally bounded above by $\eta T$, because one can always defer to the oracle to guarantee an $\eta$-accurate solution at each round. Consequently, the true or effective regret lower bound is determined by the minimum between the inherent learning difficulty (driven by dimensionality and recourse radius) and the maximum regret incurred by repeatedly using the oracle.

Our recourse regret upper bounds nearly match the lower bound in Theorem~\ref{thm::lower-bounds} in the regime, where the perturbation radius $\gamma$ is constant and the mutable context dimension satisfies $d_M \asymp d$. In this setting, our \textsf{GLRB} algorithm (Algorithm~\ref{alg:GLRB}), which follows a LinUCB-based design, achieves regret that is optimal up to logarithmic factors. While more sophisticated methods such as SupLinUCB could theoretically eliminate the remaining logarithmic gap, we intentionally adopt the simpler LinUCB-based design due to its ease of practical implementation and strong empirical performance. In practice, the added complexity of more advanced methods such as SupLinUCB is often unnecessary, a conclusion supported by both prior theoretical and empirical studies (see e.g., \citealt{li2010contextual,chu2011contextual}). %

\section{Generalized Linear Recourse Bandits with Non-Compliance}\label{sec::noncompliance}

In \Cref{sec: GLRB,sec: LIBRA}, we assumed that patients faithfully follow the prescribed recourses; that is, they modify their mutable features exactly as recommended. In practice, however, this assumption can be overly optimistic \citep{osterberg2005adherence, pruitt2025silent}. Specifically, patients may fail to comply with physician guidance for different reasons: they may forget to exercise, find it difficult to maintain a recommended diet, or simply lack the motivation or resources to achieve the prescribed lifestyle changes 
\citep{dusetzina2023cost, patel2025understanding}. As a result, the intended recourse may be only partially implemented—or ignored altogether \citep{ho2006effect, diaz2025delivery}. Such forms of non-compliance are not exceptions, but rather common realities in many healthcare applications.

To address this gap, we extend our framework to explicitly incorporate {\em non-compliance}, where deviations from the recommended recourse may occur either (i) {\em randomly} (e.g., when a patient unintentionally underperforms the prescribed exercise routine; see \citealt{stewart2023medication}) or (ii) {\em adversarially} (e.g., systematic deviations due to persistent behavioral barriers; see \citealt{patel2025understanding}). 
Formally, we consider the following {\em non-compliance setting}: in each round $t$, the learner first observes a clean context $x_t$, and selects a recourse $\checkx_t$ and an action $a_t$ jointly. The environment (i.e., the patient) may subsequently perturb the implemented context to $\barx_t$ in a random or adversarial manner. Finally, the learner observes the perturbed context $\barx_t$ and receives a stochastic reward $r_t=\mu\left(\theta_{a_t}^{\top}\barx_t\right)+\xi_t$. In what follows, we analyze two distinct perturbation mechanisms for generating $\barx_t$, including (i) random non-compliance and (ii) adversarial non-compliance, and propose principled mitigation strategies for each.

\subsection{Random Patient Non-Compliance}

We begin with the random non-compliance setting. At each round $t$, the learner intends to implement the recourse-adjusted context $\checkx_t$; however, the {\em observed} context may differ due to imperfect adherence. Specifically, the patient may not follow the prescribed recourse precisely, resulting in the realized context $\barx_t=\checkx_t+\epsilon_t$, where $\epsilon_t$ is a random perturbation satisfying $|\epsilon_t|\leq \epsilon$ almost surely. Importantly, we do not assume that $\epsilon_t$ has zero mean; that is, the non-compliance can be systematically biased in a particular direction.

To handle the random non-compliance setting, we adapt our \textsf{GLRB} algorithm  (Algorithm~\ref{alg:GLRB}) with a slight modification to the construction of the uncertainty set $\Theta_{t, a}$. Specifically, at each round $t$, we construct the confidence ellipsoid using all past \textit{realized} (possibly perturbed) contexts and observed rewards as:
\begin{equation}
    \Theta_{t, a}:=
        \left\{
        \theta_a:\|\theta_a-\widehat{\theta}_{t, a}\|_{V_{t, a}}\leq \rho_{t, a}
        \right\},
\end{equation}
where $V_{t, a}:= \lambda I + \sum_{s \in \mathcal{I}_{a, t-1}} \barx_s   \barx_s^\top$ and $\widehat{\theta}_{t, a} = 
\argmin_{\theta\in \mathbb{R}^d} 
\left\{
\sum_{s \in \mathcal{I}_{a, t-1}}
\ell(r_s, \barx_s^\top \theta) + \frac{\lambda}{2} \; \norm{\theta}_2^2
\right\}$. Given this uncertainty set and the regularized maximum-likelihood estimator (MLE), we 
select the recourse $\checkx_t$ and action $a_t$ by solving a {\em robust} variant of the optimistic recourse optimization problem \eqref{eq::ORO} as follows:
\begin{align}
\label{eq::robust-ORO}
   \max_{a\in \mathcal{A}} \max_{\checkx_M\in \mathbb{R}^{d_M}, \; \theta_a\in \Theta_{t, a}} &\quad \bE_{\epsilon}\left[\mu\left((\checkx_M+\epsilon)^\top \theta_{a,M} +x_I^\top \theta_{a,I}\right)\right]\tag{\textsf{robust-ORO}}\\
    \quad \text{subject to}\hspace{1cm} &  \quad \norm{\checkx_M- x_M} \leq \gamma. \notag
    \end{align}

\begin{remark}
    In practice, the observed deviation data $\{\epsilon_{s}:=\barx_s-\checkx_s: s\in \cI_{a, t-1}\}$ can be exploited to construct an empirical distribution of the random non-compliance. As the time horizon increases, the resulting estimation error due to this sample average approximation shall converge to zero.
\end{remark}

Algorithmically, this variant is almost identical to the \textsf{GLRB} algorithm; the key structural difference is that the current perturbed context $\barx_t$ is not available at decision time. 
The learner therefore acts based on the intended recourse $\checkx_t$, but update its model using feedback generated from the realized context $\barx_t$.
Intuitively, this creates a measurement/implementation mismatch: we optimize using the intended adjustment, but nature feeds back outcomes from the actually realized adjustment. Our next result formally characterizes its effect on the corresponding regret.

\begin{theorem}[Random Non-Compliance Recourse Regret]
\label{thm::random-noncompliance}
    Suppose the random non-compliance $\epsilon_t$ is almost surely bounded by $\epsilon$, i.e., $|\epsilon_t|\leq \epsilon$ for all $t\geq 0$. With probability at least $1-\delta$,
\begin{equation}
    \Regret_\pi(T) 
    \leq 2L_{\mu} \cdot \rho_{T}  \sqrt{2dKT \log\left(1 + \frac{(\beta_{\mathcal{X}}+\epsilon)^2 T}{\lambda d} \right)}.
\end{equation}
\end{theorem}
Compared to the regular setting, we observe that random non-compliance increases the regret only through a logarithmic term, 
$\log(\beta_\cX+\epsilon)$. In particular, bounded random deviations do not alter the leading-order regret rate. As we will show later, the regret analysis becomes substantially more challenging in the adversarial setting, and non-compliance has a much more significant impact on the resulting regret.

\subsection{Adversarial Patient Non-Compliance}
We now consider the adversarial non-compliance setting in which the environment (i.e., the patient) may alter the recommended recourse arbitrarily within a fixed perturbation budget. Specifically, at each round $t$, after the learner suggests a recourse $\checkx_{t,M}$, the patient may instead shift to a perturbed recourse $\barx_t$ such that $\norm{\checkx_{t,M}-\barx_{t,M}}\leq \epsilon$, where $\epsilon$ quantifies the severity of non-compliance. Larger values of $\epsilon$ correspond to more severe or unpredictable deviations, while $\epsilon = 0$ recovers the full-compliance setting.

\smallskip
{\bf Optimistic Recourse Optimization under Adversarial Non-Compliance}: To address this setting, we retain the same procedure for constructing the uncertainty set $\Theta_{t, a}$ as in the random non-compliance setting, but modify the decision rule. Instead of \eqref{eq::ORO}, we solve the following robust optimization problem:
\begin{align}\label{eq::ORO-NC}
   &\max_{a\in \mathcal{A}} \max_{\checkx_{t,M}\in \bR^{d_M},\theta_a\in \Theta_{t, a}} \min_{\barx_{t,M}\in \bR^{d_M}} \quad (\barx_{t,M}, x_{t,I})^\top \theta_a \tag{\textsf{ORO-NC}}\\
   & \text{subject to} \quad \norm{\checkx_{t,M}- x_{t,M}}\leq \delta  \ 
   \text{ and }\  
    \norm{\checkx_{t,M}-\barx_{t,M}}\leq \epsilon, \notag
\end{align}
where $\barx_t = (x_{t,M}, \barx_{t,I})$ is the realized context.
Since the patient's non-compliance behavior is unknown, the learner adopts a worst-case perspective through this formulation \eqref{eq::ORO-NC}. The learner first selects a recourse $\checkx_t$ within distance $\delta$ of the original features, but then anticipates that the patient may subsequently deviate to some $\barx_t$ within distance $\epsilon$ of the target. By optimizing against this worst-case perturbation, the learner ensures robustness to adversarial non-compliance. Accordingly, the learner needs to solve the minimization problem over $\barx_t$ such that $\|\checkx_t-\barx_t\|\leq \epsilon$. That is, we attempt to maximize the minimum reward when $\barx_t$ perturbs around $\checkx_t$. Then, similarly to the compliance setting in \Cref{sec: GLRB,sec: LIBRA}, the learner selects an optimistic solution in the face of uncertainty over the uncertainty set $\Theta_{t, a}$. 

\smallskip
{\bf Block Coordinate Descent for Solving \eqref{eq::ORO-NC}}: Our formulation \eqref{eq::ORO-NC} is a {\em bilevel optimization} problem with a nonlinear objective and nonlinear constraints, making it challenging to solve to global optimality even when the Mahalanobis norms are applied. To address this challenge, we propose a block coordinate descent approach that efficiently finds a local stationary point. To streamline the presentation, we consider the following simplified single-arm problem:
\begin{align}
   &\max_{\checkx_{t,M},\theta} \min_{\barx_{t,M}} \quad \barx_{t, M}^\top \theta_{a, M} + x_{t, I}^\top \theta_{a, I} \tag{\textsf{ORO-NC-Arm}}\label{eq::ORO-NC-Arm}\\
   & \text{subject to} \hspace{0.3cm}  \norm{\checkx_{t,M}- x_{t,M}}\leq \delta, 
   \notag \\
  &
   \hspace{1.82cm}\norm{\checkx_{t,M}-\barx_{t,M}}\leq \epsilon, \notag\\
   & \hspace{1.82cm} \big\|\theta_a - \hat\theta_{t, a} \big\|_{V_{t, a}}\leq \rho_{t, a}. \notag
\end{align}

We design a two-block coordinate descent algorithm (\Cref{alg:alternative-optimization-noncompliance}) that alternates between updating (i) the model parameter $\theta$ given the current context and (ii) the recourse/perturbation pair $(\checkx_{M}, \barx_{M})$
given the updated parameter. 
Due to the decomposable structure of \eqref{eq::ORO-NC}, both sub-problems admit closed-form solutions. The closed-form solution to the $\theta$-update \eqref{eq::theta-optimization} is derived through \Cref{lem: closed-form-solution}, while the closed-form solution to the $(\checkx_{M}, \barx_{M})$-update \eqref{eq::x-optimization} is characterized by the following lemma.

\begin{algorithm}
\caption{
Two-Block Coordinate Descent Algorithm
for Solving \eqref{eq::ORO-NC-Arm}}\label{alg:alternative-optimization-noncompliance}
\begin{algorithmic}
\State \textbf{Initialization:}
$\checkx_M^{(0)}=\barx_M^{(0)}=x_M$.  
\State \textbf{Repeat for $k = 0, 1, ...$ as:}
\begin{align}
&  \theta_a^{(k+1)} = \argmax_{\norm{\theta_a -\hat{\theta}_a}_{V_a}\leq \rho_a} \barx_M^{(k)\top} \theta_{a,M} +x_I^{\top} \theta_{a,I}
\tag{\textrm{\emph{Optimize $\theta_a$}}}\label{eq::theta-optimization}\\
& \checkx_M^{(k+1)}, \barx_M^{(k+1)} = \argmax_{\norm{\checkx_M- x_M} \leq \delta}\min_{\norm{\checkx_M-\barx_M} \leq \epsilon}  \barx_M^{\top} \theta_{a,M}^{(k+1)}
\tag{\textrm{\emph{Optimize $\checkx_M, \barx_M$}}}\label{eq::x-optimization}
\end{align}
\end{algorithmic}
\end{algorithm}

\begin{lemma}
    \label{lem: xx-optimization}
    The optimal solution to the $(\checkx_{M}, \barx_{M})$-update \eqref{eq::x-optimization} is given by $\checkx_M^\star=x_M+\gamma\cdot v$ and $\barx_M^\star=x_M+(\delta-\epsilon)\cdot v$, where $v\in \partial \norm{\theta_M}_{\star}$.
\end{lemma}

Analogous to \Cref{lem::converge-to-critical-points}, \Cref{alg:alternative-optimization-noncompliance} is guaranteed to converge to a critical point of \eqref{eq::ORO-NC-Arm}. Furthermore, we show that \eqref{eq::ORO-NC-Arm} satisfies the KL property (see Definition \ref{def: KL}) under a mild condition in the following lemma. This enables us to characterize the theoretical convergence rate of \Cref{alg:alternative-optimization-noncompliance} for specific problem instances, as we formalized in Theorem~\ref{thm::convergence-rate-KL}. Together, these results ensure that our robust approach is not only theoretically sound but also computationally practical.

\begin{lemma}[The KL Property of the \eqref{eq::ORO-NC-Arm} Problem]
    Suppose that the norm function $\norm{\cdot}$ is semialgebraic. Then, the \eqref{eq::ORO-NC-Arm} problem 
    \begin{equation}
        \max_{\norm{\theta_a-\hat{\theta}_a}_{V_a}\leq \rho_a, \norm{\checkx_{t,M}-x_{t,M}}\leq \delta}\min_{\norm{\checkx_{t,M}-\barx_{t,M}}\leq \epsilon} \barx_{t, M}^\top \theta_{a, M} + x_{t, I}^\top \theta_{a, I}
    \end{equation}
    satisfies the KL property.
    \label{lem::KL}
\end{lemma}
Before formally characterizing the regret bound, we introduce the following assumption, which we call the \emph{benign coverage condition}. To ensure the learner can effectively distinguish between the effects of different actions, the mutable features must exhibit a degree of ``richness" or diversity. Without sufficient variation, an adversary could potentially hide suboptimal actions within the static directions of the context, potentially leading to linear regret. We formalize this requirement through a coverage condition, which ensures that the mutable context provides enough information to explore the parameter space effectively.
\begin{assumption}[Benign Coverage Condition]
\label{ass::coverage}
There exists a constant $\gamma>0$ such that for every arm $a$ and time $t$, we have the following:
\[
\sum_{s\in\mathcal I_{t,a}} P_M \bar x_s\bar x_s^\top P_M\ \succeq\ \gamma^2\sum_{s\in\mathcal I_{t,a}} P_M,
\]
where $P_M$ is the projection matrix onto the coordinates corresponding to the mutable context $x_M$. 
\end{assumption}
This condition ensures that the mutable context exhibits sufficient variability, which is essential for achieving sublinear regret. %
From a practical standpoint, this condition is expected to hold with high probability in most real-world applications. Mutable contexts, such as physiological signals like heart rate or blood pressure, naturally exhibit inherent stochastic fluctuations over time (see \citealt{kulkarni2025blood, jarczok2022heart}). Indeed, if these variables did not vary naturally, they would likely be unresponsive to the interventions or recourse strategies being studied.
\begin{theorem}[Adversarial Non-Compliance Recourse Regret]
\label{thm::adversarial-noncompliance}
    Suppose that the adversarial non-compliance $\epsilon_t$ satisfies $\norm{\epsilon_t}\leq \epsilon$ for all $t\geq 0$. Under Assumption~\ref{ass::coverage}, with probability at least $1-\delta$,
\begin{equation}
    \Regret_\pi(T) 
    \leq 2L_{\mu} \cdot \rho_{T}  \sqrt{2dKT \log\left(1 + \frac{(\beta_{\mathcal{X}}+\epsilon)^2 T}{\lambda d} \right)}+\frac{4L_\mu \epsilon}{\gamma}\cdot \sqrt{KT}.
\end{equation}
\end{theorem}
This technical result reveals two distinct sources of regret under adversarial non-compliance. The first term, which scales as \( \tilde{O}(d\sqrt{KT}) \), corresponds to the standard statistical regret arising from learning the unknown model parameters \( \{\theta_a^\star\}_{a\in \cA} \), while the second term, \( O(\frac{\epsilon}{\gamma} \cdot \sqrt{KT}) \), quantifies the additional price of robustness against worst-case non-compliance. This penalty scales linearly with the perturbation budget \( \epsilon \) and inversely with the variability constant \( \gamma \). Therefore, stronger non-compliance (larger $\epsilon$) or limited diversity in the mutable contexts (smaller $\gamma$) can significantly amplify regret.
Importantly, this behavior contrasts sharply with random non-compliance setting, where the impact of the perturbation radius \(\epsilon\) enters only logarithmically through a term of order \(\log(\beta_{\mathcal{X}}+\epsilon)\).
Under the adversarial non-compliance setting, the dependence on \(\epsilon\) is {\em polynomial}, reflecting the fundamentally harder nature of worst-case deviations.
Nevertheless, when the benign coverage condition (Assumption~\ref{ass::coverage}) holds and the level of non-compliance remains moderate, the overall regret remains sublinear, demonstrating that our robust approach effectively mitigates the worst-case behavioral deviations without sacrificing long-run learning performance.

\section{Case Study: Personalized Hypertension Management}

In this section, we show how our proposed algorithms (\textsf{GLRB} and \textsf{LIBRA}) can be instantiated in a real clinical decision-making problem. Our goal is to illustrate the full pipeline, from context modeling to recourse optimization to LLM–bandit collaboration, and to highlight the clinical insights that emerge from our theoretical framework. In addition to the real-world case, we include experiments based on both synthetic in \Cref{sec:syn} and semi-synthetic data in Appendix \ref{app:semi-synthetic} to more rigorously evaluate the behavior and performance of the proposed algorithms.

\subsection{Background and Current Clinical Gap}
We focus on the critical problem of hypertension control for patients with type 2 diabetes (T2D), a setting where both {\em treatment selection} and {\em lifestyle adjustments} (recourse) play essential roles in shaping patient outcomes. A large body of medical literature emphasizes that {\em lifestyle modification is not optional but fundamental} to the hypertension management, particularly for patients with T2D at elevated cardiovascular risk (CVD). Major clinical guidelines emphasize that pharmacologic therapy alone is often {\em insufficient}. For example, the 2017 ACC/AHA hypertension guideline \citep{whelton20182017}  explicitly highlights that nonpharmacological interventions remain a cornerstone of hypertension management and improve control even when medications are required. This includes diet, sodium restriction, weight loss, and physical activity. Similarly, the landmark DASH (Dietary Approaches to Stop Hypertension) trial showed that a diet rich in fruits, vegetables, and low-fat foods lowers systolic blood pressure (SBP) by 11.4 mmHg in hypertensive adults, a reduction comparable to or larger than many first-line antihypertensive medications \citep{sacks2001effects, yang2025responsible}. The Look AHEAD trial \citep{look2013cardiovascular} demonstrates that the  weight loss and diet improve SBP control significantly beyond medication alone in patients with T2D. Moreover, physical activity has been shown to exert similarly strong antihypertensive effects. For example, \cite{cornelissen2013exercise} found that structured exercise interventions yield SBP reductions comparable to antihypertensive medications. Through a randomized clinical trial, \cite{blumenthal2021effects} shows that {\em both} lifestyle modifications and medications produce {\em additive effects}: combining lifestyle intervention with pharmacotherapy achieved significantly greater SBP reductions than medication alone. Likewise, \cite{forouzanfar2017global} highlights lifestyle adjustments—including diet quality, sodium restriction, and physical activity—as high-impact, low-risk interventions essential to all stages of hypertension management.

Collectively, this body of medical evidence underscores a critical insight: {\em for most hypertensive patients, the most effective clinical strategy is not simply choosing the right medication, but pairing pharmacologic treatment with personalized, actionable lifestyle adjustments tailored to each patient's modifiable characteristics}. Despite the strength and consistency of this evidence, clinicians still {\em lack} principled decision-support tools that quantify {\em which} lifestyle modification, such as dietary improvement or increased physical activity is likely to yield the greatest benefit for a given patient. This central research gap strongly motivates our theory: a recourse-aware decision-making framework that jointly optimizes medication choice and personalized lifestyle adjustments, aligning directly with contemporary clinical practice and medical science.

\subsection{Data Description}

We illustrate how our methods are applied to a personalized medicine problem in hypertension management. We use the Action to Control Cardiovascular Risk in Diabetes (ACCORD) dataset \citep{accord2010effects, action2008effects} derived from a landmark clinical study involving patients diagnosed with type 2 diabetes (T2D). Specifically, we focus on a subset of ACCORD participants who have high blood pressure or hypertension, and are at elevated risk for cardiovascular disease (CVD) events, such as stroke and heart failure. The resulting data provides detailed information on how clinicians prescribed and adjusted different treatment strategies to control hypertension. We should note that we include those patient features that have been proven in the medical literature \citep{echouffo2013risk, chowdhury2022prediction, qin2023machine} to have a clinically meaningful impact on  systolic blood pressure (SBP), which serves as the {\em primary outcome} for evaluating hypertension. Accordingly, our dataset includes the following patient information: 
\begin{itemize}
	\item {\em Demographics}: baseline age, gender, and race group.
	\item {\em Baseline conditions}: 
    history of CVD events (cvd hx baseline), and smoking habit (cigarette baseline). 
	\item {\em Clinical variables}: 
    systolic blood pressure (current SBP). 
    \item {\em Diet and physical activities}: diet score (DietScore) capturing how much the patient takes low-fat and lean meat choices and avoids fried foods and fat), and physical activity hours (PhyActHours) capturing how many hours of physical activities the patients have per week. 
    \item {\em Antihypertensive agents}: The medications the patient takes, focusing on beta-blocker, angiotensin-converting enzyme (ACE) inhibitors, diuretics (Diur), and calcium channel blockers (CCB). 
\end{itemize}

The primary research question we seek to address is how our collaborative LLM-Bandits framework can act as a practical decision-support tool to help physicians recommend tailored treatments for these patients. We should note that among the above-described patient features, DietScore and PhyActHours naturally play the role of {\em lifestyle recourses} in our empirical results.  %

\subsection{Empirical Results and Insights}

We compare the following four algorithms to better evaluate the performance of our methodology:
\begin{itemize}
    \item \textbf{LinUCB:} This is the standard contextual linear bandit algorithm \citep{chu2011contextual}, which serves as a baseline that can only recommend treatments but cannot suggest or learn from lifestyle recourses.
    \item \textbf{LLM-only:} This benchmark exclusively follows recommendations generated by the LLM. This benchmark measures the effectiveness of the LLM's general, pre-trained knowledge without any online learning or adaptation. We use GPT-3.5 Turbo as the LLM model. The  prompt example used for querying the LLM is included in \Cref{fig:promptbox}. 
    \item \textbf{GLRB:} This algorithm learns to recommend both treatments and lifestyle recourses but does not incorporate any LLM-based prior knowledge.
    \item \textbf{LIBRA:} This is our proposed Language Model Informed Bandit Recourse Algorithm. This integrates LLM guidance with online bandit learning to provide a warm-start and adaptively recommend treatments and lifestyle recourses. We set $\Delta=1$ and examine the impact of $\Delta$ in \Cref{sec:ablation}. 
\end{itemize}

To construct an environment that enables counterfactual evaluation, namely, assessing what would have occurred under alternative recourse or treatment decisions, we fit separate linear regression models for each treatment arm using our observational dataset. This standard approach \citep{hill2011bayesian} allows us to simulate outcomes for any given context-action-recourse triplet. This modeling choice is consistent with the medical literature, where linear regression models are widely used to predict SBP outcomes (see e.g., \citealt{chowdhury2022prediction, echouffo2013risk}). 

We consider two primary treatment groups: (1) \textbf{Treatment 1:} Beta-blocker (114 samples), and (2) \textbf{Treatment 2:} ACE+CCB+Diur (63 samples).
We focus on these two antihypertensive treatment strategies that are both highly prevalent and clinically central in patients with T2D: beta-blockers and combination therapy based on ACE, CCB, and Diur. 
Major clinical trials and guidelines, including ACCORD and the ACC/AHA hypertension guidelines \citep{whelton20182017}, identify these two strategies as dominant and clinically meaningful alternatives with distinct risk–benefit profiles. Focusing on these two treatment arms thus captures a large fraction of real-world clinical practice while enabling a principled comparison of treatment–recourse tradeoffs in our high-risk population. In the experiments, we bootstrap the patients at each time step from the dataset. 
We also standardize all features so that each coefficient reflects the expected change in SBP at the next visit per one standard deviation change in the corresponding feature. The full set of coefficients for the fitted linear regression models are presented in Table \ref{tab:coefficients}.

\begin{table}[htp]
\centering
\caption{Fitted Coefficients for Counterfactual Simulation Models}
\label{tab:coefficients}
\scalebox{0.86}{
\begin{tabular}{@{}lrr@{}}
\toprule
\textbf{Patient Feature} & \textbf{Treatment 1 (Beta-blocker)} & \textbf{Treatment 2 (ACE+CCB+Diur)} \\ \midrule
Female & -0.51 & 2.73 \\
Baseline age & 1.91 & -1.51 \\
Cvd hx baseline & 0.29 & 1.09 \\
Raceclass & -0.00 & -0.00 \\
Cigarette baseline & 1.43 & -1.78 \\
\textbf{DietScore (Mutable)} & \textbf{-2.12} & \textbf{-2.18} \\
\textbf{PhyActHours (Mutable)} & \textbf{-0.48} & \textbf{-1.02} \\
Current SBP & 8.79 & 10.82 \\ \bottomrule
\end{tabular}}
\end{table}

Our data-driven insights from the results of Table \ref{tab:coefficients} indicates that increases in \textbf{DietScore} have a significantly larger impact on reducing SBP than comparable increases in \textbf{PhyActHours}. In our analysis, we define the {\em patient reward} as 170-(SBP of the next visit), so higher rewards correspond to greater reductions in SBP. We choose 170 mmHg as the baseline SBP for high-blood-pressure patients; this value serves as a normalization constant and can be adjusted without affecting the qualitative conclusions.

\begin{figure}[ht]
    \centering
    \begin{tcolorbox}[
        enhanced,
        colback=blue!3!white,      %
        colframe=blue!40!black,    %
        title=\textbf{\textsc{Prompt Example}}, %
        fonttitle=\bfseries\large,
        boxrule=0.8pt,
        arc=3mm,        %
        left=6mm, right=6mm, top=4mm, bottom=4mm
    ]
    
    \textbf{System Instruction:} You are a helpful medical assistant helping to manage patients with hypertension.

    \vspace{0.2cm}
    \noindent A patient has hypertension. There are two treatment options:
    \begin{itemize}
        \item \textbf{Treatment 1:} Prescribing Beta-blocker.
        \item \textbf{Treatment 2:} ACE Inhibitor + Calcium channel blockers + Diuretics.
    \end{itemize}

    \vspace{0.2cm}
    \noindent \textbf{Input Data:} Given the following patient features:
    
    \begin{tcolorbox}[colback=white, colframe=gray!20, boxrule=0.5pt, arc=1mm, left=2mm, right=2mm]
    \footnotesize\ttfamily
    \{"female": Female, "age": 0.75, "cvd\_hx\_baseline": 1.63, "race": Non-White, \\
    "smoking": 0.30, "DietScore": 1.02, "PhyActHours": -0.43, \\
    "Systolic Blood Pressure for this visit": -0.75\}
    \end{tcolorbox}

    \vspace{0.1cm}
    {\small \textit{Note: Features are standardized (mean=0, std=1). A larger `DietScore' indicates a healthier diet.}}

    \vspace{0.3cm}
    \noindent \textbf{Task:}
    The goal is to minimize the Systolic Blood Pressure for the NEXT visit. Please recommend the optimal treatment (1 or 2) and suggest changes ONLY to `DietScore' and `PhyActHours'.
    
    \vspace{0.2cm}
    \noindent \textbf{Constraints:}
    \begin{itemize}
        \item Make sure the two norms of the feature changes are within 3 units.
        \item Do not analyze, only Respond in the format: \texttt{treatment=..., DietScore=..., PhyActHours=...}
    \end{itemize}

    \end{tcolorbox}
    \caption{Structure of the medical assistant (LLM) prompt example used in our case study.}
    \label{fig:promptbox}
\end{figure}

\begin{figure}[htp]
  \centering
  \begin{subfigure}[b]{0.45\textwidth}
    \includegraphics[width=\textwidth]{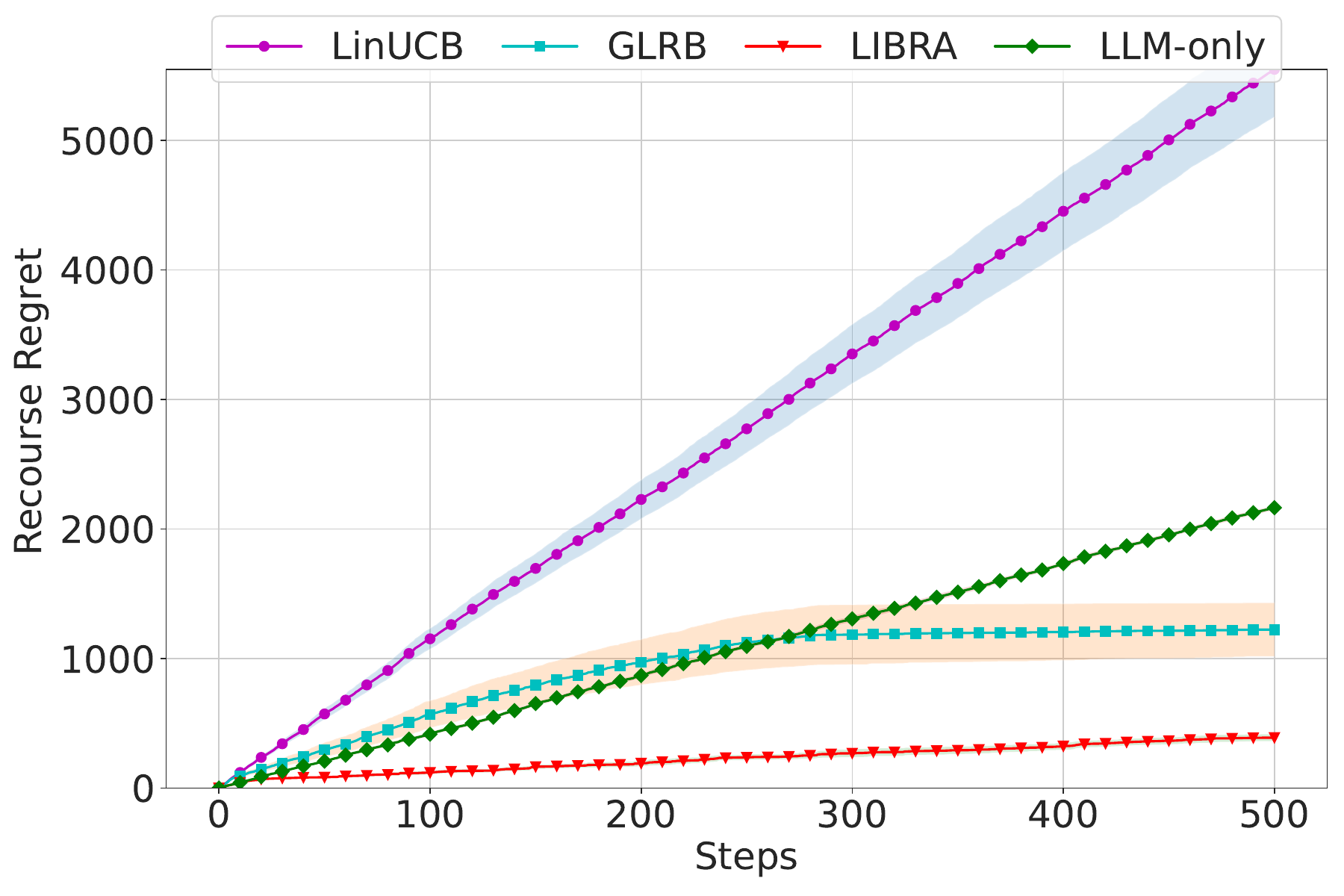}
    \caption{Cumulative Recourse Regret.}
    \label{fig:case_regret}
  \end{subfigure}
  \hfill 
  \begin{subfigure}[b]{0.45\textwidth}
    \includegraphics[width=\textwidth]{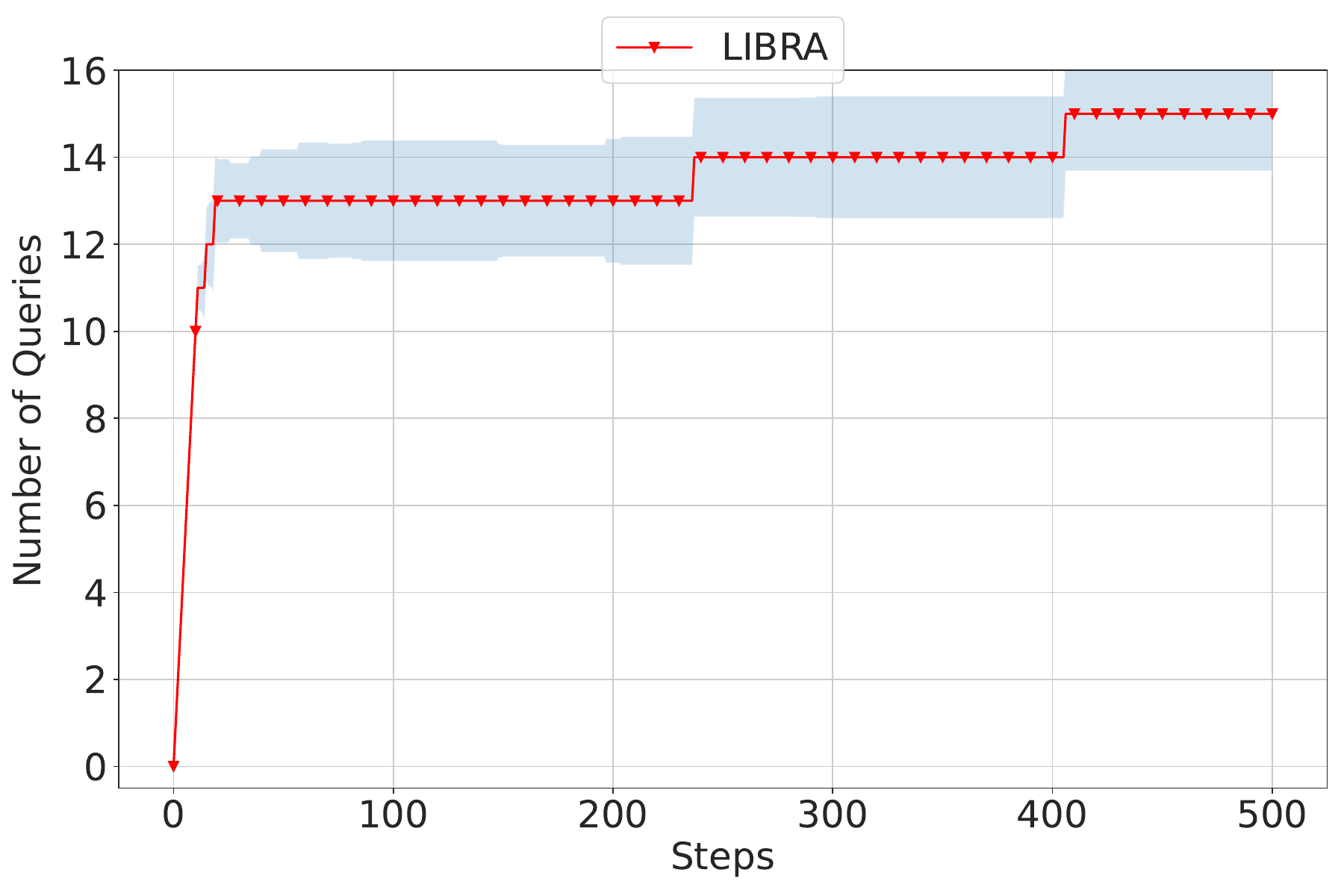}
    \caption{The Number of LLM Queries by LIBRA.}
    \label{fig:case_human}
  \end{subfigure}
  \caption{Performance evaluation of different algorithms in terms of cumulative recourse regret and number of LLM queries for patients with hypertension from the ACCORD dataset.}
  \label{fig:case}
\end{figure}

Figure~\ref{fig:case_regret} plots the cumulative recourse regret over $T=500$ patient interactions, while Figure~\ref{fig:case_human} tracks the cumulative number of times our \textsf{LIBRA} framework queried the LLM. 
As shown in Figure~\ref{fig:case_regret}, the standard LinUCB incurs the highest regret, which grows linearly over time. This behavior is expected, as LinUCB is fundamentally unable to optimize for recourses and thus fails to exploit modifiable patient characteristics. The LLM-only baseline also exhibits linear regret, though lower than LinUCB, indicating that the LLM's general advice provides some value but fails to adapt or learn from patient feedback. Notably, the LLM-only shows substantially lower variance than the bandit algorithms with no LLMs involved, because we only query once with no variation in the LLM response to each patient. 
In contrast, \textsf{GLRB} achieves  sublinear regret, demonstrating its ability to effectively learn and optimize recourse-treatment recommendations over time. This confirms the value of explicitly modeling lifestyle recourses within a principled bandit framework.
Most notably, \textsf{LIBRA} achieves the lowest regret by a significant margin. Its regret curve flattens almost immediately, showcasing the practical impact of the {\em warm-start guarantee} (\Cref{thm: warm start guarantee}). By leveraging the LLM's prior knowledge during early, high-uncertainty round, \textsf{LIBRA} avoids the costly initial exploration phase that \textsf{GLRB} must undergo. As uncertainty decreases, \textsf{LIBRA} rapidly transitions to data-driven decision-making, combining fast early gains with strong long-run performance.

Figure~\ref{fig:case_human} empirically validates our {\em LLM-effort guarantee} (\Cref{thm: LLM-effort}). The number of LLM queries increases sharply during the initial rounds where uncertainty is high, but then declines quickly as learning progresses. This pattern shows that once \textsf{LIBRA} has collected sufficient data and accordingly its own uncertainty (confidence interval) shrinks, it optimally stops relying on the costly LLM queries and transitions to a fully data-driven policy.
Importantly, this behavior arises naturally from the algorithm’s decision rule, without requiring any manual tuning, and directly reflects our theoretical development that LLM interaction is needed only during a finite initial learning phase.

\subsubsection*{Clinical Insights (Patient Outcomes and Recourse Personalization):}
Beyond the regret performance, we next examine the practical clinical impact of our proposed algorithms, measured by the average reduction in SBP. Table~\ref{tab:case_bp} shows that \textsf{LIBRA} achieves the largest average SBP reduction ($51.14$ mmHg), outperforming all other methods and indicating the most favorable clinical outcomes for patients.
\begin{table}[htp]
    \centering
    \caption{Average Systolic Blood Pressure Dropped from the target of 170.}
    \scalebox{0.86}{
    \begin{tabular}{ccccc} \toprule
        &  LLM & LinUCB & GLRB& LIBRA\\ \hline 
        Average SBP Dropped from 170 & 45.84$\pm$ 0.37 & 44.45 $\pm$ 0.28 & 48.28 $\pm$ 2.45 & 51.14$\pm$0.50\\ \bottomrule
    \end{tabular}}
    \label{tab:case_bp}
\end{table}
\vspace{-0.1cm}

Notably, while both \textsf{GLRB} and  LLM-only improve upon LinUCB, their gains are more modest. \textsf{GLRB} benefits from learning personalized lifestyle recourses over time, whereas  LLM-only relies solely on static, general medical knowledge. In contrast, our \textsf{LIBRA} combines the strengths of both: it leverages clinically informed guidance early on and then refines recommendations via patient-specific learning. This synergy enables \textsf{LIBRA} not only to accelerate improvement but also to achieve superior long-run SBP control.

\begin{figure}[htp]
  \centering
  \begin{subfigure}[b]{0.45\textwidth}
    \includegraphics[width=\textwidth]{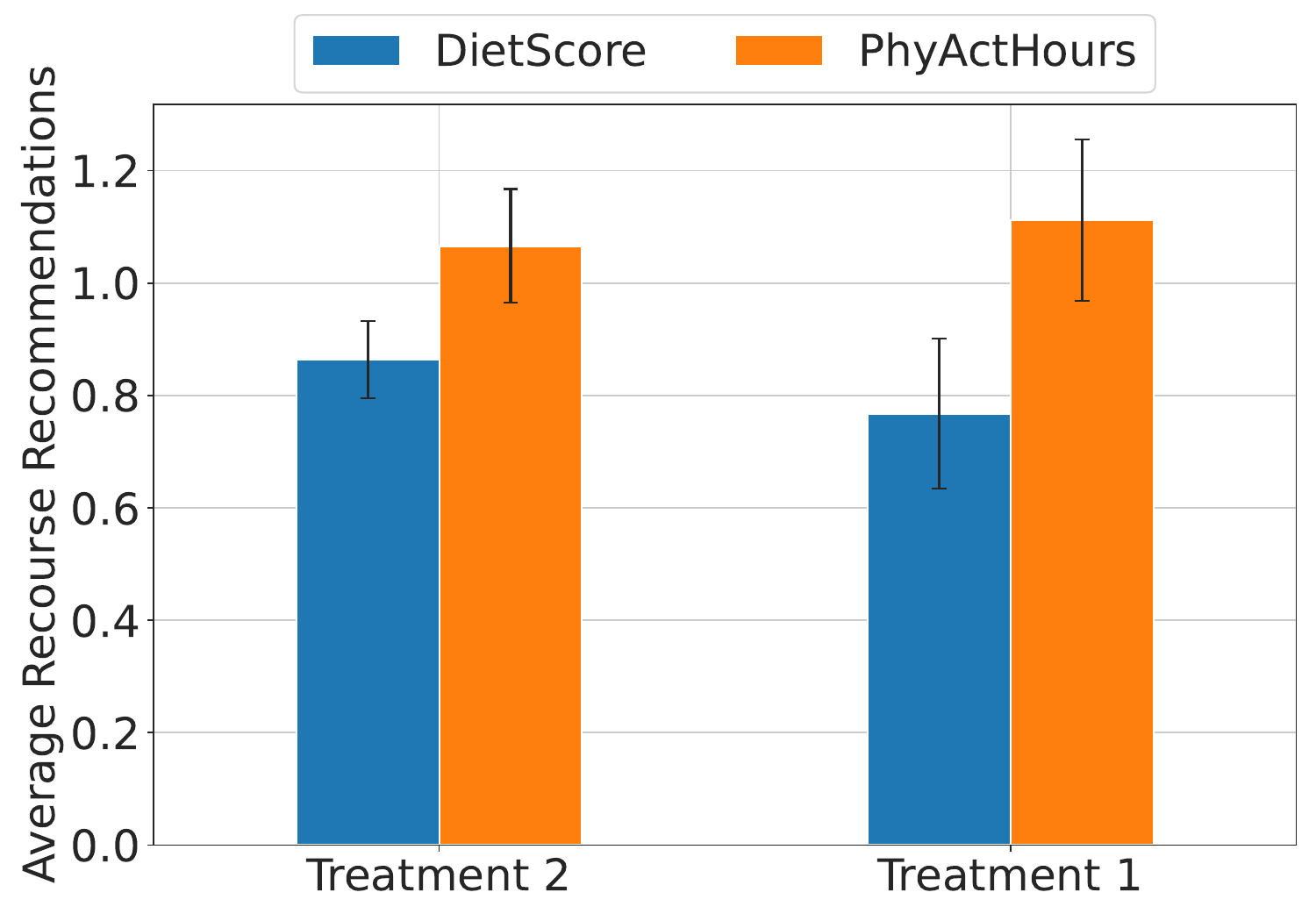}
    \caption{LLM's Average Recourse Recommendation.}
    \label{fig:llm_recourse}
  \end{subfigure}
  \hfill 
  \begin{subfigure}[b]{0.45\textwidth}
    \includegraphics[width=\textwidth]{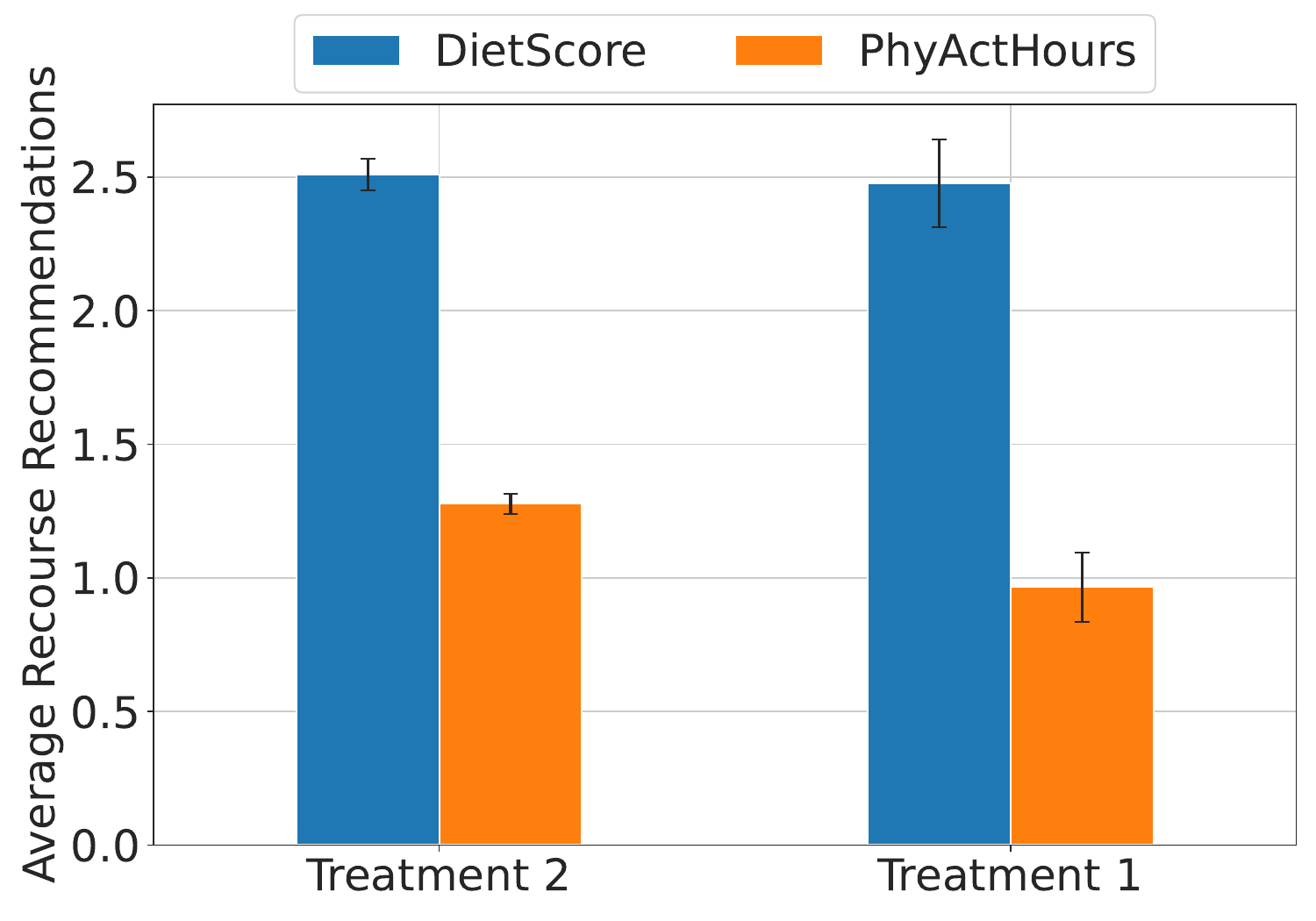}
    \caption{LIBRA Suggested Recourses.}
    \label{fig:libra_recourses}
  \end{subfigure}
  \caption{Recommended Recourses in terms of DietScore and PhyActHours Suggested by LLM and LIBRA.}
  \label{fig:case_recourse}
\end{figure}

Furthermore, Figure~\ref{fig:case_recourse} presents the qualitative results for our case study.  \Cref{fig:llm_recourse} shows the LLM's average recourse recommendation. The LLM suggests a larger increase in \texttt{PhyActHours} (approx. 1.1) than in \texttt{DietScore} (approx. 0.8). This reflects general public health knowledge that emphasizes the importance of both diet and physical activity. In contrast, Figure~\ref{fig:libra_recourses} shows that \textsf{LIBRA} learns a substantially more nuanced and effective strategy. %
Based on the fitted model coefficients (see Table \ref{tab:coefficients}), \texttt{DietScore} is between two and four times more impactful for lowering SBP than \texttt{PhyActHours}, depending on the treatment arm (Treatment 1: \texttt{DietScore} -2.12 vs. \texttt{PhyActHours} -0.48; Treatment 2: \texttt{DietScore} -2.18 vs. \texttt{PhyActHours} -1.02). Lacking access to these treatment-specific effect sizes, the LLM provides reasonable but ultimately suboptimal recommendations.

However, \textsf{LIBRA} uncovers these relationships directly from the data. 
For both treatments, it recommends a substantial increase in the most effective recourse: \texttt{DietScore} (approx. 2.5). Concurrently, it personalizes recommendations for \texttt{PhyActHours}, suggesting a larger increase under Treatment 2 (where physical activity is moderately effective) and a smaller increase under Treatment 1 (where it is least effective). This behavior highlights \textsf{LIBRA}'s central advantage: it successfully leverages the LLM's initial guidance while rapidly learning which lifestyle modifications provide the greatest clinical impact under each treatment. As a result, \textsf{LIBRA} delivers more targeted, personalized recourse recommendations that translate into improved patient outcomes. Clinically, these findings suggest that effective hypertension management may benefit from prioritizing the most impactful lifestyle changes rather than uniformly emphasizing all behaviors.

\subsection{Non-Compliance Patient Behaviors}
\begin{figure}[htp]
  \centering
  \begin{subfigure}[b]{0.45\textwidth}
    \includegraphics[width=\textwidth]{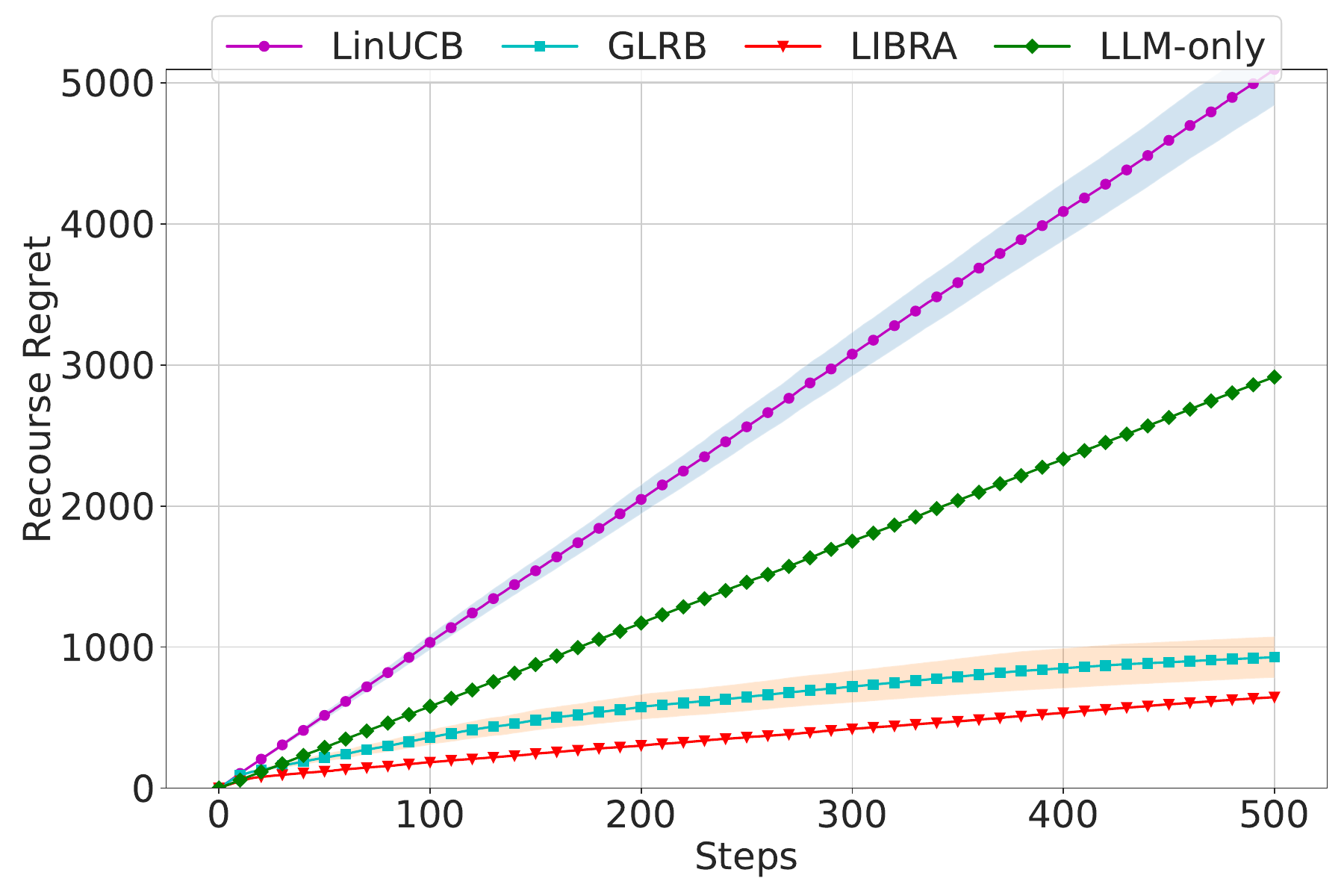}
    \caption{Regret (random non-compliance).}
    \label{fig:randnoncomp_recourse}
  \end{subfigure}
  \hfill 
  \begin{subfigure}[b]{0.45\textwidth}
    \includegraphics[width=\textwidth]{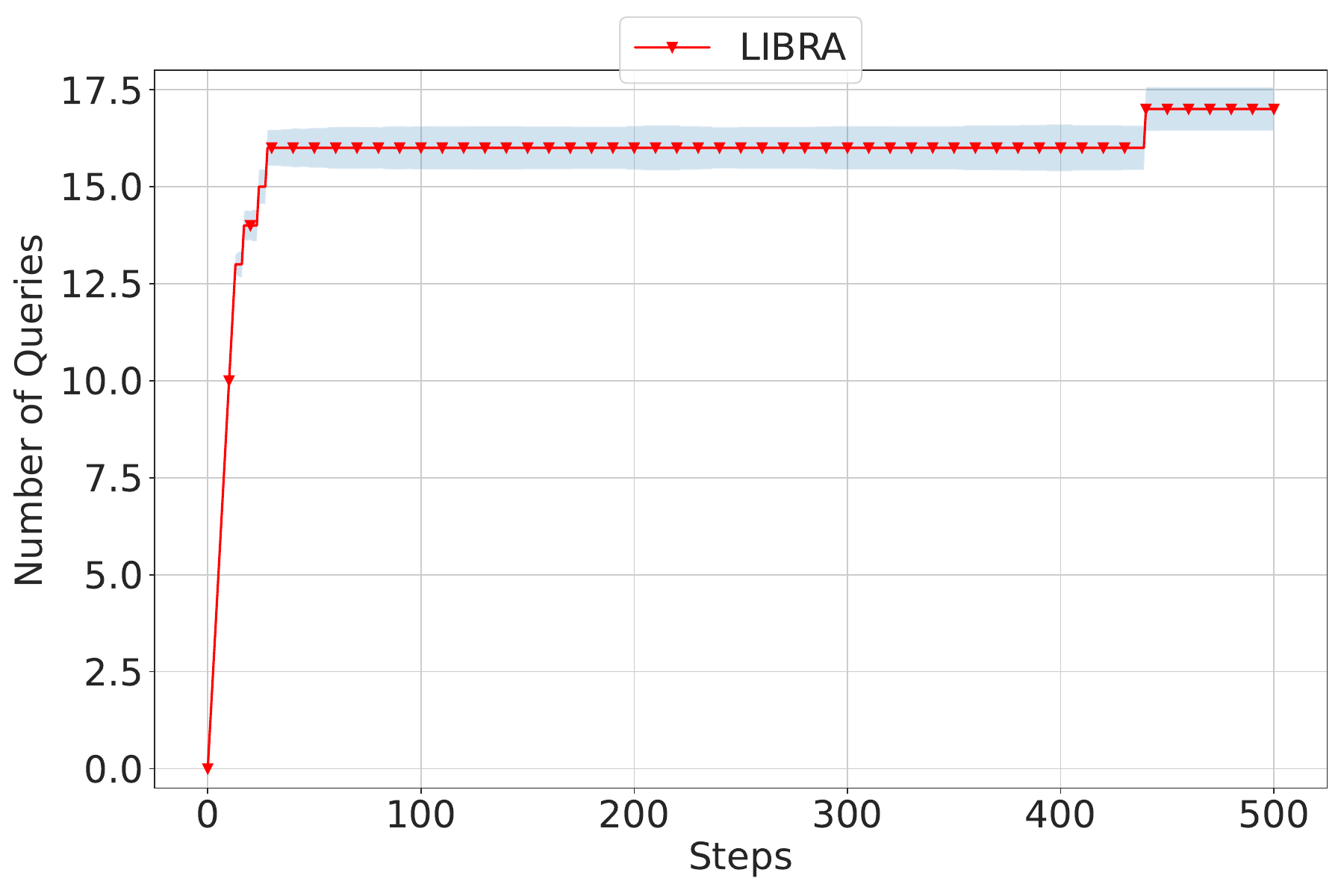}
    \caption{LLM Queries (random non-compliance).}
    \label{fig:randnoncomp_recourses}
  \end{subfigure} \\
  \begin{subfigure}[b]{0.45\textwidth}
    \includegraphics[width=\textwidth]{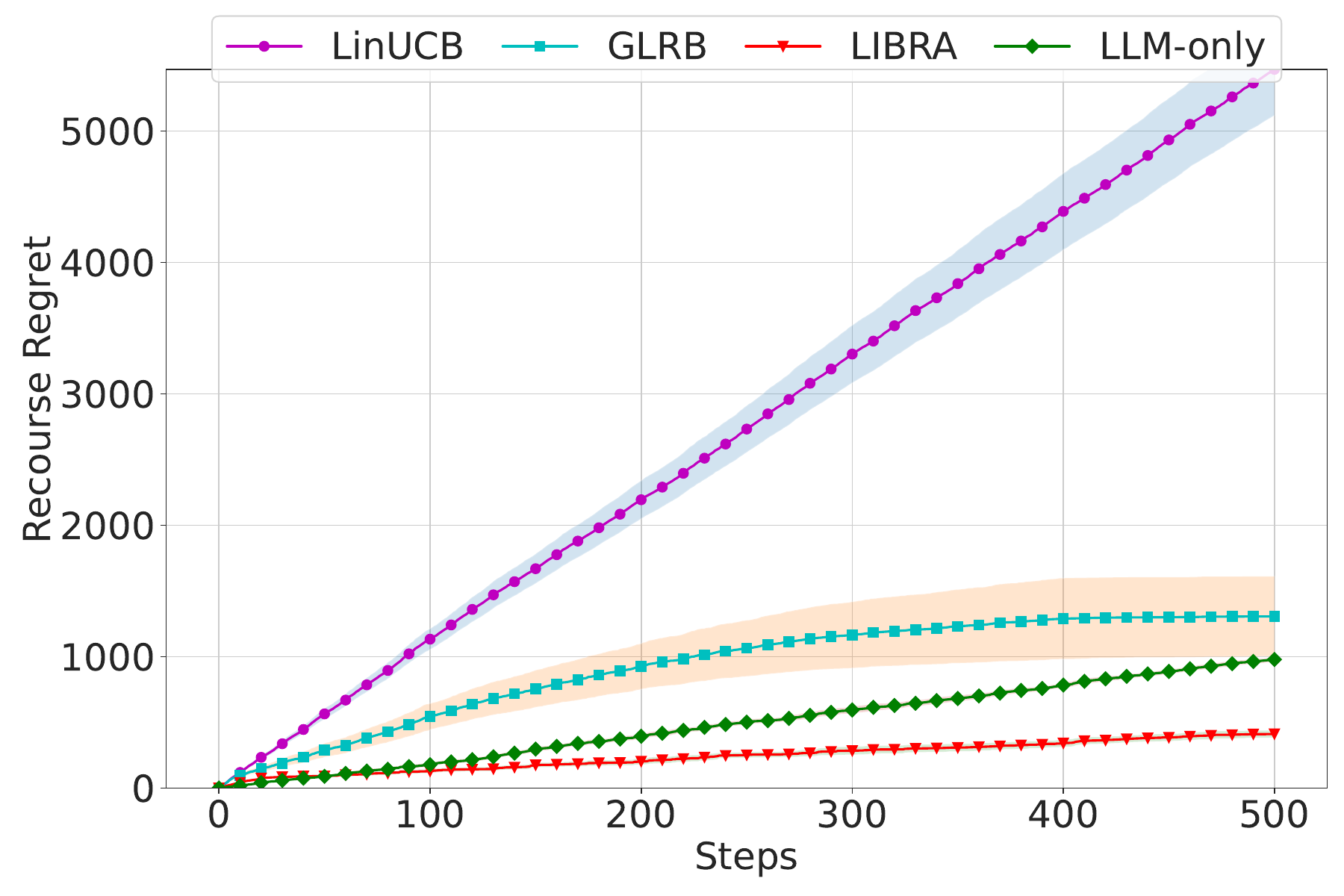}
    \caption{Regret (adversarial non-compliance).}
    \label{fig:advnoncomp_recourse}
  \end{subfigure}
  \hfill 
  \begin{subfigure}[b]{0.45\textwidth}
    \includegraphics[width=\textwidth]{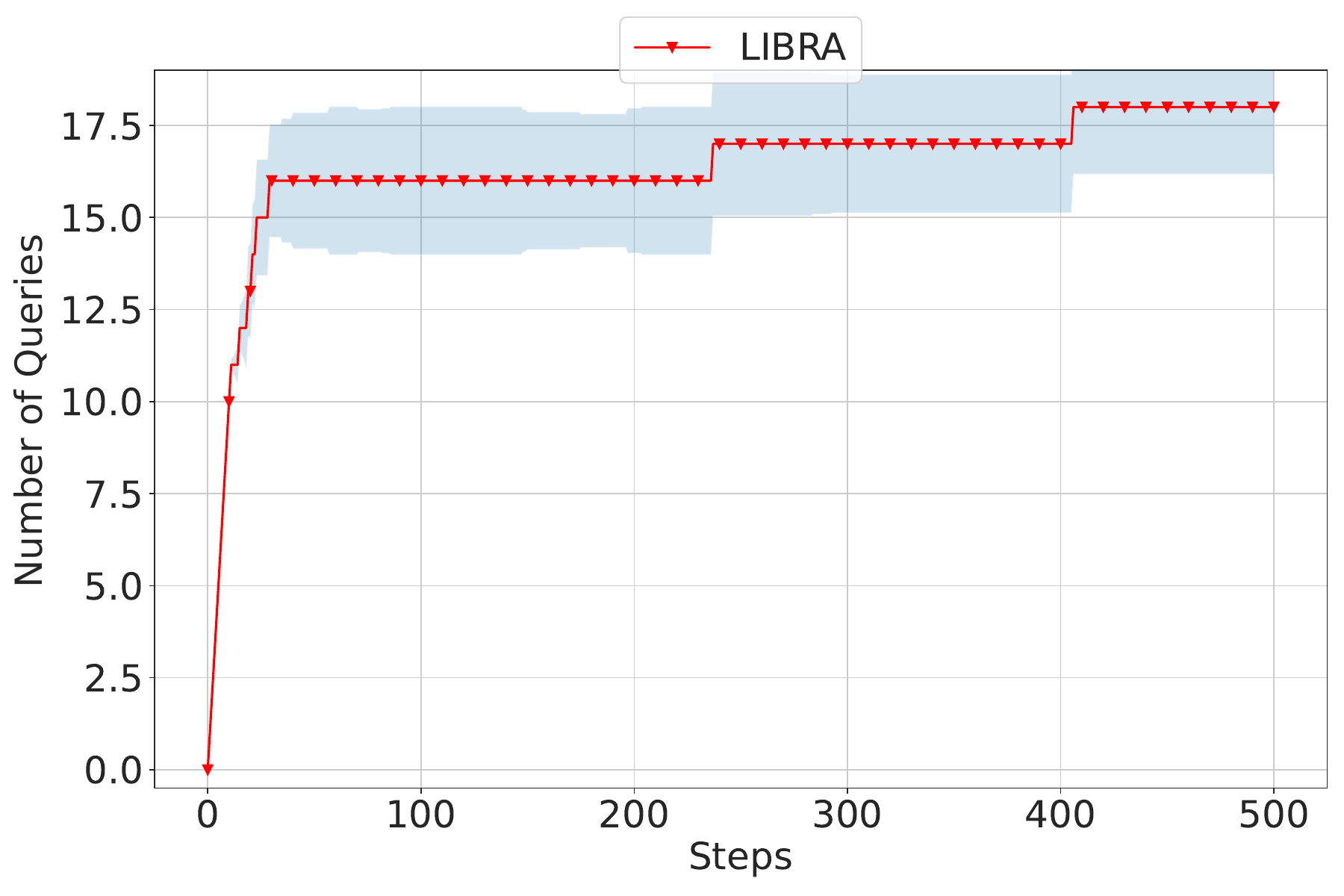}
    \caption{LLM Queries (adversarial non-compliance).}
    \label{fig:advnoncomp_recourses}
  \end{subfigure}
  \caption{Recourses Under Non-Compliance Behaviors under both random and adversarial settings.}
  \label{fig:noncomp_recourse}
\end{figure}

We next examine how the proposed algorithms perform under patient non-compliance behaviors. 
\Cref{fig:randnoncomp_recourse} and \Cref{fig:randnoncomp_recourses} show the cumulative regret and the number of queries under {\em random} non-compliance behaviors, while \Cref{fig:advnoncomp_recourse} and \Cref{fig:advnoncomp_recourses} present the corresponding results under {\em adversarial} non-compliance behaviors. The results are consistent with our previous findings that \textsf{LIBRA} provides the best model performance under a limited number of LLM queries. This demonstrates that the benefits of LLM-guided warm starts and adaptive learning persist even when patient behavior deviates from prescribed recourses. 

\subsection{Robustness Analysis with Different LLMs}

To assess the robustness of our findings and insights across different LLMs, we further examine the proposed method with GPT-4o-mini and GPT-5.2. For all experiments, we use the same system prompt as in \Cref{fig:promptbox} and consider the full-compliance setting to isolate the effect of the underlying LLM. 

\begin{figure}[htp]
  \centering
  \begin{subfigure}[b]{0.45\textwidth}
    \includegraphics[width=\textwidth]{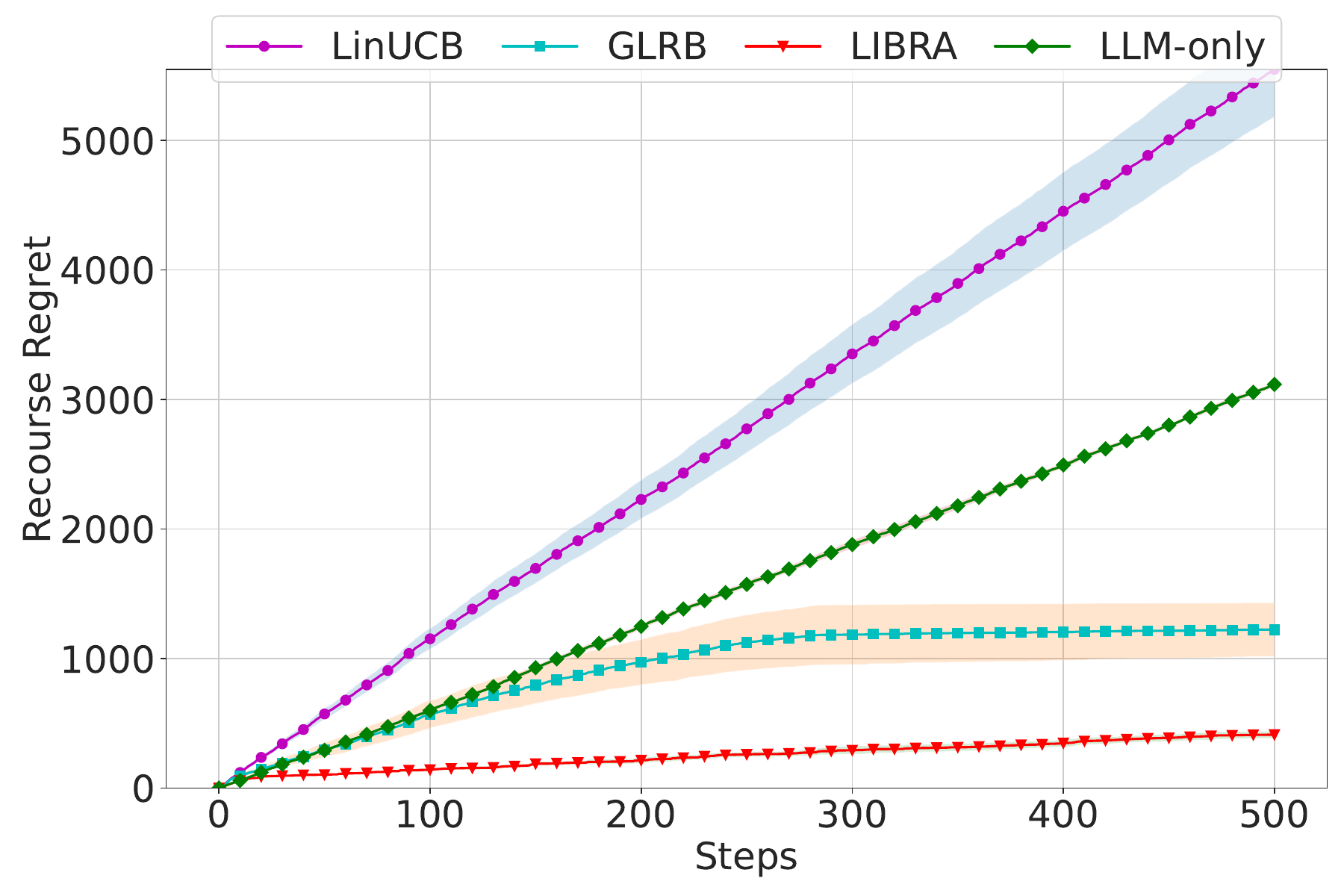}
    \caption{Regret (GPT-4o-mini).}
    \label{fig:regret_4omini}
  \end{subfigure}
  \hfill 
  \begin{subfigure}[b]{0.45\textwidth}
    \includegraphics[width=\textwidth]{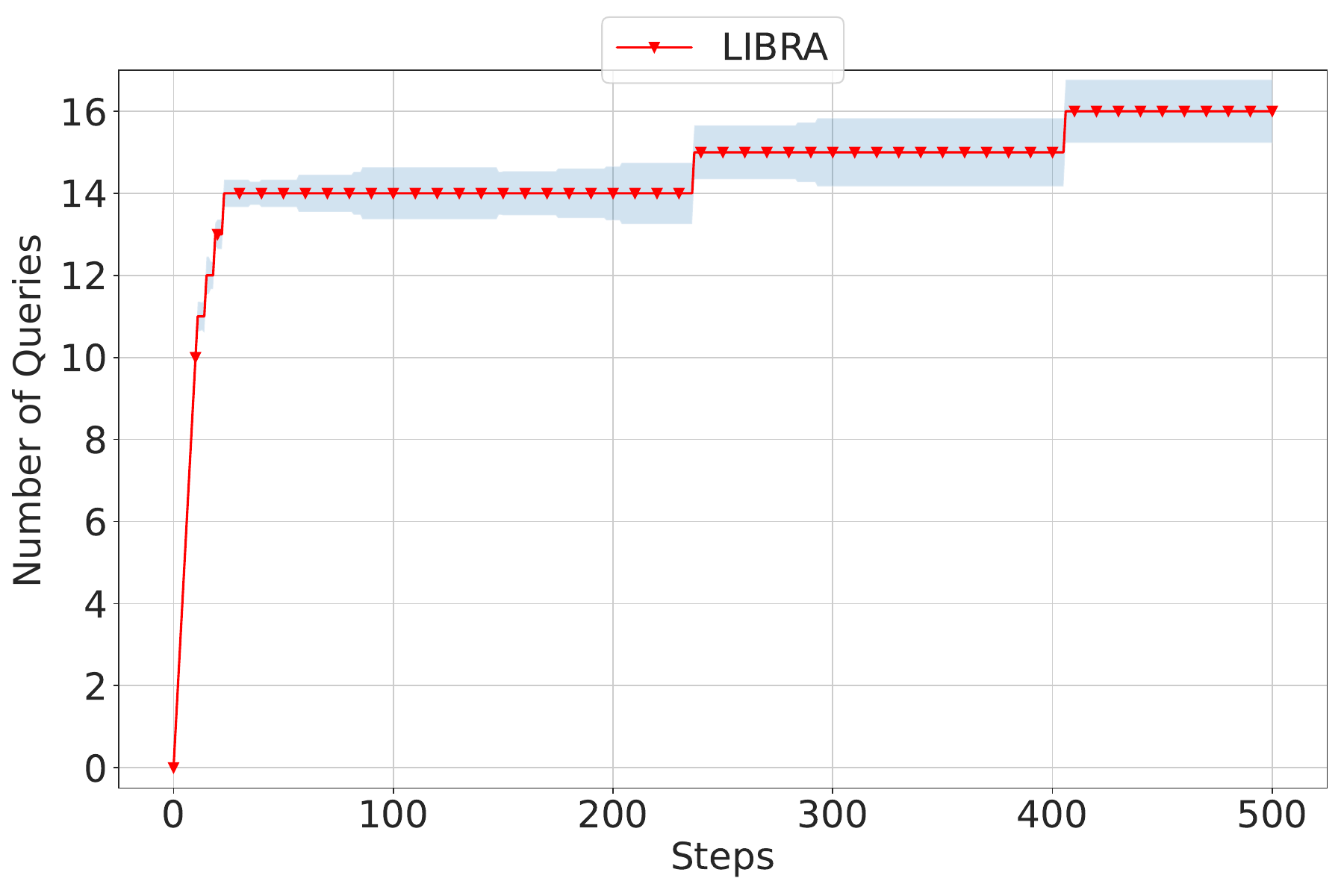}
    \caption{LLM Queries (GPT-4o-mini).}
    \label{fig:asked_4omini}
  \end{subfigure} \\
  \begin{subfigure}[b]{0.45\textwidth}
    \includegraphics[width=\textwidth]{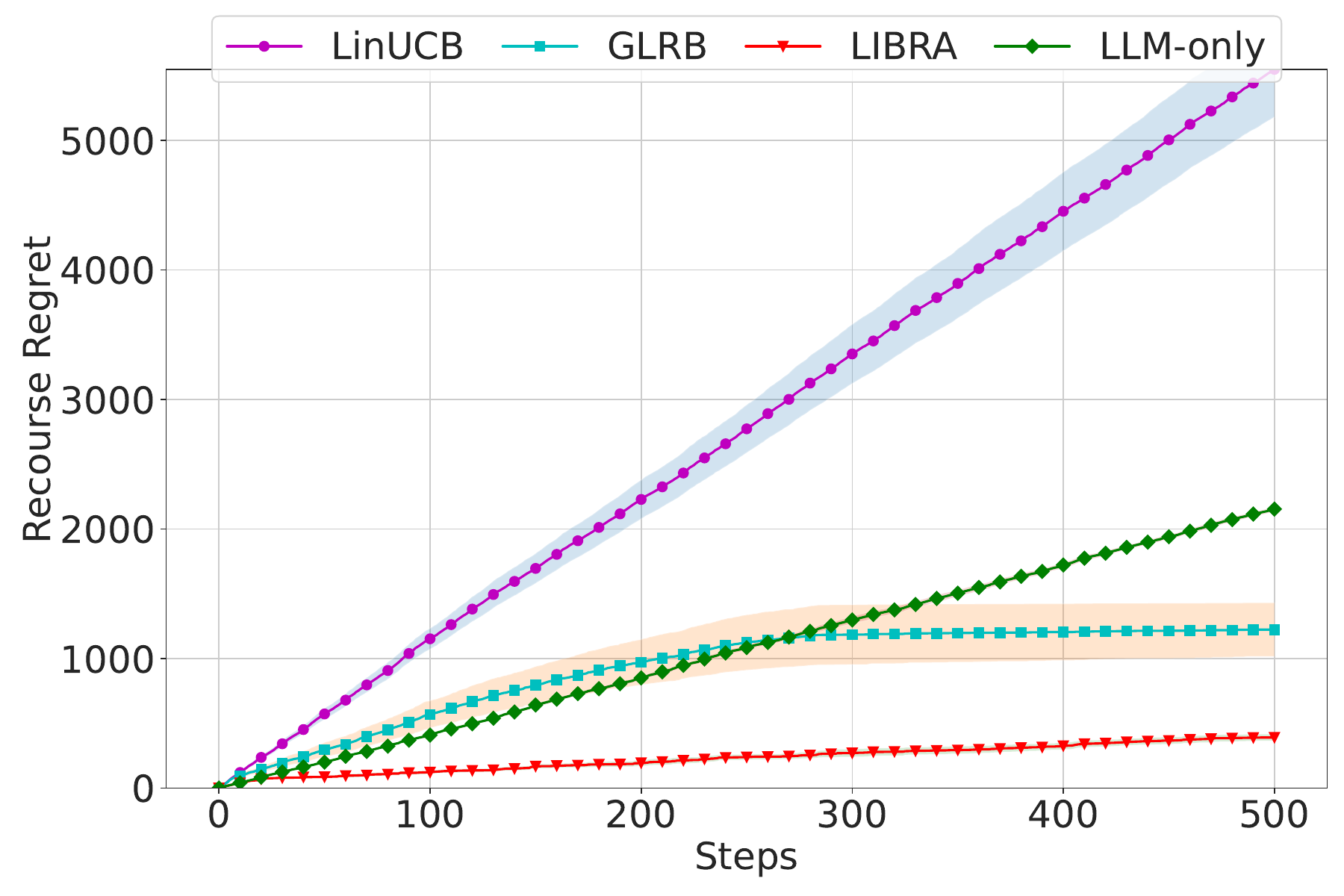}
    \caption{Regret (GPT-5.2).}
    \label{fig:regret52}
  \end{subfigure}
  \hfill 
  \begin{subfigure}[b]{0.45\textwidth}
    \includegraphics[width=\textwidth]{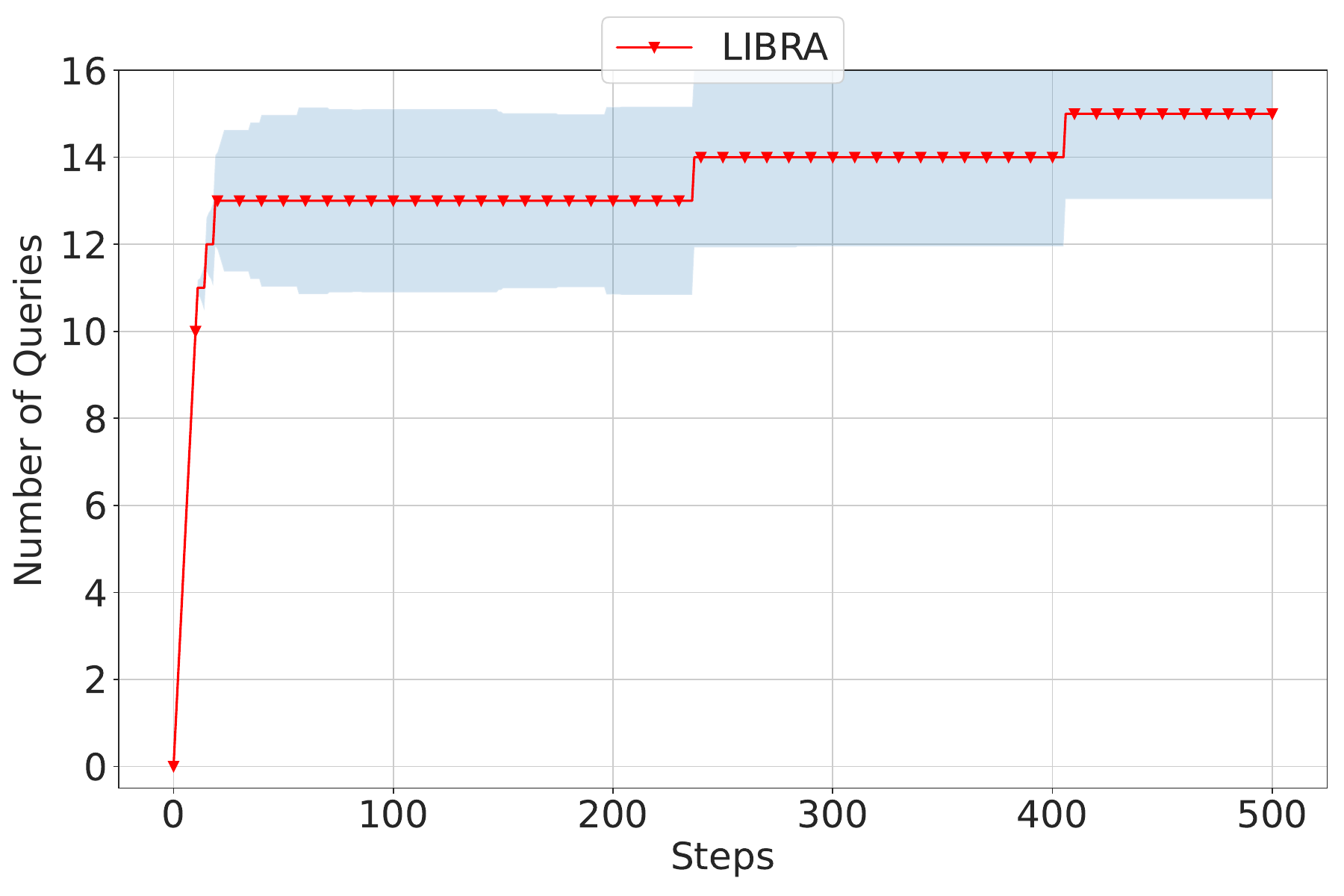}
    \caption{LLM Queries (GPT-5.2).}
    \label{fig:asked52}
  \end{subfigure}
  \caption{Results with Different LLMs.}
  \label{fig:llms}
\end{figure}

The resulting cumulative regret and LLM queries are reported in \Cref{fig:llms}. The overall patterns are consistent with the main results in the previous section. Across all tested models, \textsf{LIBRA} successfully leverages LLM prior knowledge to achieve sublinear regret while requiring only a limited number of LLM queries. The LLM Queries performance between GPT-4o-mini and GPT-5.2 is very close with only slight differences at the beginning. These findings indicate that our results are robust to the choice of underlying LLM, including state-of-the-art models, and that \textsf{LIBRA} provides a stable and effective mechanism for integrating powerful, but inherently imperfect generative priors with the adaptive, data-driven learning of bandit algorithms.

\begin{figure}[htp]
    \centering
    \includegraphics[width=0.5\linewidth]{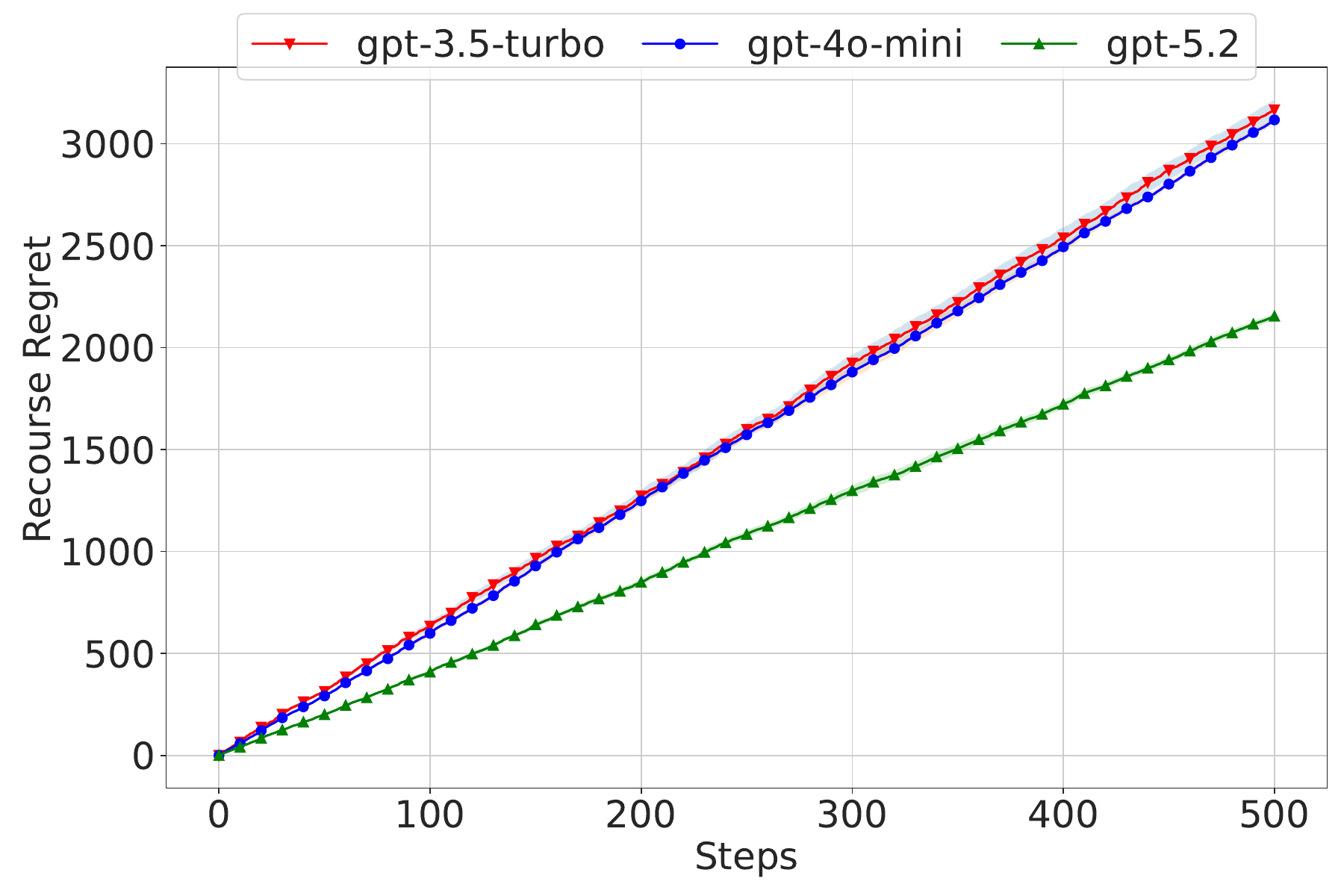}
    \caption{Regret Comparisons for Different LLMs.}
    \label{fig:regret_compare}
\end{figure}

As another interesting investigation, we further examine whether performance differs across LLM versions. We test the regret for GPT-3.5, GPT-4o-mini, and GPT-5.2 in \Cref{fig:regret_compare}. Interestingly, we find that the more advanced version of LLM indeed achieves lower regret, indicating that the more recent LLM does have an increasing knowledge in healthcare problems, though optimal decision-making remains a challenge. 

\section{Synthetic Data and Experiments} \label{sec:syn}

We further demonstrate the efficacy of the proposed methods on a controlled synthetic dataset. 

\noindent \textbf{LLM Behavior Model:} 
To model the quality of LLM guidance in a controlled and interpretable manner, we characterize the LLM’s decision quality using a parameter $q \in [0,1]$. For each decision instance, the LLM outputs the optimal action–recourse pair with probability $q$, and a random action–recourse pair with probability $1-q$. This abstraction captures the idea that LLMs may provide high-quality but imperfect guidance. 
We set $q=0.8$, $\gamma=3$, and $\Delta=1$ in the experiments. All experimental results are averaged over 10 independent runs. The sensitivity of the results to the hyperparameters is examined in \Cref{sec:ablation}.

In addition to the fully synthetic setting, we conduct experiments on a semi-synthetic benchmark based on another real-world dataset with simulated counterfactual outcomes in Appendix \ref{app:semi-synthetic}. The results are consistent with the findings reported in this section, further validating the robustness of our insights.

\subsection{Synthetic Data}

We generate patient contexts by sampling $x \sim \mathcal{N}(0,I_d)$, where $d = 20$ and $I_d$ is a $d$-dimensional identity matrix. We assume that all features are mutable for simplicity and clarity. The reward is generated by a linear model $r(a,x) = \theta_a^\top x + \epsilon$, where $\epsilon \sim \mathcal{N}(0,1)$ represents stochastic noise, $a\in\{1,2,\cdots, 10\}$ denotes the action, and the action-specific parameters satisfy $\theta_a \sim \mathcal{N}(0,I_d)$. At each time step, the learner must jointly decide which action to take and what recourse to recommend for the arriving patient with context $x_t$. Following most algorithmic recourse papers \citep{karimi2020survey,gao2023impact}, we assume that patients will adhere to the recourse recommendations that satisfy the distance constraint. 

\begin{figure}[htp]
  \centering
  \begin{subfigure}[b]{0.45\textwidth}
    \includegraphics[width=\textwidth]{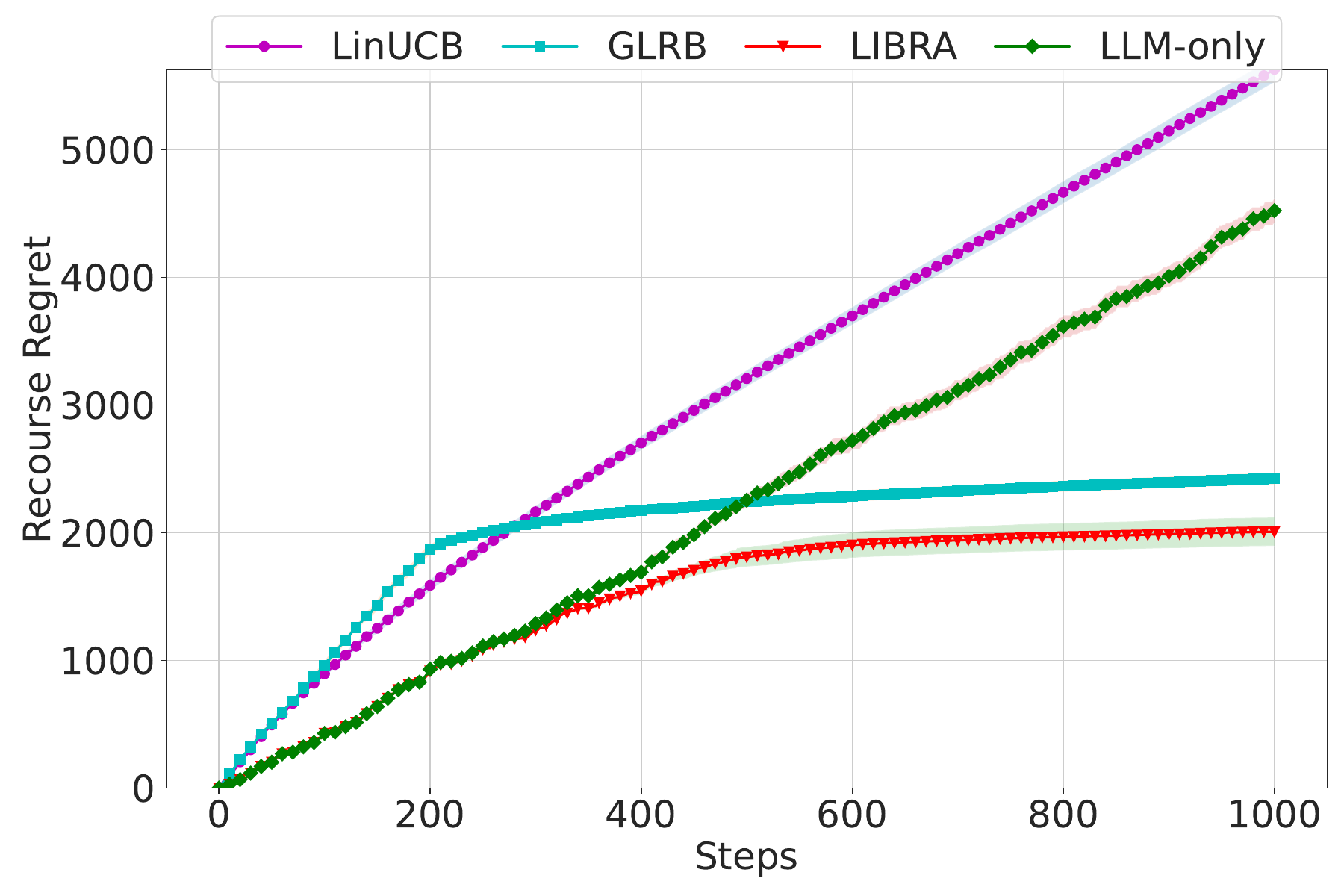}
    \caption{Cumulative Recourse Regret.}
    \label{fig:syn_regret}
  \end{subfigure}
  \hfill 
  \begin{subfigure}[b]{0.45\textwidth}
    \includegraphics[width=\textwidth]{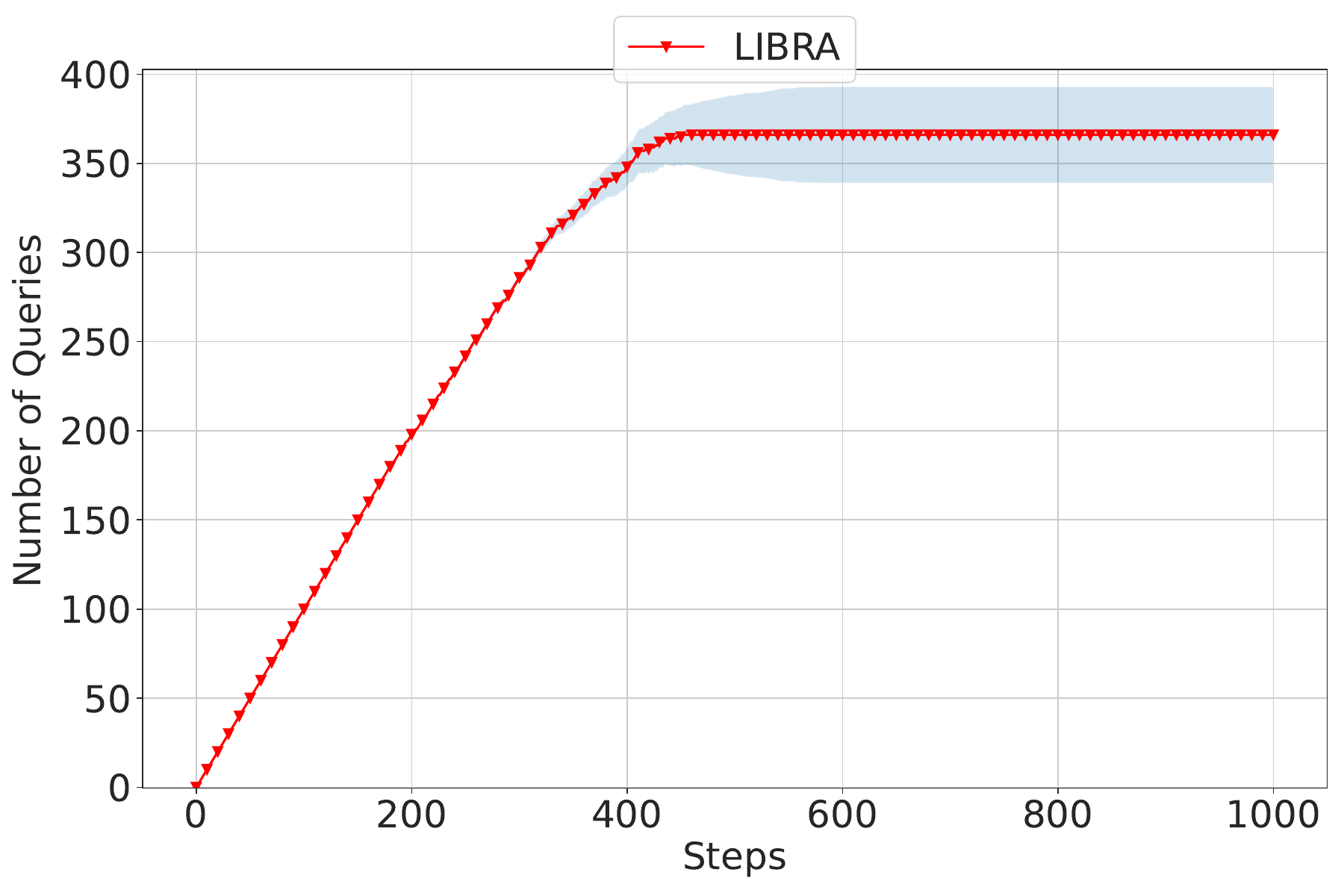}
    \caption{LLM Queries.}
    \label{fig:syn_human}
  \end{subfigure}
  \caption{Performance evaluation of different algorithms in terms of cumulative recourse regret and number of LLM queries using synthetic Data.}
  \label{fig:syn}
\end{figure}

The cumulative recourse regret and the number of LLM queries are shown in \Cref{fig:syn_regret} and \Cref{fig:syn_human}, respectively. 
As expected, LinUCB exhibits linear recourse regret, since it does not model or optimize recourse decisions. In contrast, \textsf{GLRB} substantially improves upon LinUCB and achieves sublinear regret, consistent with our theoretical findings in \Cref{thm: regretbound-GLRB}. This behavior indicates that \textsf{GLRB} successfully learns to prescribe increasingly effective recourses over time. Notably, \textsf{LIBRA} attains the lowest recourse regret among all benchmarks by selectively leveraging guidance from LLM experts, empirically validating the improvement guarantee in \Cref{thm: LIBRA improvementguarantee}. At the same time, Figure~\ref{fig:syn_human} shows that \textsf{LIBRA} relies on only a limited number of LLM queries, providing empirical support for the LLM-effort guarantee in \Cref{thm: LLM-effort}. Together, these results demonstrate that \textsf{LIBRA} effectively balances external guidance and data-driven learning—achieving strong performance gains without sustained dependence on LLM input.

\subsection{Ablation Studies}
\label{sec:ablation}

\begin{figure}[htp]
  \centering
  \begin{subfigure}[b]{0.45\textwidth}
    \includegraphics[width=\textwidth]{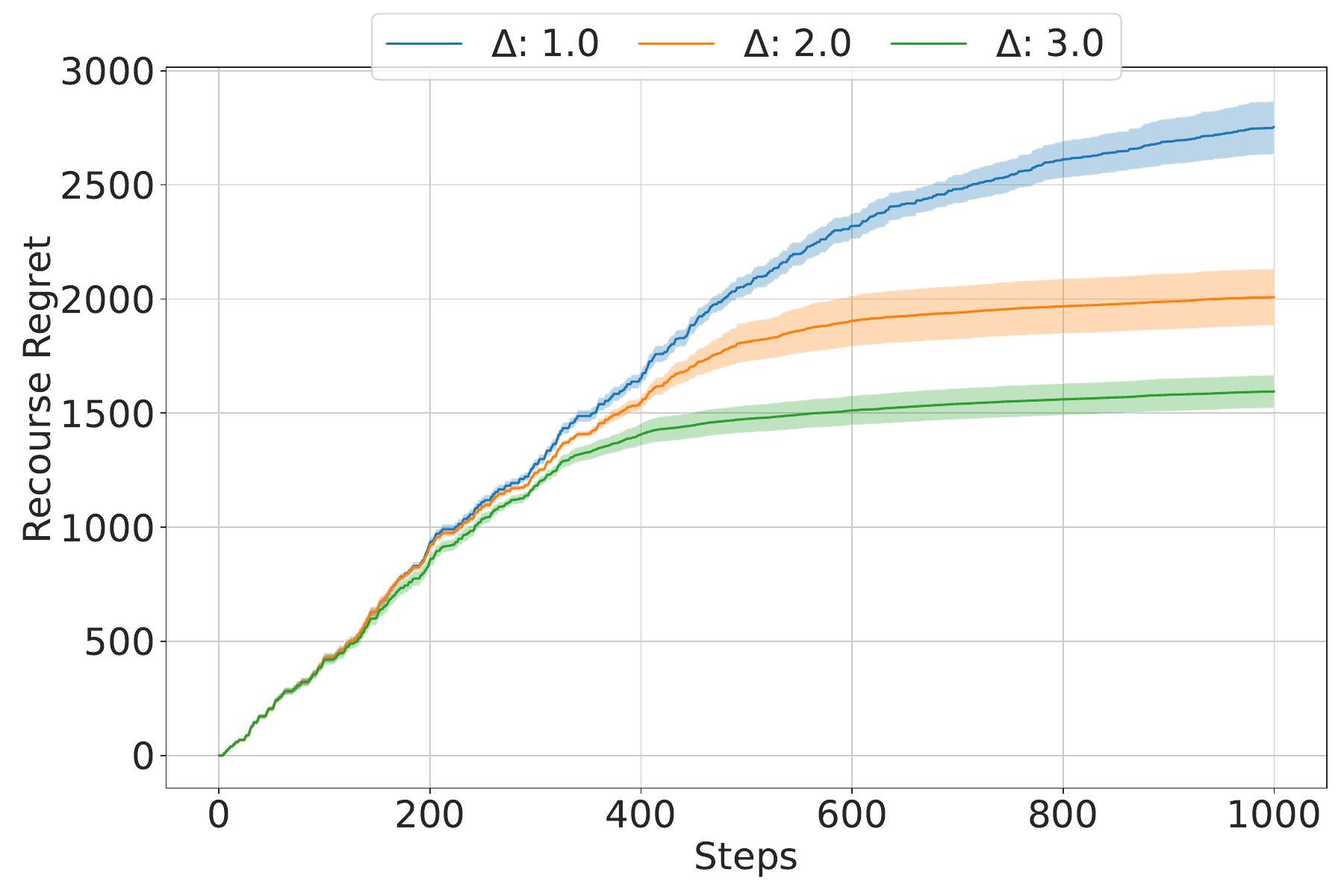}
    \caption{Sensitivity analysis on parameter $\Delta$.}
    \label{fig:delta_regret}
  \end{subfigure}
  \hfill 
  \begin{subfigure}[b]{0.45\textwidth}
    \includegraphics[width=\textwidth]{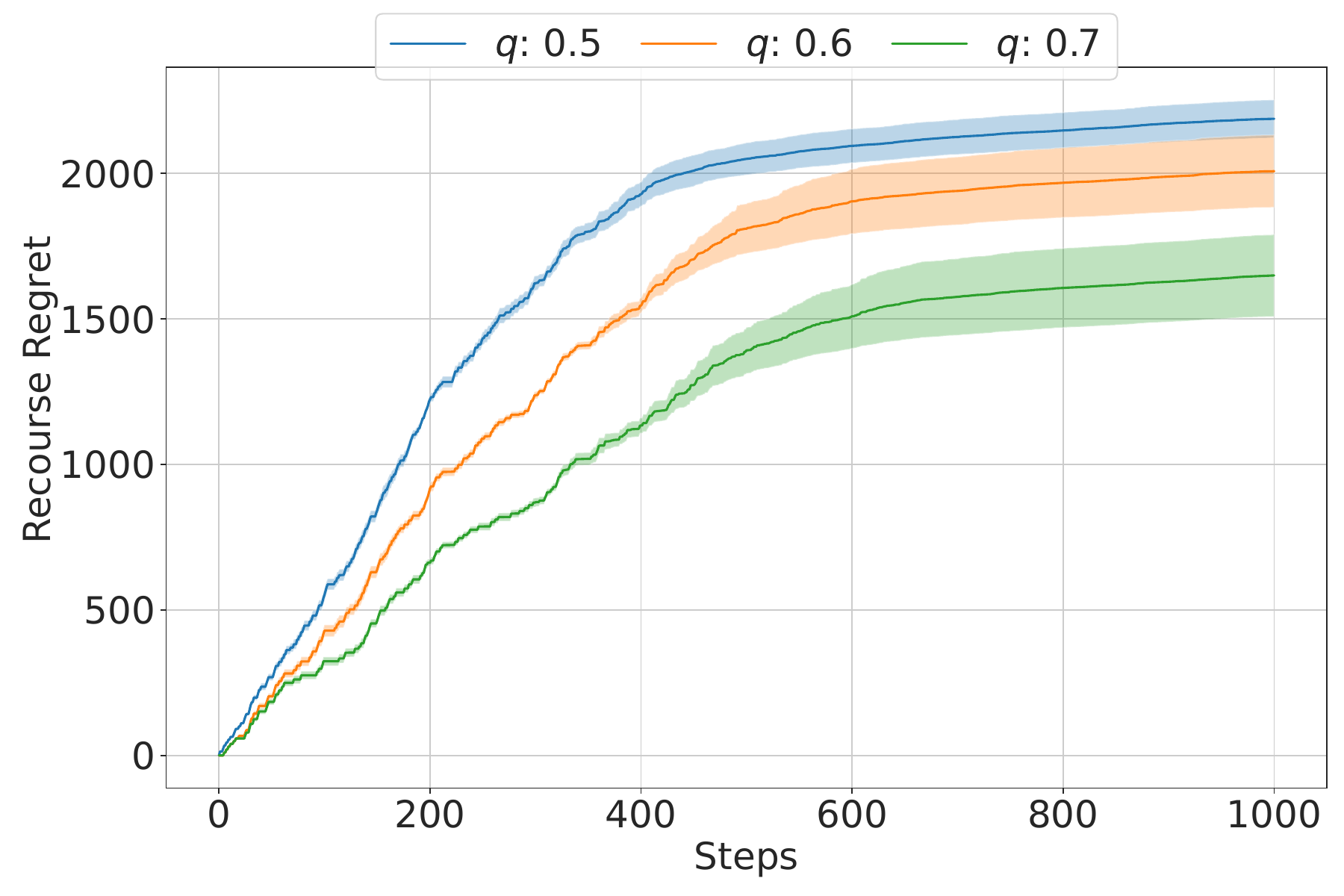}
    \caption{Sensitivity analysis on parameter $q$.}
    \label{fig:q_regret}
  \end{subfigure}
  \caption{Empirical evaluations on the impacts of $\Delta$ and $q$ parameters on the recourse regret bound.}
  \label{fig:ablation}
\end{figure}

We examine the sensitivity of \textsf{LIBRA} to two design parameters: the confidence threshold $\Delta$ and the LLM quality $q$. Figures~\ref{fig:delta_regret} and \ref{fig:q_regret} report the resulting recourse regret as each parameter is varied. The threshold $\Delta$ controls how frequently \textsf{LIBRA} consults the LLM—smaller values lead to more frequent reliance on LLM guidance, while larger values encourage earlier transition to autonomous, data-driven decision-making.

Empirically, \textsf{LIBRA} exhibits strong robustness, achieving sublinear recourse regret across all values of $\Delta$. However, extreme choices of $\Delta$ are suboptimal. When $\Delta$ is very small, \textsf{LIBRA} relies too heavily on LLM guidance and effectively behaves like the LLM-only baseline. Conversely, when $\Delta$ is large, \textsf{LIBRA} rarely queries the LLM and reduces to the purely data-driven \textsf{GLRB} baseline. Performance is maximized when $\Delta$ lies in a moderate range, where \textsf{LIBRA} successfully balances external prior knowledge with online learning, outperforming both individual baselines. This behavior empirically validates our theoretical guarantees and highlights the value of integrating imperfect but informative LLM priors with principled data-driven algorithms. Additionally, our results show that higher LLM quality consistently leads to lower regret, confirming that \textsf{LIBRA} can effectively capitalize on improvements in generative models while retaining robustness when such guidance is imperfect.

\section{Conclusion and Future Work}

In this work, we introduced \textsf{GLRB} and \textsf{LIBRA}, extending algorithmic recourse and LLM-Bandits collaboration frameworks to online learning with rigorous theoretical guarantees. \textsf{LIBRA} offers three key theoretical properties: (i) improvement guarantee, (ii) LLM-effort guarantee, (iii) robustness guarantee, and (iv) near-optimality guarantee. Empirical results show its superior performance over standard bandit algorithms. Our experimental results, conducted on both synthetic environments and a real-world hypertension management case study, demonstrate that \textsf{LIBRA} consistently outperforms traditional contextual bandits and LLM-only benchmarks in terms of regret and treatment quality. These findings highlight the potential of recourse-aware, LLM-assisted systems for creating more effective and personalized clinical interventions.

While this study provides a rigorous foundation for recourse bandits, several avenues for future research remain. First, while we assumed a generalized linear structure, exploring non-linear or non-parametric functions could broaden applicability to more complex biological systems. Second, although we addressed random and adversarial non-compliance, further modeling of stochastic patient behaviors would better capture the complexities of real-world adherence.
Furthermore, the emergence of diverse foundation models suggests a transition toward multi-LLM architectures. Since different LLMs may possess varying expertise—for instance, one model specialized in clinical pharmacology and another in behavioral psychology for lifestyle recourse—how to adaptively select the right model for a specific patient remains a critical question. Future research could investigate ``meta-bandit" gating mechanisms that learn to route queries to the most reliable expert for specific patient subgroups, further enhancing the personalization and cost-efficiency of the treatment-recourse pipeline. Finally, extending this framework to multi-objective settings by balancing treatment effectiveness, recourse cost, and patient preference, is an important step toward truly human-centered AI in medicine.

\def\bibfont{\scriptsize}
\bibliographystyle{informs2014}
\bibliography{references}

\begin{thebibliography}{85}
\providecommand{\natexlab}[1]{#1}
\providecommand{\url}[1]{\texttt{#1}}
\providecommand{\urlprefix}{URL }

\bibitem[{Abbasi-Yadkori et~al.(2011)Abbasi-Yadkori, P{\'a}l, \protect\BIBand{} Szepesv{\'a}ri}]{abbasi2011improved}
Abbasi-Yadkori Y, P{\'a}l D, Szepesv{\'a}ri C (2011) Improved algorithms for linear stochastic bandits. \emph{Advances in neural information processing systems} 24.

\bibitem[{Abeille \protect\BIBand{} Lazaric(2017)}]{abeille2017linear}
Abeille M, Lazaric A (2017) Linear thompson sampling revisited. \emph{Artificial Intelligence and Statistics}, 176--184 (PMLR).

\bibitem[{ACCORD(2008)}]{action2008effects}
ACCORD (2008) Effects of intensive glucose lowering in type 2 diabetes. \emph{New England journal of medicine} 358(24):2545--2559.

\bibitem[{ACCORD(2010)}]{accord2010effects}
ACCORD (2010) Effects of intensive blood-pressure control in type 2 diabetes mellitus. \emph{New England Journal of Medicine} 362(17):1575--1585.

\bibitem[{Agrawal \protect\BIBand{} Goyal(2013)}]{agrawal2013thompson}
Agrawal S, Goyal N (2013) Thompson sampling for contextual bandits with linear payoffs. \emph{International conference on machine learning}, 127--135 (PMLR).

\bibitem[{Alamdari et~al.(2024)Alamdari, Cao, \protect\BIBand{} Wilson}]{alamdari2024jump}
Alamdari PA, Cao Y, Wilson KH (2024) Jump starting bandits with llm-generated prior knowledge. \emph{arXiv preprint arXiv:2406.19317} .

\bibitem[{Attouch et~al.(2010)Attouch, Bolte, Redont, \protect\BIBand{} Soubeyran}]{attouch2010proximal}
Attouch H, Bolte J, Redont P, Soubeyran A (2010) Proximal alternating minimization and projection methods for nonconvex problems: An approach based on the kurdyka-{\l}ojasiewicz inequality. \emph{Mathematics of operations research} 35(2):438--457.

\bibitem[{Attouch et~al.(2013)Attouch, Bolte, \protect\BIBand{} Svaiter}]{attouch2013convergence}
Attouch H, Bolte J, Svaiter BF (2013) Convergence of descent methods for semi-algebraic and tame problems: proximal algorithms, forward--backward splitting, and regularized gauss--seidel methods. \emph{Mathematical programming} 137(1):91--129.

\bibitem[{Auer(2002)}]{auer2002using}
Auer P (2002) Using confidence bounds for exploitation-exploration trade-offs. \emph{Journal of Machine Learning Research} 3(Nov):397--422.

\bibitem[{Bhaskara et~al.(2023)Bhaskara, Cutkosky, Kumar, \protect\BIBand{} Purohit}]{bhaskara2023bandit}
Bhaskara A, Cutkosky A, Kumar R, Purohit M (2023) Bandit online linear optimization with hints and queries. \emph{International Conference on Machine Learning}, 2313--2336 (PMLR).

\bibitem[{Bienstock(2016)}]{bienstock2016note}
Bienstock D (2016) A note on polynomial solvability of the cdt problem. \emph{SIAM Journal on Optimization} 26(1):488--498.

\bibitem[{Blumenthal et~al.(2021)Blumenthal, Hinderliter, Smith, Mabe, Watkins, Craighead, Ingle, Tyson, Lin, Kraus et~al.}]{blumenthal2021effects}
Blumenthal JA, Hinderliter AL, Smith PJ, Mabe S, Watkins LL, Craighead L, Ingle K, Tyson C, Lin PH, Kraus WE, et~al. (2021) Effects of lifestyle modification on patients with resistant hypertension: results of the triumph randomized clinical trial. \emph{Circulation} 144(15):1212--1226.

\bibitem[{Bordt \protect\BIBand{} Von~Luxburg(2022)}]{bordt2022bandit}
Bordt S, Von~Luxburg U (2022) A bandit model for human-machine decision making with private information and opacity. \emph{International Conference on Artificial Intelligence and Statistics}, 7300--7319 (PMLR).

\bibitem[{Bubeck et~al.(2012)Bubeck, Cesa-Bianchi et~al.}]{bubeck2012regret}
Bubeck S, Cesa-Bianchi N, et~al. (2012) Regret analysis of stochastic and nonstochastic multi-armed bandit problems. \emph{Foundations and Trends{\textregistered} in Machine Learning} 5(1):1--122.

\bibitem[{Cao et~al.(2025)Cao, Keyvanshokooh, \protect\BIBand{} Liu}]{cao2025safe}
Cao J, Keyvanshokooh E, Liu T (2025) Safe reinforcement learning with contextual information: Theory and application to comorbidity management. \emph{Available at SSRN 4583667} .

\bibitem[{Chowdhury et~al.(2022)Chowdhury, Naeem, Quan, Leung, Sikdar, O’Beirne, \protect\BIBand{} Turin}]{chowdhury2022prediction}
Chowdhury MZI, Naeem I, Quan H, Leung AA, Sikdar KC, O’Beirne M, Turin TC (2022) Prediction of hypertension using traditional regression and machine learning models: A systematic review and meta-analysis. \emph{PloS one} 17(4):e0266334.

\bibitem[{Chu et~al.(2011)Chu, Li, Reyzin, \protect\BIBand{} Schapire}]{chu2011contextual}
Chu W, Li L, Reyzin L, Schapire R (2011) Contextual bandits with linear payoff functions. \emph{Proceedings of the Fourteenth International Conference on Artificial Intelligence and Statistics}, 208--214 (JMLR Workshop and Conference Proceedings).

\bibitem[{Cornelissen \protect\BIBand{} Smart(2013)}]{cornelissen2013exercise}
Cornelissen VA, Smart NA (2013) Exercise training for blood pressure: a systematic review and meta-analysis. \emph{Journal of the American heart association} 2(1):e004473.

\bibitem[{Cutkosky et~al.(2022)Cutkosky, Dann, Das, \protect\BIBand{} Zhang}]{cutkosky2022leveraging}
Cutkosky A, Dann C, Das A, Zhang Q (2022) Leveraging initial hints for free in stochastic linear bandits. \emph{International Conference on Algorithmic Learning Theory}, 282--318 (PMLR).

\bibitem[{Dandl et~al.(2020)Dandl, Molnar, Binder, \protect\BIBand{} Bischl}]{dandl2020multi}
Dandl S, Molnar C, Binder M, Bischl B (2020) Multi-objective counterfactual explanations. \emph{International Conference on Parallel Problem Solving from Nature}, 448--469 (Springer).

\bibitem[{Dani et~al.(2008)Dani, Hayes, \protect\BIBand{} Kakade}]{dani2008stochastic}
Dani V, Hayes TP, Kakade SM (2008) Stochastic linear optimization under bandit feedback .

\bibitem[{Denton(2023)}]{t2023frontiers}
Denton B (2023) Frontiers of medical decision-making in the modern age of data analytics. \emph{IISE Transactions} 55(1):94--105.

\bibitem[{Diaz et~al.(2025)Diaz, Satheesh, Moran, Perel, Rodgers, \protect\BIBand{} Schutte}]{diaz2025delivery}
Diaz RR, Satheesh G, Moran AE, Perel P, Rodgers A, Schutte AE (2025) Delivery models to improve adherence to medicines for chronic diseases in primary care. \emph{The Lancet Primary Care} .

\bibitem[{Dominguez-Olmedo et~al.(2021)Dominguez-Olmedo, Karimi, \protect\BIBand{} Schölkopf}]{dominguezolmedo2021adversarial}
Dominguez-Olmedo R, Karimi AH, Schölkopf B (2021) On the adversarial robustness of causal algorithmic recourse. \emph{arXiv:2112.11313} .

\bibitem[{Drusvyatskiy \protect\BIBand{} Lewis(2013)}]{drusvyatskiy2013semi}
Drusvyatskiy D, Lewis AS (2013) Semi-algebraic functions have small subdifferentials. \emph{Mathematical Programming} 140(1):5--29.

\bibitem[{Dusetzina et~al.(2023)Dusetzina, Besaw, Whitmore, Mattingly, Sinaiko, Keating, \protect\BIBand{} Everson}]{dusetzina2023cost}
Dusetzina SB, Besaw RJ, Whitmore CC, Mattingly TJ, Sinaiko AD, Keating NL, Everson J (2023) Cost-related medication nonadherence and desire for medication cost information among adults aged 65 years and older in the us in 2022. \emph{JAMA Network Open} 6(5):e2314211--e2314211.

\bibitem[{Echouffo-Tcheugui et~al.(2013)Echouffo-Tcheugui, Batty, Kivim{\"a}ki, \protect\BIBand{} Kengne}]{echouffo2013risk}
Echouffo-Tcheugui JB, Batty GD, Kivim{\"a}ki M, Kengne AP (2013) Risk models to predict hypertension: a systematic review. \emph{PloS one} 8(7):e67370.

\bibitem[{Filippi et~al.(2010)Filippi, Cappe, Garivier, \protect\BIBand{} Szepesv{\'a}ri}]{filippi2010parametric}
Filippi S, Cappe O, Garivier A, Szepesv{\'a}ri C (2010) Parametric bandits: The generalized linear case. \emph{Advances in neural information processing systems} 23.

\bibitem[{Forouzanfar et~al.(2017)Forouzanfar, Liu, Roth, Ng, Biryukov, Marczak, Alexander, Estep, Abate, Akinyemiju et~al.}]{forouzanfar2017global}
Forouzanfar MH, Liu P, Roth GA, Ng M, Biryukov S, Marczak L, Alexander L, Estep K, Abate KH, Akinyemiju TF, et~al. (2017) Global burden of hypertension and systolic blood pressure of at least 110 to 115 mm hg, 1990-2015. \emph{Jama} 317(2):165--182.

\bibitem[{Gan et~al.(2024)Gan, Keyvanshokooh, Liu, \protect\BIBand{} Murphy}]{gan2024contextual}
Gan K, Keyvanshokooh E, Liu X, Murphy S (2024) Contextual bandits with budgeted information reveal. \emph{International Conference on Artificial Intelligence and Statistics}, 3970--3978 (PMLR).

\bibitem[{Gao \protect\BIBand{} Lakkaraju(2023)}]{gao2023impact}
Gao R, Lakkaraju H (2023) On the impact of algorithmic recourse on social segregation. \emph{International Conference on Machine Learning}, 10727--10743 (PMLR).

\bibitem[{Gao et~al.(2021)Gao, Saar-Tsechansky, De-Arteaga, Han, Lee, \protect\BIBand{} Lease}]{gao2021human}
Gao R, Saar-Tsechansky M, De-Arteaga M, Han L, Lee MK, Lease M (2021) Human-ai collaboration with bandit feedback. \emph{arXiv preprint arXiv:2105.10614} .

\bibitem[{Gao \protect\BIBand{} Yin(2023)}]{gao2023confounding}
Gao R, Yin M (2023) Confounding-robust policy improvement with human-ai teams. \emph{arXiv preprint arXiv:2310.08824} .

\bibitem[{Gil et~al.(2012)Gil, Girela, De~Juan, Gomez-Torres, \protect\BIBand{} Johnsson}]{gil2012predicting}
Gil D, Girela JL, De~Juan J, Gomez-Torres MJ, Johnsson M (2012) Predicting seminal quality with artificial intelligence methods. \emph{Expert Systems with Applications} 39(16):12564--12573.

\bibitem[{Grand-Cl{\'e}ment \protect\BIBand{} Pauphilet(2024)}]{grand2024best}
Grand-Cl{\'e}ment J, Pauphilet J (2024) The best decisions are not the best advice: Making adherence-aware recommendations. \emph{Management Science} .

\bibitem[{Grippo \protect\BIBand{} Sciandrone(2000)}]{grippo2000convergence}
Grippo L, Sciandrone M (2000) On the convergence of the block nonlinear gauss--seidel method under convex constraints. \emph{Operations research letters} 26(3):127--136.

\bibitem[{Group(2013)}]{look2013cardiovascular}
Group LAR (2013) Cardiovascular effects of intensive lifestyle intervention in type 2 diabetes. \emph{New England journal of medicine} 369(2):145--154.

\bibitem[{Harris et~al.(2022)Harris, Podimata, \protect\BIBand{} Wu}]{harris2022strategy}
Harris K, Podimata C, Wu S (2022) Strategy-aware contextual bandits. \emph{Workshop on Trustworthy and Socially Responsible Machine Learning, NeurIPS 2022}.

\bibitem[{Hill(2011)}]{hill2011bayesian}
Hill JL (2011) Bayesian nonparametric modeling for causal inference. \emph{Journal of Computational and Graphical Statistics} 20(1):217--240.

\bibitem[{Ho et~al.(2006)Ho, Rumsfeld, Masoudi, McClure, Plomondon, Steiner, \protect\BIBand{} Magid}]{ho2006effect}
Ho PM, Rumsfeld JS, Masoudi FA, McClure DL, Plomondon ME, Steiner JF, Magid DJ (2006) Effect of medication nonadherence on hospitalization and mortality among patients with diabetes mellitus. \emph{Archives of internal medicine} 166(17):1836--1841.

\bibitem[{Jarczok et~al.(2022)Jarczok, Weimer, Braun, Williams, Thayer, Guendel, \protect\BIBand{} Balint}]{jarczok2022heart}
Jarczok MN, Weimer K, Braun C, Williams DP, Thayer JF, Guendel HO, Balint EM (2022) Heart rate variability in the prediction of mortality: A systematic review and meta-analysis of healthy and patient populations. \emph{Neuroscience \& Biobehavioral Reviews} 143:104907.

\bibitem[{Karimi et~al.(2020{\natexlab{a}})Karimi, Barthe, Balle, \protect\BIBand{} Valera}]{karimi2019model}
Karimi AH, Barthe G, Balle B, Valera I (2020{\natexlab{a}}) Model-agnostic counterfactual explanations for consequential decisions. \emph{International Conference on Artificial Intelligence and Statistics (AISTATS)}.

\bibitem[{Karimi et~al.(2020{\natexlab{b}})Karimi, Barthe, Sch{\"o}lkopf, \protect\BIBand{} Valera}]{karimi2020survey}
Karimi AH, Barthe G, Sch{\"o}lkopf B, Valera I (2020{\natexlab{b}}) A survey of algorithmic recourse: definitions, formulations, solutions, and prospects. \emph{arXiv:2010.04050} .

\bibitem[{Karimi et~al.(2020{\natexlab{c}})Karimi, von K{\"u}gelgen, Sch{\"o}lkopf, \protect\BIBand{} Valera}]{karimi2020probabilistic}
Karimi AH, von K{\"u}gelgen J, Sch{\"o}lkopf B, Valera I (2020{\natexlab{c}}) Algorithmic recourse under imperfect causal knowledge: a probabilistic approach. \emph{Conference on Neural Information Processing Systems (NeurIPS)}.

\bibitem[{Keyvanshokooh et~al.(2025{\natexlab{a}})Keyvanshokooh, Gan, Guo, Liu, \protect\BIBand{} Murphy}]{keyvanshokooh2025learning}
Keyvanshokooh E, Gan K, Guo Y, Liu X, Murphy S (2025{\natexlab{a}}) Learning when to nudge: A dual-agent bandit framework for behavioral interventions. \emph{Available at SSRN 5909422} .

\bibitem[{Keyvanshokooh et~al.(2025{\natexlab{b}})Keyvanshokooh, Zhalechian, Shi, Van~Oyen, \protect\BIBand{} Kazemian}]{keyvanshokooh2025contextual}
Keyvanshokooh E, Zhalechian M, Shi C, Van~Oyen MP, Kazemian P (2025{\natexlab{b}}) Contextual learning with online convex optimization: Theory and application to medical decision-making. \emph{Management Science} 71(12):10442--10464.

\bibitem[{Kulkarni et~al.(2025)Kulkarni, Parati, Bangalore, Bilo, Kim, Kario, Messerli, Stergiou, Wang, Whiteley et~al.}]{kulkarni2025blood}
Kulkarni S, Parati G, Bangalore S, Bilo G, Kim B, Kario K, Messerli F, Stergiou G, Wang J, Whiteley W, et~al. (2025) Blood pressure variability: a review. \emph{Journal of Hypertension} 43(6):929--938.

\bibitem[{Kurdyka(1998)}]{kurdyka1998gradients}
Kurdyka K (1998) On gradients of functions definable in o-minimal structures. \emph{Annales de l'institut Fourier}, volume~48, 769--783.

\bibitem[{Lattimore \protect\BIBand{} Szepesv{\'a}ri(2020)}]{lattimore2020bandit}
Lattimore T, Szepesv{\'a}ri C (2020) \emph{Bandit algorithms} (Cambridge University Press).

\bibitem[{Li et~al.(2010)Li, Chu, Langford, \protect\BIBand{} Schapire}]{li2010contextual}
Li L, Chu W, Langford J, Schapire RE (2010) A contextual-bandit approach to personalized news article recommendation. \emph{WWW}, 661--670.

\bibitem[{Li et~al.(2017)Li, Lu, \protect\BIBand{} Zhou}]{li2017provably}
Li L, Lu Y, Zhou D (2017) Provably optimal algorithms for generalized linear contextual bandits. \emph{International Conference on Machine Learning}, 2071--2080 (PMLR).

\bibitem[{Liao et~al.(2025{\natexlab{a}})Liao, Keyvanshokooh, \protect\BIBand{} Garcia}]{liao2025constraint}
Liao CY, Keyvanshokooh E, Garcia GG (2025{\natexlab{a}}) Constraint-aware self-improving large language model for clinical role model generation. \emph{Available at SSRN 5642250} .

\bibitem[{Liao et~al.(2025{\natexlab{b}})Liao, Keyvanshokooh, Pasquel, \protect\BIBand{} Garcia}]{liao2025rolemodel}
Liao CY, Keyvanshokooh E, Pasquel FJ, Garcia GG (2025{\natexlab{b}}) Augmenting individualized treatment planning via data-driven clinical role model selection. \emph{Available at SSRN 5642250} .

\bibitem[{Mahajan et~al.(2019)Mahajan, Tan, \protect\BIBand{} Sharma}]{mahajan2019preserving}
Mahajan D, Tan C, Sharma A (2019) Preserving causal constraints in counterfactual explanations for machine learning classifiers. \emph{arXiv preprint arXiv:1912.03277} .

\bibitem[{Osterberg \protect\BIBand{} Blaschke(2005)}]{osterberg2005adherence}
Osterberg L, Blaschke T (2005) Adherence to medication. \emph{New England journal of medicine} 353(5):487--497.

\bibitem[{Park et~al.(2024)Park, Liu, Ozdaglar, \protect\BIBand{} Zhang}]{park2024llm}
Park C, Liu X, Ozdaglar A, Zhang K (2024) Do llm agents have regret? a case study in online learning and games. \emph{arXiv preprint arXiv:2403.16843} .

\bibitem[{Patel et~al.(2025)Patel, Huang, \protect\BIBand{} Miliara}]{patel2025understanding}
Patel S, Huang M, Miliara S (2025) Understanding treatment adherence in chronic diseases: challenges, consequences, and strategies for improvement. \emph{Journal of Clinical Medicine} 14(17):6034.

\bibitem[{Pawelczyk et~al.(2020)Pawelczyk, Broelemann, \protect\BIBand{} Kasneci}]{pawelczyk2020learning}
Pawelczyk M, Broelemann K, Kasneci G (2020) Learning model-agnostic counterfactual explanations for tabular data. \emph{Proceedings of The Web Conference 2020}, 3126--3132.

\bibitem[{Pruitt et~al.(2025)Pruitt, Khan, Chaiyakunapruk, Phrommintikul, Aguilera, Tan, Afzal, da~Silva van~der Laan, \protect\BIBand{} Weinman}]{pruitt2025silent}
Pruitt SD, Khan R, Chaiyakunapruk N, Phrommintikul A, Aguilera MAD, Tan NC, Afzal S, da~Silva van~der Laan A, Weinman J (2025) The silent epidemic of non-adherence--insights from the 2024 a: care congress. \emph{BMC proceedings}, volume~19, 13 (Springer).

\bibitem[{Qin et~al.(2023)Qin, Huang, Zhang, \protect\BIBand{} Tang}]{qin2023machine}
Qin K, Huang W, Zhang T, Tang S (2023) Machine learning and deep learning for blood pressure prediction: a methodological review from multiple perspectives. \emph{Artificial Intelligence Review} 56(8):8095--8196.

\bibitem[{Rawal et~al.(2021)Rawal, Kamar, \protect\BIBand{} Lakkaraju}]{rawal2021modelshifts}
Rawal K, Kamar E, Lakkaraju H (2021) Algorithmic recourse in the wild: Understanding the impact of data and model shifts. \emph{arXiv:2012.11788} .

\bibitem[{Rigollet(2015)}]{rigollet201518}
Rigollet P (2015) 18. s997: High dimensional statistics. \emph{Lecture Notes), Cambridge, MA, USA: MIT Open-CourseWare} .

\bibitem[{Russo \protect\BIBand{} Van~Roy(2014)}]{russo2014learning}
Russo D, Van~Roy B (2014) Learning to optimize via posterior sampling. \emph{Mathematics of Operations Research} 39(4):1221--1243.

\bibitem[{Sacks et~al.(2001)Sacks, Svetkey, Vollmer, Appel, Bray, Harsha, Obarzanek, Conlin, Miller, Simons-Morton et~al.}]{sacks2001effects}
Sacks FM, Svetkey LP, Vollmer WM, Appel LJ, Bray GA, Harsha D, Obarzanek E, Conlin PR, Miller ER, Simons-Morton DG, et~al. (2001) Effects on blood pressure of reduced dietary sodium and the dietary approaches to stop hypertension (dash) diet. \emph{New England journal of medicine} 344(1):3--10.

\bibitem[{Slivkins et~al.(2019)}]{slivkins2019introduction}
Slivkins A, et~al. (2019) Introduction to multi-armed bandits. \emph{Foundations and Trends{\textregistered} in Machine Learning} 12(1-2):1--286.

\bibitem[{Stewart et~al.(2023)Stewart, Moon, \protect\BIBand{} Horne}]{stewart2023medication}
Stewart SJF, Moon Z, Horne R (2023) Medication nonadherence: health impact, prevalence, correlates and interventions. \emph{Psychology \& health} 38(6):726--765.

\bibitem[{Tunc et~al.(2014)Tunc, Alagoz, \protect\BIBand{} Burnside}]{tunc2014opportunities}
Tunc S, Alagoz O, Burnside E (2014) Opportunities for operations research in medical decision making. \emph{IEEE intelligent systems} 29(3):59.

\bibitem[{Upadhyay et~al.(2021)Upadhyay, Joshi, \protect\BIBand{} Lakkaraju}]{upadhyay2021robust}
Upadhyay S, Joshi S, Lakkaraju H (2021) Towards robust and reliable algorithmic recourse. \emph{Advances in Neural Information Processing Systems (NeurIPS)}, volume~34.

\bibitem[{Ustun et~al.(2019)Ustun, Spangher, \protect\BIBand{} Liu}]{ustun2019actionable}
Ustun B, Spangher A, Liu Y (2019) Actionable recourse in linear classification. \emph{Proceedings of the conference on fairness, accountability, and transparency}, 10--19.

\bibitem[{Van~Looveren \protect\BIBand{} Klaise(2019)}]{van2019interpretable}
Van~Looveren A, Klaise J (2019) Interpretable counterfactual explanations guided by prototypes. \emph{arXiv preprint arXiv:1907.02584} .

\bibitem[{Verma et~al.(2020)Verma, Dickerson, \protect\BIBand{} Hines}]{verma2020counterfactual}
Verma S, Dickerson J, Hines K (2020) Counterfactual explanations for machine learning: A review. \emph{arXiv:2010.10596} .

\bibitem[{Wachter et~al.(2017)Wachter, Mittelstadt, \protect\BIBand{} Russell}]{wachter2017counterfactual}
Wachter S, Mittelstadt B, Russell C (2017) Counterfactual explanations without opening the black box: Automated decisions and the gdpr. \emph{Harv. JL \& Tech.} 31:841.

\bibitem[{Wang et~al.(2022)Wang, Qi, \protect\BIBand{} Shi}]{wang2022blessing}
Wang J, Qi Z, Shi C (2022) Blessing from experts: Super reinforcement learning in confounded environments. \emph{arXiv preprint arXiv:2209.15448} .

\bibitem[{Wang et~al.(2024{\natexlab{a}})Wang, Zhang, \protect\BIBand{} Zhang}]{wang2024large}
Wang M, Zhang DJ, Zhang H (2024{\natexlab{a}}) Large language models for market research: A data-augmentation approach. \emph{arXiv preprint arXiv:2412.19363} .

\bibitem[{Wang et~al.(2024{\natexlab{b}})Wang, Yahav, \protect\BIBand{} Padmanabhan}]{wang2024smart}
Wang Y, Yahav I, Padmanabhan B (2024{\natexlab{b}}) Smart testing with vaccination: A bandit algorithm for active sampling for managing covid-19. \emph{Information Systems Research} 35(1):120--144.

\bibitem[{Whelton et~al.(2018)Whelton, Carey, Aronow, Casey, Collins, Dennison~Himmelfarb, DePalma, Gidding, Jamerson, Jones et~al.}]{whelton20182017}
Whelton PK, Carey RM, Aronow WS, Casey DE, Collins KJ, Dennison~Himmelfarb C, DePalma SM, Gidding S, Jamerson KA, Jones DW, et~al. (2018) 2017 acc/aha/aapa/abc/acpm/ags/apha/ash/aspc/nma/pcna guideline for the prevention, detection, evaluation, and management of high blood pressure in adults: a report of the american college of cardiology/american heart association task force on clinical practice guidelines. \emph{Journal of the American College of Cardiology} 71(19):e127--e248.

\bibitem[{Xu \protect\BIBand{} Yin(2013)}]{xu2013block}
Xu Y, Yin W (2013) A block coordinate descent method for regularized multiconvex optimization with applications to nonnegative tensor factorization and completion. \emph{SIAM Journal on imaging sciences} 6(3):1758--1789.

\bibitem[{Yang \protect\BIBand{} Burer(2016)}]{yang2016two}
Yang B, Burer S (2016) A two-variable approach to the two-trust-region subproblem. \emph{SIAM Journal on Optimization} 26(1):661--680.

\bibitem[{Yang et~al.(2025)Yang, Liao, Keyvanshokooh, Shao, Weber, Pasquel, \protect\BIBand{} Garcia}]{yang2025responsible}
Yang Y, Liao CY, Keyvanshokooh E, Shao H, Weber MB, Pasquel FJ, Garcia GGP (2025) A responsible framework for assessing, selecting, and explaining machine learning models in cardiovascular disease outcomes among people with type 2 diabetes: Methodology and validation study. \emph{JMIR Medical Informatics} 13:e66200.

\bibitem[{Ye et~al.(2025)Ye, Yoganarasimhan, \protect\BIBand{} Zheng}]{ye2025lola}
Ye Z, Yoganarasimhan H, Zheng Y (2025) Lola: Llm-assisted online learning algorithm for content experiments. \emph{Marketing Science} .

\bibitem[{Zhalechian et~al.(2022)Zhalechian, Keyvanshokooh, Shi, \protect\BIBand{} Van~Oyen}]{zhalechian2022online}
Zhalechian M, Keyvanshokooh E, Shi C, Van~Oyen MP (2022) Online resource allocation with personalized learning. \emph{Operations Research} 70(4):2138--2161.

\bibitem[{Zhalechian et~al.(2023)Zhalechian, Keyvanshokooh, Shi, \protect\BIBand{} Van~Oyen}]{zhalechian2023data}
Zhalechian M, Keyvanshokooh E, Shi C, Van~Oyen MP (2023) Data-driven hospital admission control: A learning approach. \emph{Operations Research} 71(6):2111--2129.

\bibitem[{Zhang et~al.(2020)Zhang, Xie, Li, \protect\BIBand{} CS~Lui}]{zhang2020conversational}
Zhang X, Xie H, Li H, CS~Lui J (2020) Conversational contextual bandit: Algorithm and application. \emph{Proceedings of the web conference 2020}, 662--672.

\bibitem[{Zhou et~al.(2023)Zhou, Wang, Yan, \protect\BIBand{} Tan}]{zhou2023spoiled}
Zhou T, Wang Y, Yan L, Tan Y (2023) Spoiled for choice? personalized recommendation for healthcare decisions: A multiarmed bandit approach. \emph{Information Systems Research} 34(4):1493--1512.

\bibitem[{Zuo et~al.(2022)Zuo, Hu, Yu, Li, Zhao, \protect\BIBand{} Joe-Wong}]{zuo2022hierarchical}
Zuo J, Hu S, Yu T, Li S, Zhao H, Joe-Wong C (2022) Hierarchical conversational preference elicitation with bandit feedback. \emph{Proceedings of the 31st ACM International Conference on Information \& Knowledge Management}, 2827--2836.

\end{thebibliography}

\newpage

\begin{center}
\centering
\textbf{{\large LIBRA: Language Model Informed Bandit Recourse
Algorithm  \\ for Personalized Treatment Planning}} \\
\textbf{\large (Online Appendices)}
\end{center}

\begin{APPENDICES}

\smallskip
\section{Detailed Proofs in \Cref{sec: GLRB}}\label{appendix: proof in Section 2}

\smallskip
\subsection{Proof of \Cref{lem: closed-form-solution}}

We note that it suffices to compute the best recourse for each arm individually and then choose the arm that yields the highest expected reward. Moreover, since the link function $\mu(\cdot)$ is strictly increasing, we can drop it from the optimization and instead directly maximize the inner product $(x_I,\checkx_M)^\top \theta_a^\star$. Accordingly, we focus on the following optimization problem for each arm/action:
\begin{align*}
    \max_{\checkx_M\in \mathbb{R}^{d_M}} &\quad
    (x_I,\checkx_M)^\top \theta^\star
    \label{eq: RO-Arm} \tag{\textsf{RO-Arm}}\\
    \text{subject to}\hspace{-0.15cm} &\quad \norm{\checkx_M-x_M} \leq \gamma. \notag
\end{align*}

The problem \eqref{eq: RO-Arm} is equivalent to the following optimization problem:
\begin{equation}
    \max_{\norm{\checkx_M-x_M}\leq \gamma}  \checkx_M^\top \theta_{a, M}^\star.
\end{equation}
According to Hölder's inequality, we have 
\begin{equation}
    \checkx_M^\top \theta_{a, M}^\star= x_M^\top \theta_{a, M}^\star + \left(\checkx_M-x_M\right)^\top \theta_{a, M}^\star\leq x_M^\top \theta_{a, M}^\star+\norm{\checkx_M-x_M}\norm{\theta_{a, M}^\star}_\star\leq x_M^\top \theta_{a, M}^\star+\gamma \norm{\theta_{a, M}^\star}_\star.  
\end{equation}
The equality is attained when $\checkx_M-x_M\in \gamma \cdot \partial \norm{\theta_{a, M}^\star}_\star$. This completes the proof. \QED

\subsection{Proof of Lemma~\ref{lemma: glm-radius}}
Recall we have the following GLM model for the stochastic reward:
\begin{align*}
r_t(x_t,a_t): 
= \mu({\theta_a^\star}^\top x_t) + \xi_t
= \mu(x_{t, I}^\top \; \theta_{a,I}^\star +  x_{t, M}^\top \; \theta_{a,M}^\star) + \xi_t,
\end{align*}
where 
$x_t = (x_{t, I}, x_{t, M}) \in \mathbb{R}^d$, $\theta_a^\star = (\theta_{a,I}^\star, \theta_{a,M}^\star)$, and $\theta_a^\star \in \mathbb{R}^d$ is the unknown true parameter for each arm $a \in [K]$. In addition, $\mu(\cdot): \mathbb{R} \rightarrow \mathbb{R}$ is a known, $L_{\mu}$-Lipschitz, link function. Additionally, we assume $\mu'(z) \geq c_{\mu} > 0$ for all $z \in \mathbb{R}$. Moreover, $\xi_t$ is zero-mean, $\sigma$-sub-Gaussian random noise (i.e., conditionally sub-Gaussian given the past history). Also, recall that $\underline{\beta}_{\mathcal{X}}\leq\|x\|_2\leq \beta_{\mathcal{X}}$ for all $x\in \mathcal{X}$ and $\sup_{\theta\in \Theta} \|\theta\|\leq \beta_{\Theta}$.

Our objective is to construct a high-probability confidence set for the unknown parameter $\theta_a^\star$. We denote $\mathcal{I}_{t, a} = \{ s < t \;|\; a_s = a \}$ as the set of time steps before $t$ where arm $a$ was selected and let $n_{t, a} = |\mathcal{I}_{t, a}|$.  Given historical observations $\{(r_s, x_s, a_s)\}_{s=1}^{t-1}$, we estimate the unknown model parameter $\theta_{a}^\star$  can be minimizing the following 
regularized negative log-likelihood loss function for $\lambda > 0$:
\begin{align*}
    \widehat{\theta}_{t, a} := 
    \argmin_{\theta\in \mathbb{R}^d} 
    \bigg\{
    \sum_{s \in \mathcal{I}_{t, a}}
    \ell(r_s, x_s^\top \theta) + \frac{\lambda}{2} \; \norm{\theta}_2^2
    \bigg\},
\end{align*}
where $\ell(r, z): = rz + m(z)$ 
with $m'(z) = \mu(z)$ 
being the negative log-likelihood loss for a single observation.

\smallskip
We let $L_{t, a} (\theta) = 
\sum_{s \in \mathcal{I}_{t, a}}
\ell(r_s, x_s^\top \theta) + \frac{\lambda}{2} \; \norm{\theta}_2^2$ and denote $V_{t, a}:= \lambda I + \sum_{s \in \mathcal{I}_{t, a}} x_s   x_s^\top$ as the covariance matrix. Next, applying the Fundamental Theorem of Calculus to the gradient yields the following:
\begin{align*}
    \nabla L_{t, a} (\widehat{\theta}_{t, a}) =
    \nabla L_{t, a} (\theta_{a}^\star) +\int_{0}^{1} \nabla^2 L_{t, a} (\theta_{a}^\star+\tau(\widehat{\theta}_{t, a} - \theta_{a}^\star)) \cdot (\widehat{\theta}_{t, a} - \theta_{a}^\star) \operatorname{d} \! \tau.
\end{align*}

Because the regularized MLE estimator $\widehat{\theta}_{t, a}$ minimizes $L_{t, a}$, we have $\nabla L_{t, a} (\widehat{\theta}_{t, a}) = 0$. Accordingly, rearranging the above expression, we have the following for the estimation error: 
\begin{align*}
    \widehat{\theta}_{t, a} - \theta_{a}^\star = 
    - \left( \int_{0}^{1} \nabla^2 L_{t, a} (\theta_{a}^\star+\tau(\widehat{\theta}_{t, a} - \theta_{a}^\star)) \cdot (\widehat{\theta}_{t, a} - \theta_{a}^\star) \operatorname{d} \! \tau \right)^{-1}
     \nabla L_{t, a} (\theta_{a}^\star).
\end{align*}

Computing the gradient of the loss at $\theta_{a}^\star$, we have: 
\begin{align*}
    \nabla L_{t, a} (\theta_{a}^\star) 
    =
    \sum_{s \in \mathcal{I}_{t, a}} 
    \left( \mu({\theta_a^\star}^\top x_t) - r_s \right) x_s 
    + \lambda \theta_{a}^\star
    =
    - \sum_{s \in \mathcal{I}_{t, a}} \xi_s x_s + \lambda \theta_{a}^\star
    =: - g_{a, t} + \lambda \theta_{a}^\star. 
\end{align*}
We next bound the estimation error $(\widehat{\theta}_{t, a} - \theta_{a}^\star)$ in the $V_{t, a}$-norm by substituting the Taylor expression into the $V_{t, a}$ norm and using the triangle inequality as follows:
\begin{align*}
    \norm{\widehat{\theta}_{t, a} - \theta_{a}^\star}_{V_{t, a}}
     &= 
    \norm{\left( \int_{0}^{1} \nabla^2 L_{t, a} (\theta_{a}^\star+\tau(\widehat{\theta}_{t, a} - \theta_{a}^\star)) \cdot (\widehat{\theta}_{t, a} - \theta_{a}^\star) \operatorname{d} \! \tau \right)^{-1}
     g_{a, t} }_{V_{t, a}}
    \\
    &\quad +
    \norm{\left( \int_{0}^{1} \nabla^2 L_{t, a} (\theta_{a}^\star+\tau(\widehat{\theta}_{t, a} - \theta_{a}^\star)) \cdot (\widehat{\theta}_{t, a} - \theta_{a}^\star) \operatorname{d} \! \tau \right)^{-1}
    \; \lambda \theta_{a}^\star }_{V_{t, a}}.
\end{align*}
We next bound the inverse Hessian. Since $\mu(\cdot)$ is strictly increasing and twice differentiable, the Hessian has the following form:
\begin{align*}
    \nabla^2 L_{t, a} (\theta) 
    = 
    \sum_{s \in \mathcal{I}_{t, a}} 
    \mu'(\theta^\top x_t) \;  x_s x_s^\top
    + \lambda I   
    \succeq   
    c_{\mu} \cdot \bigg( \sum_{s \in \mathcal{I}_{t, a}} x_s x_s^\top + \frac{\lambda}{c_{\mu}}   I \bigg)
    = c_{\mu} \cdot V_{t, a},
\end{align*}
where we assume $\mu'(z) \geq c_{\mu} > 0$ for all $z \in \mathbb{R}$. This implies $\left( \nabla^2 L_{t, a} (\overline{\theta}_a) \right)^{-1} \preceq \frac{1}{c_{\mu}}   V_{t, a}^{-1}$. Therefore, using the bound, we have the following:  
\begin{align*}
    \norm{\widehat{\theta}_{t, a} - \theta_{a}^\star}_{V_{t, a}}
     \leq 
     \frac{1}{c_{\mu}} 
     \left(
    \norm{g_{a, t} }_{V_{t, a}^{-1}}
    + \lambda
    \norm{\theta_{a}^\star}_{V_{t, a}^{-1}}
    \right).
\end{align*}
We note that $\norm{\theta_{a}^\star}_{V_{t, a}^{-1}} \leq \norm{\theta_{a}^\star}_2 / \sqrt{\lambda} \leq \beta_{\Theta} / \sqrt{\lambda}$. Recall the noise term $\xi_s = r_s - \mu({\theta_a^\star}^\top x_s)$ satisfies the condition $\mathbb{E}[\xi_s | \mathcal{F}_{s-1}] = 0$ and is conditionally $\sigma$-sub-Gaussian. Accordingly, leveraging the self-normalized martingale concentration inequalities \citep[Theorem~2]{abbasi2011improved} and the union bound, 
for each arm $a \in [K]$ and time $t\geq 0$, 
with probability $1-\delta$, we have 
\begin{align*}
    \norm{g_{a, t} }_{V_{t, a}^{-1}} \leq  \sigma \sqrt{d \log \left( 1 + \frac{\beta_{\mathcal{X}}^2 n_{t, a}}{\lambda} \right) + d\log \left( \frac{K}{\delta} \right)}.
\end{align*}
Thus, we have the following confidence bound with probability at least $1-\delta$ for the unknown parameter 
\begin{align*}
\Theta_{t, a}:=
\left\{
\theta_a:\|\theta_a-\widehat{\theta}_{t, a}\|_{V_{t, a}}\leq \rho_{t, a}
\right\},
\end{align*}
where 
$$
\rho_{t, a} = \frac{1}{c_{\mu}} 
\left(
 \sigma \sqrt{d \log \left( 1 + \frac{\beta_{\mathcal{X}}^2 n_{t, a}}{\lambda} \right) + d\log \left( \frac{K}{\delta} \right)} +
\sqrt{\lambda} \beta_{\Theta}
\right).
$$
This completes the proof.
\QED

\subsection{Proof of \Cref{thm::global-optimality}}
    We first introduce the following result on quadratically constrained quadratic programs (QCQPs).
    \begin{lemma}[Adapted from \citealt{yang2016two,bienstock2016note}]
    \label{lem::qcqp}
        The QCQP of the following form:
        \begin{align*}
            \max_{x}\quad &x^{\top}Cx+2c^{\top}x\\
            \text{subject to}\quad &(x-a_i)^{\top}A_i(x-a_i)\leq 1, \quad i=1, 2,
        \end{align*}
        can be solved to global optimality in polynomial time when each $A_i$ is positive semidefinite (PSD).
    \end{lemma}

\proof{Proof of \Cref{thm::global-optimality}.}
When the Mahalanobis norm is used in \eqref{eq::ORO-Arm}, the optimistic recourse optimization problem can be reformulated as a QCQP. Specifically, defining the Mahalanobis norm by $\norm{x}_{\Sigma} = \sqrt{x^{\top} \Sigma x}$, we rewrite \eqref{eq::ORO-Arm} as follows:
\begin{align*}
   \max_{\checkx_M, \theta=(\theta_M, \theta_I)} \quad &\checkx_M^\top \theta_{M} +x_I^\top \theta_{I} \\
    \text{subject to}\hspace{0.2cm}\quad &\hspace{-0.05cm}(\checkx_M-x_M)^{\top}\Sigma (\checkx_M-x_M) \leq \gamma^2,\\
    &\hspace{-0.05cm}(\theta - \hat\theta)^{\top}V(\theta - \hat\theta)\leq \rho^2.
    \end{align*}
    This is a QCQP with a linear objective and two convex quadratic constraints. Therefore, by \Cref{lem::qcqp}, the problem is globally solvable in polynomial time, which completes the proof.
\QED
\endproof

\subsection{Proof of \Cref{thm::convergence-rate-KL}}
    First, we show that the optimistic recourse optimization problem~\eqref{eq::ORO-Arm} satisfies the KL property. To this goal, we reformulate it as an unconstrained optimization problem by incorporating the constraints via indicator functions:
    \begin{equation}
        \max_{\theta, \checkx_M}\checkx_M^\top \theta_{M}+x_I^\top \theta_{I} - \iota_{\{\norm{\checkx_M- x_M} \leq \gamma\}}(\checkx_M)- \iota_{\left\{\norm{\theta-\hat{\theta}} \leq \rho\right\}}(\theta).\nonumber
    \end{equation}
    Here, $\iota_A(x)=0$ if $x\in A$ and $\iota_A(x)=+\infty$ otherwise. Each term in the objective is a semi-algebraic function. Since the sum of semi-algebraic functions is also semi-algebraic, the overall objective is semi-algebraic. It then follows from standard results (e.g., \citealp{attouch2010proximal}) that the objective satisfies the KL property.

    The convergence rate result is classical; see, for example, \citet[Theorem 2.9]{xu2013block}.
  \QED

\subsection{Proof of Theorem~\ref{thm: regretbound-GLRB}}

Assume the optimal action selected at time $t$ is $a_t^\star$ and the optimal recourse selected at time $t$ is $\checkx_t^\star$. We then define the recourse regret as follows:
\[
   \Regret_\pi(T) = \mathbb{E}\left[\sum_{t=1}^T   r(\checkx_t^\star, a_t^\star) 
   - 
   r(\checkx_t, a_t) \right],
\]
where $\checkx_t=(x_{t, I},\checkx_{t, M})$ and $a_t$ is the optimal solution to \eqref{eq::ORO-Arm}. 

According to Lemma~\ref{lemma: glm-radius}, with probability at least $1-\delta$, for all $t\in [T]$ and $a\in \mathcal{A}$, we have:
\begin{align*}
\theta_a^\star\in \Theta_{t, a}:=
\left\{
\theta_a:\|\theta_a-\widehat{\theta}_{t, a}\|_{V_{t, a}}\leq \rho_{t, a}
\right\},
\end{align*}
where 
$$
\rho_{t, a} = \frac{1}{c_{\mu}} 
\left(
 \sigma \sqrt{d \log \left( 1 + \frac{\beta_{\mathcal{X}}^2 n_{t, a}}{\lambda} \right) + d\log \left( \frac{K}{\delta} \right)} +
\sqrt{\lambda} \beta_{\Theta}
\right).
$$
In addition, since $\checkx_t$, $a_t$, and $\theta_{a_t}$ is the optimal solution to \eqref{eq::ORO-Arm}, we have
\[
\mu(\theta_{a_t}^\top \checkx_t)
\geq \mu((\theta^\star_{ a^\star_t})^\top \checkx_t^\star).
\]
Accordingly, we can then bound the one-step regret with probability at least $1-\delta$ as follows: 
\begin{align*}
    \mathbb{E}\left[r_t(x_t^\star, a_t^\star) 
   - r_t(x_t, a_t)\right] 
   &= \mu\Big(\big(\theta_{a^\star_t}^{\star}\big)^\top \checkx_t^\star\Big) -
   \mu((\theta_{a_t}^\star)^\top \checkx_t) \\
   &\leq \mu(\theta_{a_t}^\top \checkx_t) - \mu((\theta_{a_t}^\star)^\top \checkx_t)
   \\
   &\stackrel{(a)}{\leq} L_{\mu} \cdot \left| \checkx_t^\top \left( \theta_{a_t} - \widehat{\theta}_{t, a} + \widehat{\theta}_{t, a} - \theta_{a_t}^\star \right) \right|\\
    &\stackrel{(b)}{\leq} 2L_{\mu} \cdot \rho_{t, a_t} \cdot \|x_t\|_{V_{t, a_t}^{-1}},
\end{align*}
where in $(a)$, we apply the Lipschitz property of the link function $\mu(\cdot)$.

\smallskip
We then bound the total recourse regret as follows:
\begin{align*}
    \Regret_\pi(T) &= \mathbb{E}\left[\sum_{t=1}^T  \  r(x_t^\star, a_t^\star) 
   - r(x_t, a_t) \right] \\
   & \leq 2L_{\mu} \sum_{t=1}^T \rho_{t,a_t} \cdot \min\Big\{\|x_t\|_{V_{t,a_t}^{-1}} , 1\Big\}
   \\
   &\leq 2L_{\mu} \sum_{a=1}^K \sum_{t \in \mathcal{I}_{t, a}} \rho_{t, a} \cdot \min\Big\{\|x_t\|_{V_{t,a_t}^{-1}} , 1\Big\} 
   \\
   &\leq 2L_{\mu} \cdot \rho_{T} \sum_{a=1}^K \sum_{t \in \mathcal{I}_{t, a}}  \min\Big\{\|x_t\|_{V_{t,a_t}^{-1}} , 1\Big\},
\end{align*}
where $\rho_{T} = \frac{1}{c_{\mu}} 
\left(
 \sigma \sqrt{d \log \left( 1 + \frac{\beta_{\mathcal{X}}^2 n_{t, a}}{\lambda} \right) + d\log \left( \frac{K}{\delta} \right)} +
\sqrt{\lambda} \beta_{\Theta}
\right)\geq 1$. To proceed,
we apply the elliptical potential lemma \cite[Lemma 19.4]{lattimore2020bandit}, for each $a\in \mathcal{A}$:
\begin{align*}
    \sum_{t \in \mathcal{I}_{t, a}}  \min\Big\{\|x_t\|_{V_{t,a_t}^{-1}}^2 , 1\Big\} \leq 2d \log\left(1 + \frac{\beta_{\mathcal{X}}^2 n_{t, a}}{\lambda d} \right),
\end{align*}
and applying Cauchy-Schwarz inequality, we can argue:
\begin{align*}
    \sum_{t \in \mathcal{I}_{t, a}}  \min\Big\{\|x_t\|_{V_{t,a_t}^{-1}} , 1\Big\}\leq \sqrt{n_{t, a}}   \sqrt{2d \log\left(1 + \frac{\beta_{\mathcal{X}}^2 n_{t, a}}{\lambda d} \right)}. 
\end{align*}
Using the Cauchy-Schwarz inequality that $\sum_{a=1}^K \sqrt{n_{t, a}} \leq \sqrt{K \cdot \sum_{a=1}^K n_{t, a}} = \sqrt{KT}$, we can get the following:
\begin{align*}
    \Regret_\pi(T) 
    &\leq 
    2L_{\mu} \cdot \rho_{T} \sum_{a=1}^K \sum_{t \in \mathcal{I}_{t, a}}  \min\Big\{\|x_t\|_{V_{t,a_t}^{-1}} , 1\Big\} 
    \leq 2L_{\mu} \cdot \rho_{T}  \sqrt{2dKT \log\left(1 + \frac{\beta_{\mathcal{X}}^2 T}{\lambda d} \right)}.
\end{align*}
This completes the proof.
\QED

\smallskip
\section{Proofs in Section~\ref{sec: LIBRA}}\label{appendix: proof}

\smallskip
\subsection{Proof of Theorem~\ref{thm: warm start guarantee}}
Let $(\checkx_t^B,a_t^B)$ denote the action selected by the base bandit algorithm at round $t$. We measure the bandit's uncertainty at this action by the width of its confidence interval as follows:
\[
\CI_t^B := \CI_t(\checkx_t^B,a_t^B) = \UCB_t(\checkx_t^B,a_t^B) - \LCB_t(\checkx_t^B,a_t^B).
\]
Let $(\checkx_t^L,a_t^L)$ denote the action proposed by the LLM when queried. Throughout the proof, we condition on the good event that the confidence sets are valid, i.e.,
\[
\theta_a^\star\in \Theta_{t, a},\qquad \forall a\in \mathcal{A}.
\]
Under this event, the true reward of any action always lies between its upper and lower confidence bounds. We analyze two cases depending on whether the bandit’s uncertainty $\CI_t^B$ exceeds the threshold $\Delta$, which governs whether \textsf{LIBRA} queries the LLM.

\smallskip
\underline{{\bf Case 1: $\CI_t^B > \Delta$ (query LLM)}}.
When the confidence interval is wide, the bandit is highly uncertain about its own action. In this case, \textsf{LIBRA} queries the LLM and plays the recommendation $(\checkx_t^L,a_t^L)$. 
By the $\eta$-suboptimality assumption, the LLM provides an $\eta$-suboptimal recommendation, meaning its reward is within $\eta$ of the optimal reward at round $t$. Therefore, we have:
\[
\regret(t) = r_t(x^\star, a^\star) - r_t(\checkx_t^L,a_t^L) \le \eta.
\]
Since $\CI_t^B > \Delta$, we have $\min\{ \Delta, \CI_t^B \} = \Delta$. Define the scaling constant:
\[
c := \max\left\{ 1, \frac{\eta}{\Delta} \right\}.
\]
which captures how the LLM’s quality compares to the uncertainty threshold. We now distinguish two subcases:
\begin{itemize}
    \item If $\eta \le \Delta$, then the LLM is sufficiently accurate relative to the threshold, and $c = 1$. In this case, we have:
    \[
    c \cdot \min\{ \Delta, \CI_t^B \} = \Delta \ge \eta \ge \regret(t).
    \]
    \item If $\eta > \Delta$, then the LLM is less accurate, but this loss is exactly accounted for by the scaling factor $c = \eta / \Delta$, yielding the following:
    \[
    c \cdot \min\{ \Delta, \CI_t^B \} = \frac{\eta}{\Delta} \cdot \Delta = \eta \ge \regret(t).
    \]
\end{itemize}
In both subcases, the regret incurred by following the LLM is bounded by:
\[
\regret(t) \le \max\left\{ 1, \frac{\eta}{\Delta} \right\} \cdot \min\{ \Delta, \CI_t^B \}.
\]

\smallskip
\underline{{\bf Case 2: $\CI_t^B \le \Delta$ (do not query LLM)}}.
When the confidence interval is narrow, the bandit is already sufficiently confident in its action. In this case, \textsf{LIBRA} does not query the LLM and instead plays the base bandit action $(\checkx_t^B,a_t^B)$. Under the confidence event, the instantaneous regret of this action is at most the width of its confidence interval:
\[
\regret(t) = r_t(x^\star, a^\star) - r_t(\checkx_t^B,a_t^B) \le \UCB_t(\checkx_t^B,a_t^B) - \LCB_t(\checkx_t^B,a_t^B) = \CI_t^B.
\]
Since $\CI_t^B \le \Delta$, it follows that $\min\{ \Delta, \CI_t^B \} = \CI_t^B$, and hence:
\[
\regret(t) \le \CI_t^B = \min\{ \Delta, \CI_t^B \} \le \max\left\{ 1, \frac{\eta}{\Delta} \right\} \cdot \min\{ \Delta, \CI_t^B \}.
\]

\smallskip
{\bf Conclusion}.
In both cases—whether \textsf{LIBRA} relies on the LLM or on the base bandit—the instantaneous regret at round $t$ is controlled by the same quantity:
\[
\regret(t) \le \max\left\{ 1, \frac{\eta}{\Delta} \right\} \cdot \min\left\{ \Delta, \CI_t(\checkx_t^B,a_t^B) \right\}.
\]
This establishes the warm-start guarantee and completes the proof. 
\hfill \QED

\smallskip
\subsection{Proof of Theorem~\ref{thm: LLM-effort}}
We work on the high-probability event under which all confidence sets $\Theta_{t,a}$ hold
simultaneously for every $t \le T$ and $a \in [K]$. On this event, for any pair $(x,a)$,
the confidence width satisfies
\begin{equation}\label{eq:width}
\CI_t(x,a) = 2 L_\mu \rho_{t,a} \|x\|_{V_{t,a}^{-1}}.
\end{equation}

By Algorithm~\ref{alg: LIBRA}, the algorithm queries the LLM at round $t$
only when the bandit's confidence width exceeds the threshold $\Delta$. Combining this
decision rule with~\eqref{eq:width} yields
\begin{equation}\label{eq:query-threshold}
\|x_t\|_{V_{t, a_t}^{-1}}
>
\frac{\Delta}{2 L_\mu \rho_{t,a_t}}
\ge
\frac{\Delta}{2 L_\mu \rho_T},
\end{equation}
where the final inequality uses the monotonicity $\rho_{t,a_t} \le \rho_T$ for all $t$. Thus, every time \textsf{LIBRA} queries the LLM, the corresponding self-normalized feature vector must exceed a fixed positive threshold.

For each arm $a$, define the index set of rounds in which arm $a$ is selected,
\[
\mathcal{I}_{T,a} := \{t \le T : a_t = a\},
\]
and the subset of those rounds in which the threshold in~\eqref{eq:query-threshold}
is exceeded:
\[
\mathcal{S}_{T,a}
:= \Bigl\{
t \in \mathcal{I}_{T,a} : \|x_t\|_{V_{t,a}^{-1}}
> \tfrac{\Delta}{2 L_\mu \rho_T}
\Bigr\}.
\]
By construction, every query to the LLM for arm $a$ must lie in $\mathcal{S}_{T,a}$; thus, the total number of LLM queries satisfies
\[
N_{\mathrm{LLM}}(T,\Delta)
\le
\sum_{a=1}^K |\mathcal{S}_{T,a}|.
\]

To bound $|\mathcal{S}_{T,a}|$, we convert the threshold condition into a summable quantity. We use the elementary inequality:
\[
\mathbf{1}\{v > \alpha\}
\le
\min\{1, (v/\alpha)^2\}
\qquad \text{for all } v,\alpha>0.
\]
Applying this inequality with
$\alpha = \Delta / (2 L_\mu \rho_T)$ gives
\begin{equation}
\begin{aligned}
|\mathcal{S}_{T,a}|
&\le
\sum_{t \in \mathcal{I}_{T,a}}
\mathbf{1}
\Bigl\{
\|x_t\|_{V_{t,a}^{-1}}
> \tfrac{\Delta}{2 L_\mu \rho_T}
\Bigr\}\le
\sum_{t \in \mathcal{I}_{T,a}}
\min\Bigl\{
1,
\Bigl(\tfrac{2 L_\mu \rho_T}{\Delta}\Bigr)^2
\|x_t\|_{V_{t,a}^{-1}}^{2}
\Bigr\}.
\end{aligned}
\end{equation}
Since $(2 L_\mu \rho_T / \Delta)^2 \ge 1$, we may pull out this factor to obtain
\begin{equation}
|\mathcal{S}_{T,a}|
\le
\Bigl(\tfrac{2 L_\mu \rho_T}{\Delta}\Bigr)^2
\sum_{t \in \mathcal{I}_{T,a}}
\min\Big\{1, \|x_t\|_{V_{t,a}^{-1}}^{2}\Big\}.
\end{equation}

We now apply the elliptical potential lemma, which gives
\[
\sum_{t \in \mathcal{I}_{T,a}}
\min\Big\{1, \|x_t\|_{V_{t,a}^{-1}}^{2}\Big\}
\le
2d \log\Bigl(1 + \frac{\beta_X^2 n_{T,a}}{\lambda d}\Bigr),
\]
where $n_{T,a} = |\mathcal{I}_{T,a}|$ is the number of times arm $a$ was pulled.

Substituting this bound, we obtain
\begin{equation}
|\mathcal{S}_{T,a}|
\le
\frac{8 L_\mu^2}{\Delta^2}
\rho_T^2 d
\log\Bigl(1 + \frac{\beta_X^2 n_{T,a}}{\lambda d}\Bigr).
\end{equation}

Finally, summing over $a=1,\dots,K$ and using Jensen's inequality,
\begin{equation}
\begin{aligned}
N_{\mathrm{LLM}}(T,\Delta)
&\le
\sum_{a=1}^K |\mathcal{S}_{T,a}|
\le
\frac{8 L_\mu^2}{\Delta^2}
\rho_T^2 d
\sum_{a=1}^K
\log\Bigl(1 + \frac{\beta_X^2 n_{T,a}}{\lambda d}\Bigr)
\le
\frac{8 L_\mu^2}{\Delta^2}
\rho_T^2 d K
\log\Bigl(1 + \frac{\beta_X^2 T}{\lambda d K}\Bigr),
\end{aligned}
\end{equation}
which proves the claimed bound. \hfill \QED

\subsection{Proof of Theorem~\ref{thm: LIBRA improvementguarantee}}
We work under the same high-probability event as in Theorem~\ref{thm: regretbound-GLRB}, under which all confidence intervals are valid and the bound used to define $\rho_T$ holds. In particular, for every round $t$ and every action $(x, a)$, we have
\[
r_t(x, a) \in \Big[\LCB_t(x, a), \UCB_t(x, a) \Big],
\]
and the base algorithm selects
\[
(\checkx_t^B, a_t^B) \in \argmax_{(x, a)} \UCB_t(x, a).
\]

\smallskip
\underline{{\bf Step 1: From Per-round to Cumulative Regret via the Warm-start Guarantee}}.
The warm-start guarantee (Theorem~\ref{thm: warm start guarantee}) provides a per-round control on the regret of \textsf{LIBRA} that interpolates between the LLM quality and the bandit’s uncertainty. Specifically, by Theorem~\ref{thm: warm start guarantee}, the per-round regret of \textsf{LIBRA} at each round $t$ satisfies:
\[
\regret(t) = r_t(x^\star, a^\star) - r_t(x_t, a_t)
\le \max\left\{1, \frac{\eta}{\Delta}\right\}
\cdot \min\big\{ \Delta, \CI_t(\checkx_t^B, a_t^B) \big\},
\]
where $(x_t, a_t)$ is the action played by \textsf{LIBRA}, and $\eta$ is the suboptimality guarantee of the LLM.

Summing over $t = 1, \dots, T$, we obtain
\[
\Regret(T)
= \sum_{t=1}^T \regret(t)
\le \max\left\{1, \frac{\eta}{\Delta}\right\}
\sum_{t=1}^T \min\big\{ \Delta, \CI_t(\checkx_t^B, a_t^B) \big\}.
\]

For any nonnegative sequence $(c_t)_{t=1}^T$ and any threshold $\Delta > 0$, the inequality
\[
\sum_{t=1}^T \min\{\Delta, c_t\} \le \min\left\{ \Delta T, \sum_{t=1}^T c_t \right\}
\]
holds. Applying this inequality with $c_t = \CI_t(\checkx_t^B, a_t^B)$ gives
\[
\sum_{t=1}^T \min\big\{ \Delta, \CI_t(\checkx_t^B, a_t^B) \big\}
\le \min\left\{ \Delta T, \sum_{t=1}^T \CI_t(\checkx_t^B, a_t^B) \right\}.
\]
Hence, combining the above bounds, we arrive at
\begin{equation}
\label{eq:libra-regret-reduction}
\Regret(T) \le \max\left\{1, \frac{\eta}{\Delta}\right\}
\cdot \min\left\{ \Delta T, \sum_{t=1}^T \CI_t(\checkx_t^B, a_t^B) \right\}.
\end{equation}
This inequality highlights the core tradeoff: cumulative regret is controlled either by the LLM-driven term $\Delta T$ or by the cumulative bandit uncertainty.

\smallskip
\underline{{\bf Step 2: Bounding the Sum of Confidence Widths}}.
By the analysis in Theorem~\ref{thm: regretbound-GLRB}, which relies on GLM confidence bounds and the elliptical potential lemma, we have with probability at least $1 - \delta$,
\[
\sum_{t=1}^T \CI_t(\checkx_t^B, a_t^B)
\le 2 L_{\mu} \rho_T \sqrt{2 d K T \log\left(1 + \frac{\beta_{\mathcal{X}}^2 T}{\lambda d} \right)},
\]
where
\[
\rho_T = \frac{1}{c_{\mu}} 
\left(
\sigma \sqrt{d \log\left(1 + \frac{\beta_{\mathcal{X}}^2 T}{\lambda} \right)
+ d \log\left( \frac{K}{\delta} \right)}
+ \sqrt{\lambda} \beta_{\Theta}
\right).
\]

Substituting into~\eqref{eq:libra-regret-reduction} gives the following bound, valid with probability at least $1 - \delta$:
\[
\Regret(T) \le \max\left\{1, \frac{\eta}{\Delta}\right\}
\cdot
\min \left\{
\Delta T,
2 L_{\mu} \rho_T \sqrt{2 d K T \log\left(1 + \frac{\beta_{\mathcal{X}}^2 T}{\lambda d} \right)}
\right\}.
\]

\smallskip
\underline{{\bf Step 3: Specialization to $\Delta = \Theta(\eta)$}}.
Finally, we interpret the bound by choosing the threshold $\Delta$ on the order of the LLM's suboptimality level.
Suppose we set $\Delta = c \eta$ for some absolute constant $c > 0$. Then
\[
\max\left\{1, \frac{\eta}{\Delta}\right\} = \max\left\{1, \frac{1}{c} \right\} = \Theta(1),
\quad
\min\{ \Delta T, \cdots \} = \min\{ \Theta(\eta T), \cdots \}.
\]
That is, the first term inside the minimum becomes $\Delta T = \Theta(\eta T)$.
The second term inside the minimum is of order $\tilde{O}(d \sqrt{K T})$, since $\rho_T$ and the logarithmic terms are polylogarithmic in $T$, $d$, $K$, and $1/\delta$, Therefore, up to logarithmic factors, we have the following:
\[
\Regret(T) = \tilde{O}\left( \min\{ \eta T, d \sqrt{K T} \} \right),
\]
as claimed. Intuitively, if the LLM is very good ($\eta$ small), \textsf{LIBRA} behaves like a strong warm-start policy and regret grows as $\eta T$. If learning eventually dominates, \textsf{LIBRA} falls back to optimal bandit behavior with regret $d \sqrt{K T}$. The algorithm automatically adapts to whichever regime is better.
\hfill \QED

\smallskip
\subsection{Proof of \Cref{thm::lower-bounds}}
We only present the proof for the setting in which a suboptimal oracle is available. The result for the other case follows by taking the suboptimality gap $\eta\to \infty$. The proof is divided into two parts.

\noindent\textbf{Part 1.} We first establish a regret lower bound of the form
\begin{equation}
    \Regret(T) = \Omega\left(\gamma d_{M}\sqrt{KT} \wedge \eta T\right).
\end{equation}
To this end, we divide the time horizon into $K$ equally sized subperiods $T_1, \dots, T_{K}$.\footnote{Without loss of generality, assume $\frac{T}{K}$ is an integer.} We then construct the arm parameters $\{\theta_k\}_{k=1}^{K}$ and the contexts $\{x_t\}_{t=1}^{T}$ accordingly.

\smallskip
\underline{{\bf Dimension Reduction}}.
Since we can only modify the mutable context, which is of dimension $d_M$, without loss of generality, we perform a dimension reduction and restrict our attention to the mutable subspace of dimension \( d_M \). This can be done by simply setting $x_{t, I}=0$ for all $1\leq t\leq T$.

\smallskip
\underline{{\bf Arm Parameter Construction}}.
Consider a \( d_M \)-dimensional hypercube \( \{\pm \sqrt{K/T}\}^{d_M} \). To proceed, we introduce the following lemma adapted from Lemma 5.12 in \citealt{rigollet201518}.

\begin{lemma}[Varshamov-Gilbert Lemma]
    \label{lem::varshamov-gilbert}
    For any $\gamma \in(0,1 / 2)$, there exist binary vectors $v_1, \ldots v_M \in\{0,1\}^d$ such that the followings hold:
    \begin{itemize}
        \item $\norm{v_i- v_j}_1 \geq\left(\frac{1}{2}-\gamma\right) d$ for all $i \neq j$;
        \item $M=\lfloor e^{\gamma^2 d}\rfloor \geq e^{\frac{\gamma^2 d}{2}}$.
    \end{itemize}
\end{lemma}

Therefore, upon choosing $\gamma=\frac{1}{4}$ in the above lemma, we know that there exists a set
\(
\{\theta_1^{\star}, \dots, \theta_{K}^{\star}\}\subset \{\pm \sqrt{K/T}\}^{d_M}
\)
satisfying $\norm{\theta_i^\star-\theta_j^\star}_1\geq \frac{1}{2}\sqrt{\frac{K}{T}}d_M$ for any $i\neq j$ when \( K \leq e^{d_M/8} \).

We define the feasible parameter set for each arm \( k \) as
\begin{equation}
    \Theta_k = \theta_{k}^{\star} + \omega\{-1, +1\}^{d_M},
\end{equation}
where $\omega := \gamma\sqrt{K/T} \wedge \frac{\eta}{d_M}$. 
Nature selects an arbitrary vector \( \theta_k \in \Theta_k \), which remains unknown to the learner.

\smallskip
\underline{{\bf Context Construction}}.
For each subperiod \( T_k \), define the context set as
\begin{equation}
    \mathcal{X}_k = \sqrt{T/K} \theta_k^{\star} + c[-1, +1]^{d_M},
\end{equation}
where $c\leq \frac{1}{32}$ is a univeral constant.
Then, for any \( x_k \in \mathcal{X}_k \) and \( j \neq k \), we have
\begin{equation}
    \begin{aligned}
        \inner{x_k}{\theta_k-\theta_j}
&\geq \inner{\sqrt{T/K} \theta_k^{\star}}{\theta_k-\theta_j}-2c\norm{\theta_k-\theta_j}_1\\
&\geq \inner{\sqrt{T/K} \theta_k^{\star}}{\theta_k^{\star}-\theta_j^{\star}}-2c\norm{\theta_k^\star-\theta_j^\star}_1-4c\omega d_M\\
&\geq \frac{1}{2}\sqrt{K/T}d_M-c\sqrt{K/T}d_M-4c\omega d_M\\
&\geq \frac{1}{4}\sqrt{K/T}d_M.
    \end{aligned}
\end{equation}
Here, we use the triangle inequality in the first two inequalities. Hence, arm \( k \) is uniquely optimal during subperiod \( T_k \). Under this prior information, the learner effectively faces \( K \) disjoint bandit subproblems, each with horizon \( T/K \).

On the other hand, for any $x_k\in \cX_k$ and $\theta_k, \theta_k'\in \Theta_k$, we have 
\begin{equation}
    \begin{aligned}
        \left|\inner{x_k}{\theta_k-\theta_k'}\right|\leq \norm{x}_{\infty}\norm{\theta_k-\theta_k'}_1\leq 4\omega d_M\leq \eta.
    \end{aligned}
\end{equation}
Hence, any parameter in $\Theta_k$ is a $\eta$-suboptimal choice. Consequently, the $\eta$-suboptimal oracle cannot provide any useful information in the worst case.

\smallskip
\underline{{\bf Lower Bound for a Single Subperiod}}.
Fix any \( k \in \{1,\dots,K\} \) and consider subperiod \( T_k \). Define the rescaled variables
\begin{equation}
    \tilde{x} = \frac{x - \sqrt{T/K} \theta_k^{\star}}{c}, \quad \tilde{\theta} = \frac{\theta_k - \theta_k^{\star}}{\omega\sqrt{T/K}},
\end{equation}
which ensures \( \tilde{\theta} \in \{-\sqrt{K/T}, \sqrt{K/T}\}^{d_M} \) and \( \tilde{x} \in [-1, 1]^{d_M} \). Therefore, the learning interaction in subperiod \( T_k \) reduces to a standard linear bandit problem with action set \( [-1,1]^{d_M} \) and parameter set \( \{-\sqrt{K/T}, \sqrt{K/T}\}^{d_M} \). To proceed, we introduce the following classic regret lower bound.

\begin{lemma}[Theorem 24.1 in \citealt{lattimore2020bandit}]\label{lem:LS}
Let the action set be \( \mathcal{A} = [-1,1]^{d_M} \) and \( \Theta = \{-T^{-1/2}, T^{-1/2}\}^{d_M} \). Then, for any policy, there exists \( \theta \in \Theta \) such that
\begin{equation}
    \Regret(T) \geq \frac{e^{-2}}{8} d_M\sqrt{T}.
\end{equation}
\end{lemma}

Applying Lemma~\ref{lem:LS} in the rescaled setting and converting back to the original scale, the expected regret in subperiod \( T_k \) is at least
\begin{equation}
    \Regret(T_k) \geq c\omega\sqrt{T} \cdot \frac{e^{-2}}{8}   d_M\sqrt{\frac{T}{K}} = \Omega\left(\gamma d_M\sqrt{T/K} \wedge \eta \frac{T}{K}\right).
\end{equation}

\smallskip
\underline{{\bf Putting Everything Together}}.
Since the \( K \) subperiods \( \{T_k\}_{k=1}^K \) are constructed to be statistically independent, and because the learner obtains no information about \( \theta_j \) while operating in subperiod \( T_k \) for \( j \neq k \), each subproblem is effectively isolated. Applying Lemma~\ref{lem:LS} to each subproblem yields an expected regret of order \( \Omega\left(\gamma d_M\sqrt{T/K} \wedge \eta \frac{T}{K}\right) \). Summing over the \( K \) subperiods, we obtain the total regret bound
\begin{equation}
    \Regret(T)=\sum_{k=1}^{K}\Regret(T_k) \geq K \cdot \Omega\left(\gamma d_M\sqrt{T/K} \wedge \eta \frac{T}{K}\right) = \Omega\left(\gamma d_M\sqrt{KT} \wedge \eta T\right),
\end{equation}
which completes the proof of \textbf{Part 1}.

\noindent\textbf{Part 2.} Next, we prove the following regret lower bound
\begin{equation}
    \Regret(T) = \Omega\left(\sqrt{d K T} \wedge \eta T\right).
\end{equation}
Without loss of generality, we consider the setting where the perturbation radius $\gamma=0$ and there is no presence of an $\eta$-suboptimal oracle. We divide the time horizon into \( d \) equally sized subperiods \( T_1, \dots, T_{d} \), each of length \( T/d \).

\smallskip
\underline{{\bf Context Construction}}.
In each subperiod \( T_k \) for \( k = 1, \dots, d \), set the context to be \( x_t = e_k \) for all \( t \in T_k \).

\smallskip
\underline{{\bf Lower Bound for the Learner}}.
The learner faces a $K$-arm bandit problem in each subperiod $T_k$. For each subperiod $T_k$, the regret lower bound is \( \Omega(\sqrt{K T_k}) \) (see Theorem 15.2 in \citealt{lattimore2020bandit}). Therefore, the total regret lower bound is
\begin{equation}
    \Regret(T) = \Omega\left(\sqrt{d_M K T}\right).
\end{equation}
This completes the proof of \textbf{Part 2}.
\QED

\smallskip
\section{Proofs in \Cref{sec::noncompliance}}
\subsection{Proof of \Cref{thm::random-noncompliance}}
Denote the optimal recourse and action at time $t$ as 
\begin{equation}
    (\checkx_t^\star, a_t^\star) :=   \argmax_{a\in \mathcal{A}, \norm{\checkx_M- x_M} \leq \gamma} \bE_{\epsilon}\Big[\mu\left((\checkx_M+\epsilon)^\top \theta_{a,M}^\star +x_I^\top \theta_{a,I}^\star\right)\Big].
\end{equation}
We then define the recourse regret as follows:
\[
   \Regret_{\pi}(T) = \mathbb{E}\left[\sum_{t=1}^T  \  r(\checkx_t^\star+\epsilon_t, a_t^\star) 
   - 
   r(\checkx_t+\epsilon_t, a_t) \right].
\] 
Similarly, from Lemma~\ref{lemma: glm-radius}, recall we have the following uncertainty set with probability at least $1-\delta$:
\begin{align*}
\Theta_{t, a}:=
\left\{
\theta_a:\|\theta_a-\widehat{\theta}_{t, a}\|_{V_{t, a}}\leq \rho_{t, a}
\right\},
\end{align*}
where 
$$
\rho_{t, a} = \frac{1}{c_{\mu}} 
\left(
 \sigma \sqrt{d \log \left( 1 + \frac{\beta_{\mathcal{X}}^2 n_{t, a}}{\lambda} \right) + d\log \left( \frac{K}{\delta} \right)} +
\sqrt{\lambda} \beta_{\Theta}
\right).
$$
In addition, since $\checkx_t$, $a_t$ and $\theta_{t, a_t}$ is the optimal solution to \eqref{eq::robust-ORO}, we have the following argument:
\[
\mathbb{E}\Big[\mu(\theta_{t, a_t}^\top (\checkx_t+\epsilon_t))\Big]
\geq \mathbb{E}\left[\mu(\theta_{a_t^\star}^\top (\checkx^\star_t+\epsilon_t))\right].
\]
Accordingly, we can then bound the one-step regret with probability at least $1-\delta$ as follows: 
\begin{align*}
    \mathbb{E}\Big[r(\checkx_t^\star+\epsilon_t, a_t^\star) 
   - 
   r(\checkx_t+\epsilon_t, a_t) \Big]
   &= \mathbb{E}\left[\mu((\theta^\star_{a^\star_t})^\top (\checkx_t^\star+\epsilon_t)) -
   \mu((\theta^\star_{a_t})^\top (\checkx_t+\epsilon_t))\right] \\
   &\leq \mathbb{E}\Big[\mu(\theta_{t, a_t}^\top (\checkx_t+\epsilon_t)) -
   \mu((\theta^\star_{a_t})^\top (\checkx_t+\epsilon_t))\Big]
   \\
   &\leq L_{\mu} \cdot \mathbb{E}\left[\left| (\checkx_t+\epsilon_t)^\top \left( \theta_{a_t}^\star - \widehat{\theta}_{t, a_t} + \widehat{\theta}_{t, a_t} - \theta_{t, a_t} \right) \right|\right]\\
    &\leq 2L_{\mu} \cdot \rho_{t, a_t} \cdot E_{\pi,\epsilon}\left[\|\checkx_t+\epsilon_t\|_{V_{t, a_t}^{-1}}\right],
\end{align*}
where in above, we apply the Lipschitz property of $\mu(\cdot)$ (i.e., $L_{\mu} := \sup_z |\mu'(z)|$).

We then bound the total recourse regret as follows:
\begin{align*}
    \Regret_\pi(T) &= \mathbb{E}\left[\sum_{t=1}^T  \  r(\checkx_t^\star+\epsilon_t, a_t^\star) 
   - 
   r(\checkx_t+\epsilon_t, a_t) \right] \\
   & \leq 2L_{\mu} \mathbb{E}\left[\sum_{t=1}^T  \min\Big\{\rho_{t,a_t} \cdot\|\checkx_t+\epsilon_t\|_{V_{t, a}^{-1}}^2, 1\Big\}\right] 
   \\
   &\leq 2L_{\mu} \cdot \mathbb{E}\left[\rho_{T} \sum_{a=1}^K \sum_{t \in \mathcal{I}_{T, a}}  \min\Big\{\|\checkx_t+\epsilon_t\|_{V_{t, a}^{-1}}^2, 1\Big\}\right],
\end{align*}
where $\rho_{T} = \frac{1}{c_{\mu}} 
\left(
 \sigma \sqrt{d \log \left( 1 + \frac{\beta_{\mathcal{X}}^2 T}{\lambda} \right) + d\log \left( \frac{K}{\delta} \right)} +
\sqrt{\lambda} \beta_{\Theta}
\right)$.
To proceed, note that $\norm{\checkx_t+\epsilon_t}\leq \beta_{\mathcal{X}}+\epsilon$. Then, we apply the elliptical potential lemma \cite[Lemma 19.4]{lattimore2020bandit}, for each $a\in \mathcal{A}$:
\begin{align*}
    \sum_{t \in \mathcal{I}_{T, a}}  \min\Big\{\|\checkx_t+\epsilon_t\|_{V_{t, a}^{-1}}^2, 1\Big\} \leq 2d \log\left(1 + \frac{(\beta_{\mathcal{X}}+\epsilon)^2 n_{T, a}}{\lambda d} \right),
\end{align*}
and applying Cauchy-Schwarz inequality, we can argue:
\begin{align*}
    \sum_{t \in \mathcal{I}_{T, a}}  \min\Big\{\|\checkx_t+\epsilon_t\|_{V_{t, a}^{-1}}, 1\Big\} \leq \sqrt{n_{T, a}} \; \sqrt{2d \log\left(1 + \frac{(\beta_{\mathcal{X}}+\epsilon)^2n_{T, a}}{\lambda d} \right)}. 
\end{align*}
Using the Cauchy-Schwarz inequality that $\sum_{a=1}^K \sqrt{n_{T, a}} \leq \sqrt{K \cdot \sum_{a=1}^K n_{T, a}} = \sqrt{KT}$, we can get the following bound:
\begin{align*}
    \Regret_\pi(T) 
    &\leq 
    2L_{\mu} \cdot \mathbb{E} \Bigg[\rho_{T} \sum_{a=1}^K \sum_{t \in \mathcal{I}_{T, a}}  \min\Big\{\|\checkx_t+\epsilon_t\|_{V_{t, a}^{-1}}^2, 1\Big\}\Bigg] 
    \leq 2L_{\mu} \cdot \rho_{T}  \sqrt{2dKT \log\left(1 + \frac{(\beta_{\mathcal{X}}+\epsilon)^2 T}{\lambda d} \right)}.
\end{align*}
This completes the proof.
\QED
\endproof

\subsection{Proof of \Cref{lem: xx-optimization}}
We first solve the inner minimization problem
\begin{align*}
   &\min_{\barx_M} \hspace{0.65cm} \barx_{M}^\top \theta_M \\
   & \hspace{-0.3cm}\text{subject to} \hspace{0.2cm}  \norm{\checkx_{M}-\barx_{M}}\leq \epsilon.
\end{align*}
Following the same argument as in the proof of \Cref{lem: closed-form-solution}, the optimal solution is $\barx_M^\star=\checkx_M-\epsilon\cdot v$ where $v\in \partial \norm{\theta_M}_{\star}$. Substituting this into the outer problem, we obtain
\begin{align*}
   &\min_{\barx_M} \hspace{0.65cm} \checkx_{M}^\top \theta_M \\
   & \hspace{-0.3cm}\text{subject to} \hspace{0.2cm}  \norm{\checkx_{M}-x_{M}}\leq \delta.
\end{align*}
whose optimal solution is given by $\checkx_M^\star=x_M+\gamma\cdot v$ where $v\in \partial \norm{\theta_M}_{\star}$. Combining these two results, we derive that $\barx_M^\star=x_M+(\delta-\epsilon)\cdot v$ where $v\in \partial \norm{\theta_M}_{\star}$. This completes the proof.
\QED

\subsection{Proof of \Cref{lem::KL}}
    We begin by reformulating the problem as a single-level optimization by explicitly solving the inner minimization. Specifically, the inner solution is given by $\barx_M^\star=\checkx_M-\epsilon\cdot \partial \norm{\theta_M}_\star$. Substituting this into the outer objective yields
    \begin{equation}
        \max_{\norm{\theta-\hat{\theta}}_V\leq \rho, \norm{\checkx_M-x_M}\leq \delta} \left(\checkx_M-\epsilon\cdot \partial \norm{\theta_M}_{\star}\right)^{\top}\theta_M+x_I^{\top} \theta_I.
    \end{equation}
    Using the property that $\partial \norm{\theta_M}_\star^\top \theta_M = \norm{\theta_M}$, we can simplify the objective and rewrite the problem equivalently as an unconstrained optimization
    \begin{equation}
        \max_{\theta, \checkx_M} \checkx_M^{\top}\theta_M-\epsilon\norm{\theta_M}+x_I^{\top} \theta_I- \iota_{\{\norm{\checkx_M- x_M}\leq \delta\}}(\checkx_M)- \iota_{\left\{\norm{\theta-\hat{\theta}}_V\leq \rho\right\}}(\theta_{M}).
    \end{equation}
    Since the norm $\norm{\cdot}$ is assumed to be semi-algebraic, and the class of semi-algebraic functions is closed under addition, scalar multiplication, and composition with continuous semi-algebraic functions, the entire objective function is semi-algebraic. It follows from standard results (e.g., \citealp{attouch2010proximal}) that the objective satisfies the KL property. This completes the proof.
\QED

\smallskip
\subsection{Proof of \Cref{thm::adversarial-noncompliance}}

First, \eqref{eq::ORO-NC-Arm} is equivalent to
\begin{equation}
    \begin{aligned}
        &\max_{a\in \mathcal{A}} \max_{\|\checkx_{t,M}- x_{t,M}\|\leq \delta,\theta_a\in \Theta_{t, a}} \quad (\checkx_{t,M}, x_{t,I})^\top \theta_a-\epsilon\norm{\theta_{a, M}}_\star.
    \end{aligned}
\end{equation}
where $\|\cdot\|_\star$ is the dual norm of $\|\cdot\|$. Denote the optimal recourse and action at time $t$ as
\begin{equation}
    (\checkx_t^\star, a_t^\star) =\argmax_{a\in \mathcal{A}, \norm{\checkx_{t, M}- x_{t, M}} \leq \gamma} \min_{\norm{\checkx_{t, M}-\barx_{t, M}}\leq \epsilon} \ \mu\left(\barx_{t, M}^\top \theta_{a,M}^\star +x_{t, I}^\top \theta_{a,I}^\star\right).
\end{equation}
We then define the recourse regret as follows:
\[
   \Regret_{\pi}(T) = \mathbb{E}\left[\sum_{t=1}^T  \  \min_{\norm{\checkx^\star_{t, M}-\barx^\star_{t, M}}\leq \epsilon} r(\barx^\star_t, a_t^\star) 
   - 
   \min_{\norm{\checkx_{t, M}-\barx_{t, M}}\leq \epsilon} r(\barx_t, a_t) \right].
\] 
To proceed, we bound the one-step regret with probability at least $1-\delta$ as follows: 
\begin{align*}
    &\min_{\norm{\checkx^\star_{t, M}-\barx^\star_{t, M}}\leq \epsilon} r(\barx^\star_t, a_t^\star) 
   - 
   \min_{\norm{\checkx_{t, M}-\barx_{t, M}}\leq \epsilon} r(\barx_t, a_t)\\
   &= \mu((\theta^\star_{a^\star_t})^\top \checkx_t^\star-\epsilon \|(\theta^\star_{a^\star_t})_M\|_{\star}) -
   \mu((\theta^\star_{a_t})^\top \checkx_t-\epsilon \|(\theta^\star_{a_t})_M\|_{\star}) \\
   &\leq \mu((\theta_{t, a_t})^\top \checkx_t-\epsilon \|(\theta_{t, a_t})_M\|_{\star}) -
   \mu((\theta^\star_{a_t})^\top \checkx_t-\epsilon \|(\theta^\star_{a_t})_M\|_{\star})
   \\
   &\leq L_{\mu} \cdot \left| (\theta_{t, a_t})^\top \checkx_t-\epsilon \|(\theta_{t, a_t})_M\|_{\star}-\left((\theta^\star_{a_t})^\top \checkx_t-\epsilon \|(\theta^\star_{a_t})_M\|_{\star}\right) \right|\\
    &\leq 2L_{\mu} \cdot \rho_{t, a_t} \cdot \|\checkx_t\|_{V_{t, a_t}^{-1}}+L_\mu\cdot \epsilon\cdot \left|((\theta_{t, a_t})_M)^{\top}\left(\partial \norm{(\theta_{t, a_t})_M}_{\star}-\partial \norm{(\theta_{t, a_t}^\star)_M}_{\star}\right)\right|.
\end{align*}
Notice that 
\begin{equation}
    \begin{aligned}
        &\left|((\theta_{t, a_t})_M)^{\top}\left(\partial \norm{(\theta_{t, a_t})_M}_{\star}-\partial \norm{(\theta_{t, a_t}^\star)_M}_{\star}\right)\right|\\
        &\leq \norm{(\theta_{t, a_t}-\theta_{t, a_t}^\star)_M}_\star+\left|(\partial \norm{(\theta_{t, a_t}^\star)_M}_\star)^{\top}(\theta_{t, a_t}-\theta_{t, a_t}^\star)_M\right|\\
        &\leq 2\norm{(\theta_{t, a_t}-\theta_{t, a_t}^\star)_M}_\star.
    \end{aligned}
\end{equation}
Moreover, we have
\begin{equation}
    \norm{(\theta_{t, a_t}-\theta_{a_t}^\star)_M}_\star
= \sup_{\|u\|\le 1} u^\top\big((\theta_{t, a_t}-\theta_{a_t}^\star)_M\big)
\le \sup_{\|u\|\le 1}\|u_M\|_{V_{t, a_t}^{-1}}\|\theta_{t, a_t}-\theta_{a_t}^\star\|_{V_{t, a_t}}\leq \rho_{t, a_t}\sup_{\|u\|\le 1}\|u_M\|_{V_{t, a_t}^{-1}}.
\end{equation}
Therefore, we have
\begin{equation}
    \min_{\norm{\checkx^\star_{t, M}-\barx^\star_{t, M}}\leq \epsilon} r(\barx^\star_t, a_t^\star) 
   - 
   \min_{\norm{\checkx_{t, M}-\barx_{t, M}}\leq \epsilon} r(\barx_t, a_t)\leq 2L_{\mu} \cdot \rho_{t, a_t} \cdot \left(\|\checkx_t\|_{V_{t, a_t}^{-1}}+\epsilon\cdot\sup_{\|u\|\le 1}\|u_M\|_{V_{t, a_t}^{-1}}\right).
\end{equation}
We then bound the total recourse regret as follows:
\begin{align*}
    \Regret_\pi(T) &= \mathbb{E}\left[\sum_{t=1}^T  \  \min_{\norm{\checkx^\star_{t, M}-\barx^\star_{t, M}}\leq \epsilon} r(\barx^\star_t, a_t^\star) 
   - 
   \min_{\norm{\checkx_{t, M}-\barx_{t, M}}\leq \epsilon} r(\barx_t, a_t) \right] \\
   & \leq 2L_{\mu}\cdot \mathbb{E}\left[\sum_{t=1}^T \min\left\{\rho_{t, a_t} \cdot \left(\|\checkx_t\|_{V_{t, a_t}^{-1}}+\epsilon\cdot\sup_{\|u\|\le 1}\|u_M\|_{V_{t, a_t}^{-1}}\right), 1\right\}\right] 
   \\
   &\leq 2L_{\mu} \cdot \rho_{T}  \sqrt{2dKT \log\left(1 + \frac{(\beta_{\mathcal{X}}+\epsilon)^2 T}{\lambda d} \right)}+2L_\mu\cdot \epsilon\cdot \mathbb{E}\left[\sum_{t=1}^T\sup_{\|u\|\le 1}\|u_M\|_{V_{t, a_t}^{-1}}\right].
\end{align*}
According to Assumption~\ref{ass::coverage}, we further have
\begin{equation}
    \begin{aligned}
        \mathbb{E}\left[\sum_{t=1}^T\sup_{\|u\|\le 1}\|u_M\|_{V_{t, a_t}^{-1}}\right]&\leq \frac{1}{\gamma}\mathbb{E}\left[\sum_{t=1}^T\frac{1}{\sqrt{n_{t, a_t}}}\right]\\
        &=\frac{1}{\gamma}\mathbb{E}\left[\sum_{a=1}^{K}\sum_{t\in \mathcal{I}_{T, a}}^T\frac{1}{\sqrt{n_{t, a}}}\right]\\
        &\leq \frac{1}{\gamma} \mathbb{E}\left[\sum_{a=1}^{K}2\sqrt{n_{T, a}}\right]\\
        &\leq \frac{2}{\gamma}\sqrt{KT}.
    \end{aligned}
\end{equation}
Putting everything together, we derive
\begin{equation}
    \Regret_\pi(T) \leq 2L_{\mu} \cdot \rho_{T}  \sqrt{2dKT \log\left(1 + \frac{(\beta_{\mathcal{X}}+\epsilon)^2 T}{\lambda d} \right)}+\frac{4L_\mu \epsilon}{\gamma}\cdot \sqrt{KT}.
\end{equation}
This completes the proof.
\QED

\section{Empirical Results based on Semi-Synthetic Data}\label{app:semi-synthetic}

We build an additional  healthcare example using real-world data on seminal quality in this section. The real-world dataset is an observational data where part of the patients has received surgical intervention and other patients have not. For our experiments, we need the potential outcomes of one patient for both under surgical treatment or not, also known as counterfactuals. Such counterfactuals are always not available for observational dataset since we can only observe the outcome under one possible treatment in the past. Therefore, we simulate the counterfactuals by fitting a linear model on each treatment arm. 

\citet{gil2012predicting} ask 100 volunteers to provide semen samples and analyze them according to the World Health Organization (WHO) 2010 criteria. The dataset is available \href{https://archive.ics.uci.edu/dataset/244/fertility}{here}. In the dataset, the authors record the age, childish diseases, accident, surgical intervention, high fevers in the last year, alcohol consumption, smoking habit, number of hours sitting per day, and the diagnosis. The outcome variable is any alteration in the sperm parameters, which include Normozoospermia, Asthenozoospermia, Oligozoospermia, and Teratozoospermia.

We consider three mutable features: alcohol consumption, smoking habit, and the number of hours sitting per day. We assume the outcome function is $\mathcal{N}(5,1) * \mathds{1}[\text{diagnosis} = \text{Normal}] + \mathcal{N}(0,1) * \mathds{1}[\text{diagnosis} = \text{Altered}].$ The treatment is whether to ask patients to perform a surgical intervention. 
To simulate the counterfactual outcome when patients implement recourse recommendations that are not in the dataset, we fit separate linear regression models for each treatment arm and use them as the underlying data-generating processes for the corresponding arm. We set LLM quality as $q=0.4$ following the simulation setup in the main paper. 

The recourse regret is shown in \Cref{fig:fert_regret}. The number of humans queried is included in \Cref{fig:fert_asked}. We observe similar qualitative conclusion as the synthetic data where \textsf{LIBRA} significantly outperforms other methods with only limited number of human queries. 

\begin{figure}[htp]
  \centering
  \begin{subfigure}[b]{0.45\textwidth}
    \includegraphics[width=\textwidth]{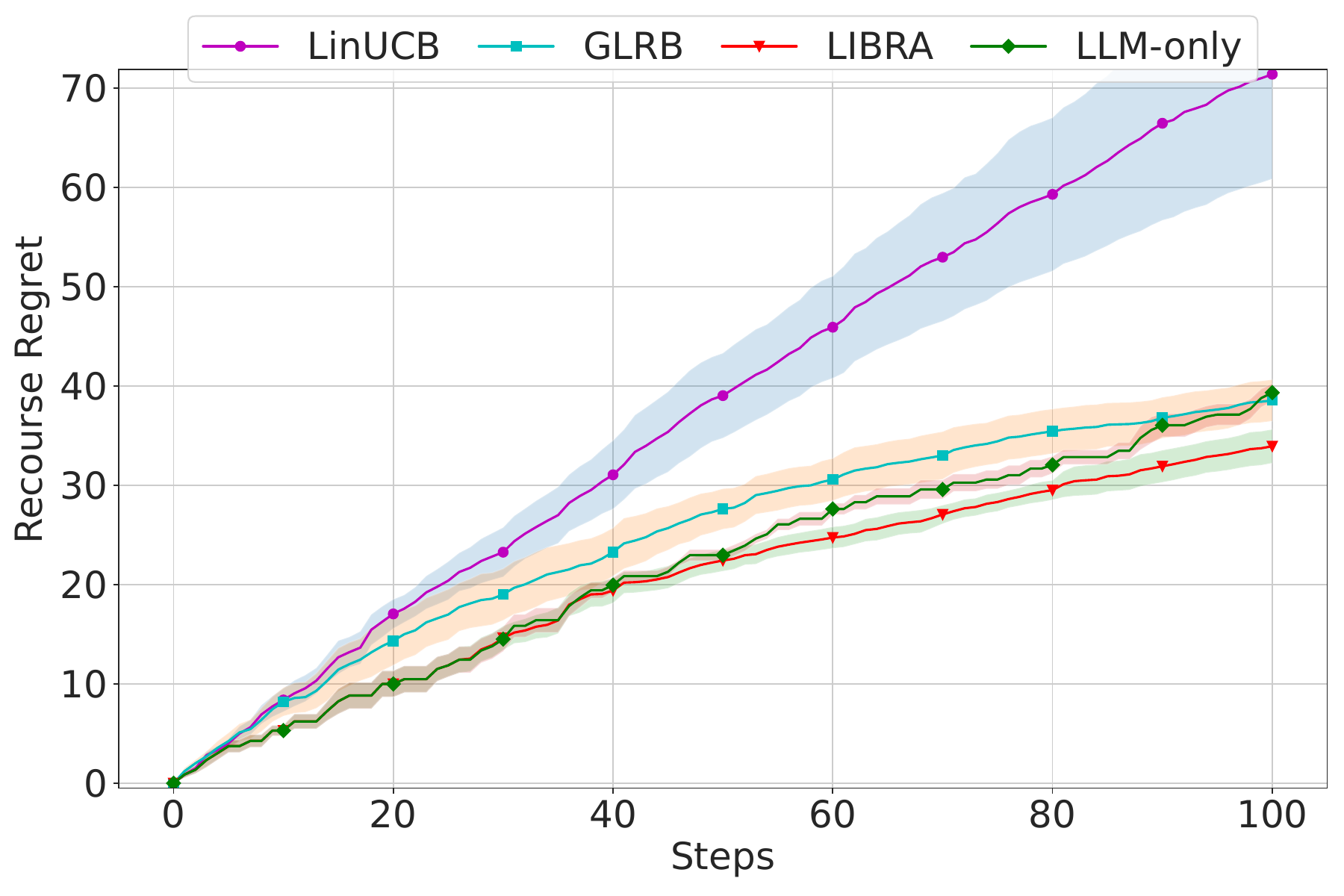}
    \caption{Recourse Regret}
    \label{fig:fert_regret}
  \end{subfigure}
  \hfill 
  \begin{subfigure}[b]{0.45\textwidth}
    \includegraphics[width=\textwidth]{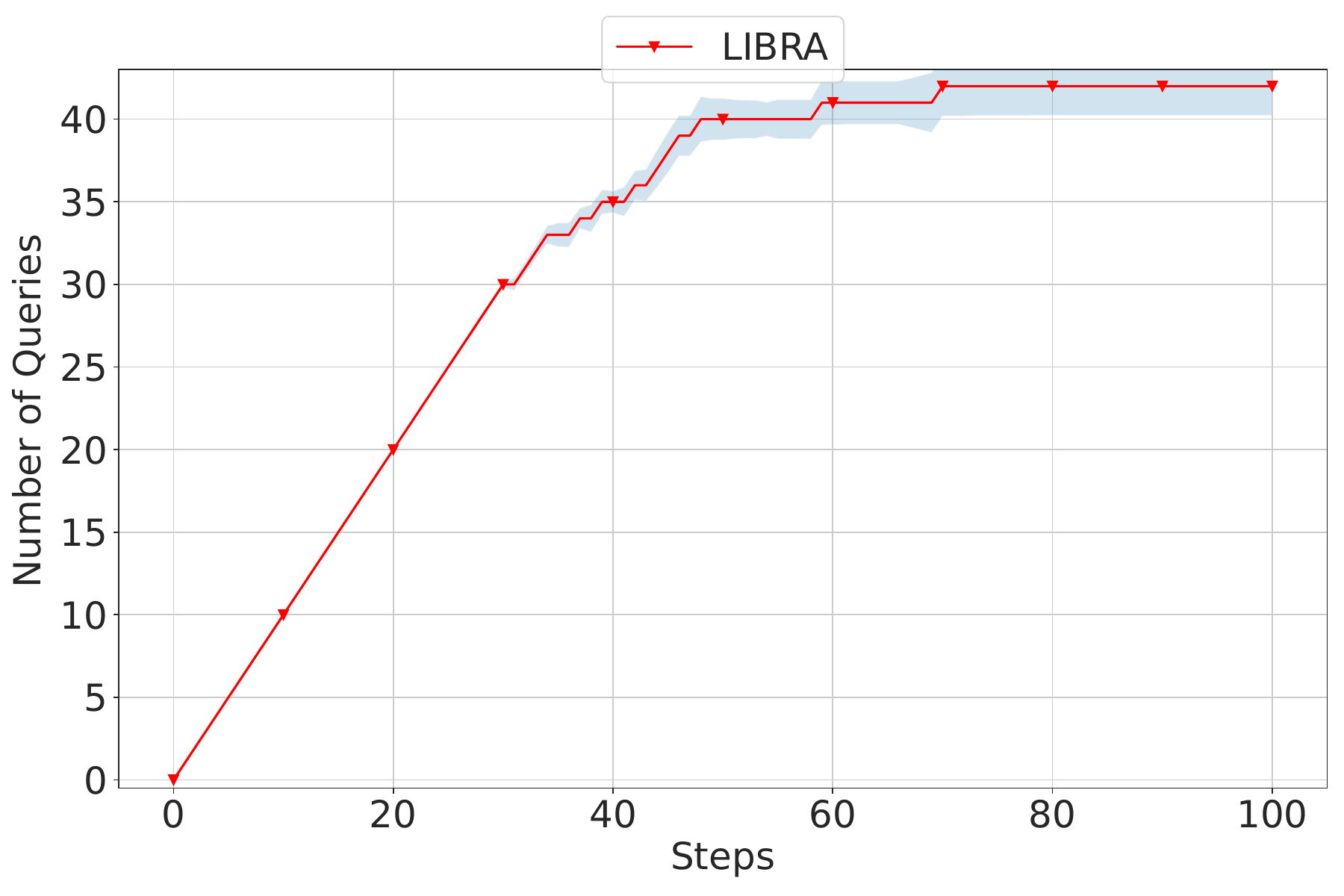}
    \caption{LLM Queries}
    \label{fig:fert_asked}
  \end{subfigure}
  \caption{Performance evaluation of different algorithms in terms of cumulative recourse regret and number of LLM queries using Fertility Dataset.}
  \label{fig:fert}
\end{figure}

In addition, we visualize the recourses output after step 70 by \textsf{LIBRA} in \Cref{fig:fert_recourses} (We note that while we did not restrict recourses to be negative, our algorithm can easily accommodate this constraint if needed).  We choose step 70 to show the recourse recommendation at a later stage since in the beginning there will be more explorations than exploitations.
The coefficients of alcohol consumption, smoking habit, and the number of hours sitting per day fitted by the linear regression are $[-0.31, -0.03, -0.15]$ and $[-0.03, -0.17, -0.12]$ for the surgery and no surgery groups, respectively. As a result, we observe that our algorithm encourages the patients to focus more on reducing alcohol consumption and sitting hours if the algorithm is going to recommend surgery and more on smoking if the algorithm decides surgery is not necessary. ~\looseness=-1

\begin{figure}[htp]
    \centering
    \includegraphics[width=0.5\linewidth]{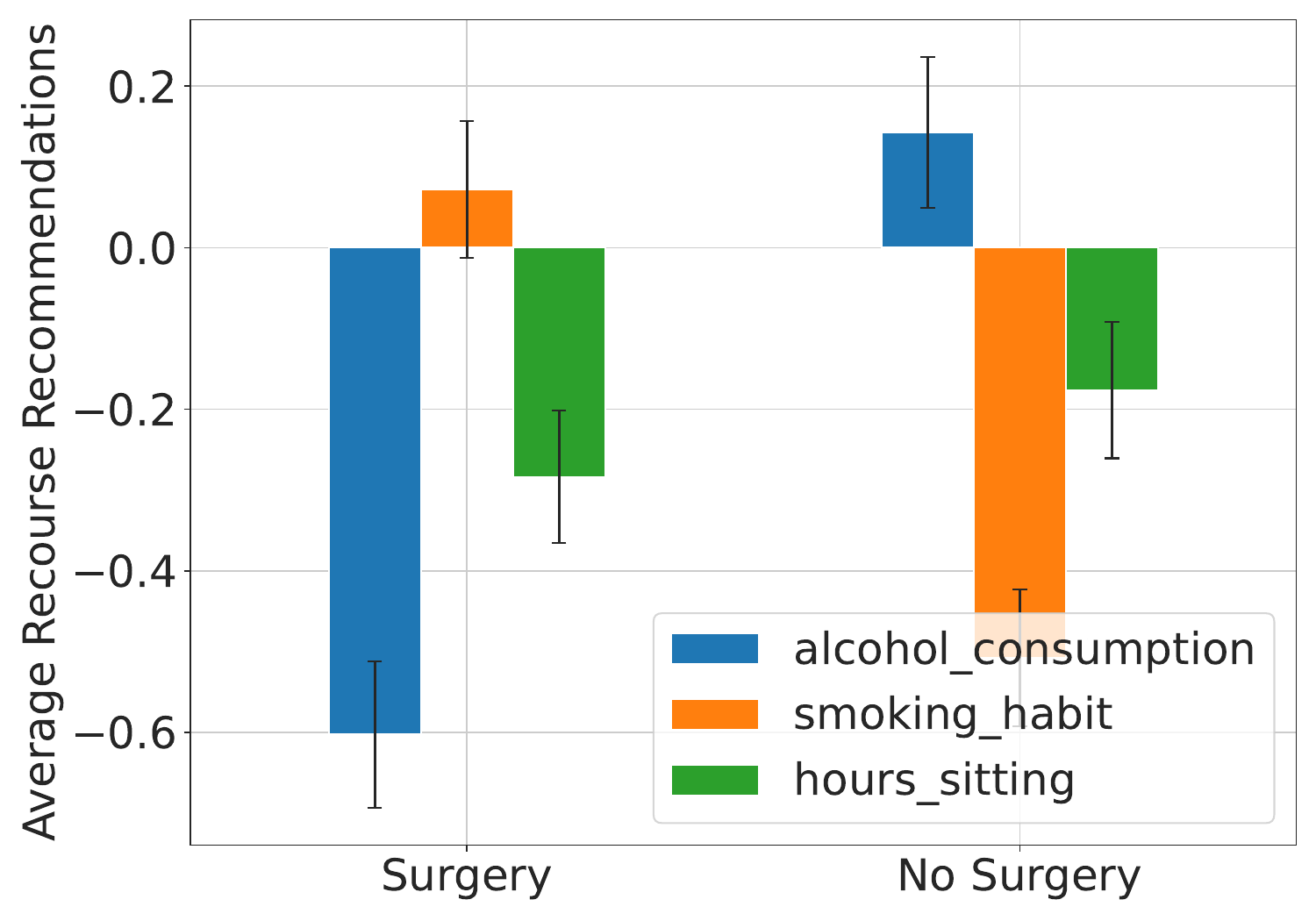}
    \caption{Recommended recourses under each treatment action.}
    \label{fig:fert_recourses}
\end{figure}

\end{APPENDICES}

\end{document}